\documentclass{article}

\usepackage[english]{babel}

\usepackage[a4paper,top=2cm,bottom=2cm,left=3cm,right=3cm,marginparwidth=2cm]{geometry}

\usepackage{setspace}
\usepackage{amsmath}
\usepackage{graphicx}
\usepackage{setspace}
\usepackage{multirow}
\usepackage{longtable}
\usepackage[english]{babel}
\usepackage[utf8]{inputenc}
\usepackage{algorithm}
\usepackage[noend]{algpseudocode}
\usepackage{csquotes}
\usepackage{subcaption}
\usepackage{graphicx}
\usepackage{tikz}
\usepackage{lipsum}
\usepackage{tcolorbox}
\usepackage{tabularx}
\usepackage{booktabs}
\usepackage{listings}
\usepackage{titlesec}
\usepackage{longtable}
\usepackage{rotating}
\usepackage{pdflscape}
\newlength{\RoundedBoxWidth}
\newsavebox{\GrayRoundedBox}
   {\setlength{\RoundedBoxWidth}{\dimexpr#1}
    \begin{lrbox}{\GrayRoundedBox}
       \begin{minipage}{\RoundedBoxWidth}}%
   {   \end{minipage}
    \end{lrbox}
    \begin{center}
    \begin{tikzpicture}%
       \draw node[draw=black,fill=black!10,rounded corners,%
             inner sep=2ex,text width=\RoundedBoxWidth]%
             {\usebox{\GrayRoundedBox}};
    \end{tikzpicture}
    \end{center}}

\usepackage{placeins} 
\newcolumntype{Y}{>{\raggedleft\arraybackslash}X}

\usepackage[colorlinks=true, allcolors=blue]{hyperref}

\usepackage{amsfonts}
\usepackage[citestyle=bwl-FU]{biblatex}
\graphicspath{ {./graphs/} }

\newcounter{noteMCctr} 

\newcounter{noteYZctr} 

\title{\textbf{Robust Detection of Lead-Lag Relationships in Lagged Multi-Factor Models}}

\author{Yichi Zhang$^{1,4}$, Mihai Cucuringu$^{1,2,4,7}$, Alexander Y. Shestopaloff$^{5,6}$, Stefan Zohren$^{3,4,7}$\\}

\date{
\textit{$^{1}$Department of Statistics, University of Oxford\\
        $^{2}$Mathematical Institute, University of Oxford\\
        $^{3}$Department of Engineering, University of Oxford\\
        $^{4}$Oxford-Man Institute of Quantitative Finance, University of Oxford\\
        $^{5}$School of Mathematical Sciences, Queen Mary University of London\\
        $^{6}$Department of Mathematics and Statistics, Memorial University of Newfoundland\\
        $^{7}$The Alan Turing Institute\\}
        }

\begin{document}
\maketitle
\centerline{\today}
\vspace{1cm}
\begin{abstract}
In multivariate time series systems, key insights can be obtained by discovering lead-lag relationships inherent in the data, which refer to the dependence between two time series shifted in time relative to one
another, and which can be leveraged for the purposes of control, forecasting or clustering. We develop a clustering-driven methodology for robust detection of lead-lag relationships in lagged multi-factor models. Within our framework, the envisioned pipeline takes as input a set of time series, and creates an enlarged universe of extracted subsequence time series from each input time series, via a sliding window approach. 
This is then followed by an application of various clustering techniques, 
(such as K-means++ and spectral clustering), employing a variety of pairwise similarity measures, including nonlinear ones. 
Once the clusters have been extracted, lead-lag estimates across clusters are robustly aggregated to enhance the identification of the consistent relationships in the original universe. We establish connections to the multireference alignment problem for both the homogeneous and heterogeneous settings. 
Since multivariate time series are ubiquitous in a wide range of domains, we demonstrate that our method is not only able to robustly detect lead-lag relationships in financial markets, but can also yield insightful results when applied to an environmental data set. 

    
\end{abstract} 

\newpage

\tableofcontents
\newpage

\section{Introduction}
When observed over time, natural physical systems frequently produce data recorded as high-dimensional, nonlinear time series, which are ubiquitous in a wide range of domains. Numerous contributions, covering different aspects of their analysis, have been made. For example,  \cite{cont2001empirical} discussed financial time series with a focus on various statistical properties, including distributional and tail properties, extreme fluctuations, etc. \cite{cartea2018enhancing} constructed a metric for assessing volume imbalance in the limit order book sourced from the Nasdaq exchange, and demonstrated it is a good predictor of the sign of the next market order. \cite{cao2020neural} utilized neural networks to analyze volatility surface movements based on daily call options data on the S\&P 500 index. \cite{drinkall2022forecasting} introduced a new technique that integrates time-series language models based on transformers into the field of infectious disease modelling. \cite{sokolov2022assessing} proposed a supervised machine learning framework for analyzing the impact of environmental, social and governance (ESG) factors on fund flows in US-domiciled equity mutual funds, and assessed whether sustainability has excess predictive power on fund-level flows as compared to benchmark driven by non-ESG factors. To detect change-points in nonlinear time series regression, \cite{cui2021state} utilized a density-weighted anti-symmetric kernel function and identified the presence of change-points within the state domain, rather than the time domain. Furthermore, \cite{vuletic2023fin} examined the feasibility of utilizing generative adversarial networks to forecast financial time series probabilistically, and to learn from asset co-movements while addressing problems associated with mode collapse.


Key insights regarding high-dimensional time series can be obtained by discovering latent structures. An example of a latent structure is lead-lag relationships, which are widely observed and found in the realms of 
finance [\cite{tolikas2018lead}, \cite{buccheri2021high}, \cite{miori2022returns}, \cite{bennett2022lead}, 
 \cite{albers2021fragmentation}, \cite{ito2020direct}, \cite{yao2020time}, \cite{li2021dynamic}], the environment [\cite{de2021detecting}, \cite{wu2010detecting}], and biology [\cite{runge2019detecting}]. For example, 
\cite{miori2022returns} explored lead-lag relationships within data-driven macroeconomic regimes by clustering the performance of diverse asset class indices in time relative to each other. \cite{bennett2022lead} constructed a directed network for encoding pairwise lead-lag relationships between time series of equity prices in the US equity market, in order to detect pairs of lead-lag clusters that exhibited a high pairwise directed flow imbalance. Another finance-related application study was derived from \cite{albers2021fragmentation}, who analyzed
the existence and strength of lead-lag relationships between pairs of Bitcoin markets. The importance and potentially high impact of this problem are broadly recognized; however, to date, there has been limited progress on
the robust detection of lead-lag relationships in high-dimensional time series. 

Clustering is widely used as part of the analysis of time series [\cite{zolhavarieh2014review}]. For example, \cite{lu2023co} developed a similarity measure between equities based on co-occurrence of trades from \cite{lu2022trade} and used spectral clustering algorithms [\cite{shi2000normalized}, \cite{ng2001spectral}, \cite{cucuringu2019sponge}] to detect dynamic communities within US equity markets. In particular, subsequence time series clustering, as one of time series clustering, groups similar subsequences into the same cluster. This type of clustering is used for detecting structures or patterns, and is typically used as a subroutine in rule discovery [\cite{uehara2002extraction}, \cite{sarker2003parallel}], indexing [\cite{li1998malm}, \cite{radhakrishnan2000alternate}], classification [\cite{cotofrei2002classification}, \cite{vijay2023earthquake}], prediction [\cite{tino2000temporal}, \cite{schittenkopf2002benefit}], and anomaly detection [\cite{yairi2001fault}, \cite{10091205}]. However, despite its wide use,  \cite{lin2003clustering} reported a surprising result: clustering subsequences extracted from a single time series via a sliding window is meaningless. They illustrated that all previous results involving clustering subsequences of a single time series were inaccurate since the resultant cluster centers appeared to be a form of sine waves, regardless of the initial patterns in the input data [\cite{keogh2005clustering}]. Afterwards, several lines of work  [\cite{madicar2013parameter}, \cite{rakthanmanon2011time}, \cite{rodpongpun2012selective}] proposed
solutions to the aforementioned meaningless results of subsequence time series clustering, and achieved meaningful time series clusters. In contrast, we take the approach of clustering every subsequence, extracted via a sliding window from a {\em set} of time series.

In this paper, we develop a clustering-driven methodology to enable the robust detection of lead-lag relationships in high-dimensional time series. In the proposed pipeline, we are given as input a set of $n$ time series. With this in mind, we create an enlarged universe $ \mathcal{U}$ 
of $N = n\times h$ time series by extracting  $h$ subsequence time series via a sliding window from each input time series. To this enlarged universe $ \mathcal{U}$, we apply various clustering techniques (e.g, K-means++ and spectral clustering) by employing various pairwise similarity measures between subsequence time series. The underlying clusters are then leveraged for the purpose of discovering the latent lead-lag relationships. In essence, the clustering step can be construed as an initial denoising step, in which we group together data relevant to estimating a subset of the lead-lag effects. Once the clusters have been extracted, the lead-lag estimates within each cluster are aggregated across clusters to enhance the identification of consistent relationships in the original universe. 
Our main contributions are summarized as follows.

\begin{tcolorbox}

\textbf{Summary of main contributions}.
\begin{enumerate}
    \item We introduce a computationally scalable pipeline for the robust detection of lead-lag relationships in high-dimensional time series.
    \item We demonstrate that our proposed methodology can reliably detect lead-lag relationships in a range of factor model-based simulated high-dimensional time series.
    \item In a financial market setting, we leverage the detected lead-lag relationships to construct a profitable trading strategy and show that our method outperforms the benchmark in most of cases.
    \item We apply our method to a data set of CO2 emissions and demonstrate that it achieves results consistent with the literature.
\end{enumerate}

\end{tcolorbox}

\textbf{Paper outline}. This paper is organized as follows. We first discuss  connections between lead-lag detection and the multireference alignment (MRA) problem in Section \ref{sec: mra}. Section \ref{sec: model} describes the {\em lagged multi-factor model} and establishes the notation used in this paper. Section \ref{sec: methodology} describes our proposed method for detecting lead-lag relationships in high-dimensional time series. In Section \ref{sec: synthetic}, we validate our method on synthetic data sets from the lagged multi-factor model, and show that our method can reliably recover them. Since high-dimensional time series data are ubiquitous in finance, our application domain mainly targeted financial time series. Within this domain, each time series corresponds to the excess return time series of a specific financial instrument. We explore lead-lag relationships in the US equity, ETF, and futures markets in Section \ref{sec: financial}. We then move on to a robustness analysis in Section \ref{sec: robustness}. To demonstrate the broader scope and applicability of our proposed algorithm, we also consider an application from the environmental sciences, looking at a data set of CO2 emissions in Section \ref{sec: CO2}. Finally, we summarize our findings in Section \ref{sec: conclusion}, and discuss possible future research directions.

\section{Connections to Multireference Alignment (MRA)} \label{sec: mra}

Our proposed framework for lead-lag detection has strong connections to the multireference alignment (MRA) problem, which we briefly describe here. MRA aims to estimate one signal from $n$ cyclically and noisy shifted copies of itself [\cite{bandeira2014multireference}]. 
In the \textbf{homogeneous} setting, let $x \in$ $\mathbb{R}^L$ be the unknown signal and let $R_r$ be the cyclic shift operator by $r$. We are given $n$ measurements of the form 

\begin{equation}
y_j=R_{r_j} x+\varepsilon_j, \quad j=1, \ldots, n,
\end{equation}

\noindent
where the $\varepsilon_j \sim \mathcal{N}\left(0, \sigma^2 I\right)$ are i.i.d.\ white Gaussian noise. The goal of MRA is to then estimate the unknown signal $x$, up to a shift, in a regime with high-noise in which the shifts $r_j$ are also unknown.

\begin{figure}[!htbp]
\centering
\includegraphics[width=\textwidth,trim=2cm 1.5cm 2cm 1.5cm,clip]{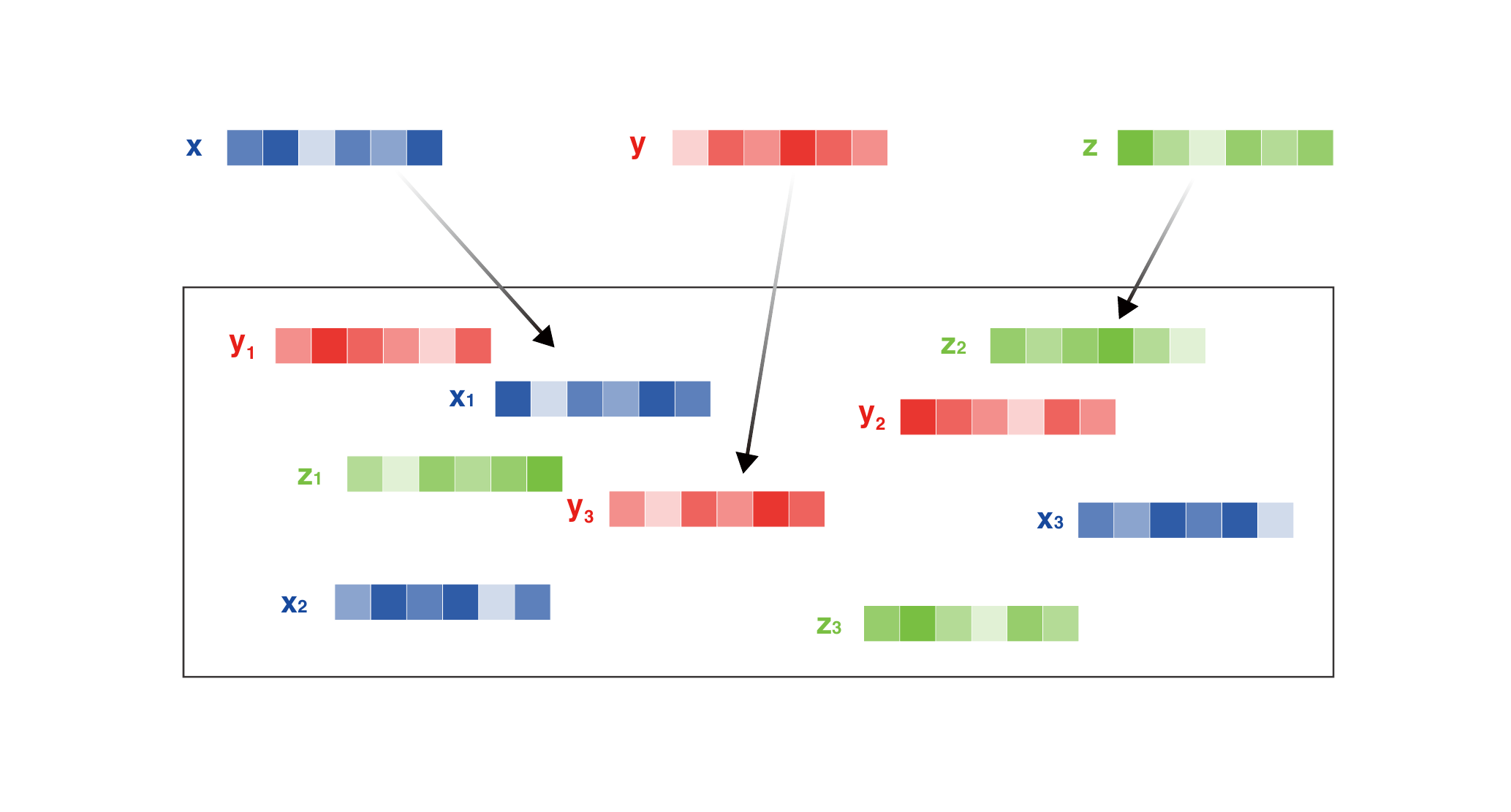}
\caption{Nine observations derived from one
of three signals with noisy cyclic shifts.} 
\label{fig:mra}
\end{figure}

In the \textbf{heterogeneous} setting that arises in areas such as cryo-electron microscopy, there are $k$ distinct signals $x_1, \ldots, x_k \in \mathbb{R}^L$ to be estimated. Each of the available $n$ observations is derived from one of these $k$ signals, yet the correspondence is unknown to the user. The model can be written as follows

\begin{equation}
y_j=R_{r_j} x_{v_j}+\varepsilon_j, \quad j=1, \ldots, n,
\end{equation}

\noindent
where the classes $v_j$ and the shifts $r_j$  are unknown, while the $\varepsilon_j$ are i.i.d.\ Gaussian noise of variance $\sigma^2$ as before. The goal is to estimate the signals $x_1, \ldots, x_k$, up to shifts and ordering. 
Figure \ref{fig:mra} illustrates an example of nine observations $x_i, y_i, z_i$ $(i = 1, 2, 3)$, each derived from one of the three signals $x, y, z \in \mathbb{R}^6$ with a noisy cyclic shift.

The work of [\cite{perry2019sample}] introduced the first known procedure to provably achieve signal recovery in a low signal-to-noise ratio (SNR) regime for heterogeneous MRA. Moreover, [\cite{boumal2018heterogeneous}] considered the application of MRA to the $2 \mathrm{D}$ class averaging problem in cryo-EM, aiming to achieve classification, alignment and averaging concurrently in a single pass, without using any of these steps explicitly. The authors of [\cite{boumal2018heterogeneous}] also proposed to use signal characteristics that are invariant under translations, and used them to recover the original signal. 
In the spirit of MRA, we will consider an analogous single-pass approach for robust detection of lead-lag relationships. 


\section{Model setup} \label{sec: model}

In this section, we will introduce the standard and lagged versions of the {\em multi-factor model}, which we will assume as a model for our time series data. The lagged multi-factor model will be used to validate our method on a synthetic data scenario before proceeding to the real-world applications. The gist of these models is to represent a time series as a (noisy) superposition of factors with varying exposures to each factor. We will also summarize the notation used in this paper. 

\subsection{Description}
We first recall the standard multi-factor model for a multivariate time series

\begin{equation} \label{eq:multi-factor_model}
X_{i}^{t} = \sum_{j=1}^{k} B_{ij}f_{j}^{t} + \epsilon_{i}^{t} \hspace{0.4cm} i = 1,\ldots,n;\quad t = 1,\ldots,T,
\end{equation}

\noindent
where $X_{i}^{t}$ is the time series $i$ (e.g., the excess return of a financial asset) at time $t$, $k$ is the number of factors, $B_{ij}$ is the exposure of time series $i$ to factor $j$, $f_{j}^{t}$ is the factor $j$ at time $t$, and $\epsilon_{i}^{t}$ is the noise at time $t$, with variance $\sigma^{2}$.  Furthermore, $n$ is the number of time series, and $T$ is the total number of time steps.

In this paper, we focus on the {\em lagged} version of the multi-factor model, which can be written as

\begin{equation} \label{eq:lagged_multi-factor_model}
X_{i}^{t} = \sum_{j=1}^{k} B_{ij}f_{j}^{t-L_{ij}} + \epsilon_{i}^{t} \hspace{0.4cm} i = 1,\ldots,n;\quad t = 1,\ldots,T, 
\end{equation}

\noindent
where the only difference compared to the standard multi-factor model is the addition of $L_{ij}$, the lag at which time series $i$ is exposed to factor $j$. Thus, $f_{j}^{t-L_{ij}}$ is the value of factor $j$ at time $t-L_{ij}$.

We introduce two main settings in the lagged multi-factor model (\ref{eq:lagged_multi-factor_model}), as follows.  

\begin{itemize}
  \item \textbf{Single Membership:} Each time series has a lagged exposure to a single factor. We consider the following two main categories. 
  \begin{itemize}
    \item \textbf{Homogeneous Setting:} The model only has one factor, i.e. $k=1$.
    \item \textbf{Heterogeneous Setting:} The model has more than one factor, i.e. $k \geq 2$. However, each time series is exposed only to a single factor.    
  \end{itemize}
  \item \textbf{Mixed Membership:} Each time series is allowed to have a lagged exposure to more than one factor, hence it is mixed. The model contains at least two factors, i.e. $k \geq 2$.
  
\end{itemize}

In this paper, our goal will be the inference of $L_{ij}$ in the lagged multi-factor model, focusing on the single membership setting. We do not focus on the inference of the unknown coefficient matrix $B$ and factors $f$. As shown later, the inference of $L_{ij}$ alone is of practical importance in certain applications, e.g.  finance. We leave the mixed membership setting for future research.

\subsection{Notation}

We first introduce the definition of a time series, subsequence time series and sliding window. 

\begin{itemize}
  \item \textbf{Time Series:} A time series $X_{i} = X_i^{1}, \ldots,X_i^{T}$ is an ordered set of $T$ real-valued variables.
  \item \textbf{Subsequence Time Series (STS):} Given a time series $X_{i} = X_{i}^{1}, \ldots, X_{i}^{T}$ of length $T$, a STS $Y_{i}^{z}$ of time series $X_{i}$ is a sample of length $q < T$ of contiguous positions from $X_{i}$ starting at $z$, that is, $Y_{i} = X_{i}^{z}, \ldots,X_{i}^{z+q-1}$ where $1 \leq z \leq T - q + 1$.
  \item \textbf{Sliding Window:} Given a set of time series $X_i^{t}$ $(i = 1,\ldots,n; t = 1,\ldots, T)$, and a user-defined STS length of $q$, an enlarged universe matrix $U$ of $N$ STS can be built by sliding a window shifted by $s$ across $X_i^{t}$. The size of the enlarged universe matrix $U$ is $N$ by $q$.
\end{itemize}

We summarize the remaining notation conventions used in the paper in Table \ref{tab: notation}.

\begin{table}[!htbp]
  \centering
  \caption{Variables and their description as used in our paper.}
    \begin{tabular}{p{2cm}p{14cm}}
        \toprule
    \textbf{Variables} & \textbf{Description} \\
        \midrule
    $k$                     &  Total number of factors \\
    $K$                     &  Total number of clusters \\
    $n$                     &  Total number of time series \\
    $N$                     &  Total number of STS in the enlarged universe \\
    $m$                     &  Number of lags \\
    $M$                     &  Maximum number of lags \\
    $t$                     &  Time \\
    $T$                     &  Length of time series \\
    $q$                     &  Length of STS \\
    $h$                     &  Number of STS in each time series \\
    $\mathcal{U}$           &  Universe set, it contains $N$ STS with length of $q$ \\
    $l$                     &  Length of sliding window \\
    $\sigma$                &  Noise \\
    $s$                     &  Value of sliding window shift \\
    $\delta$                &  Forward looking horizon windows \\
    $p$                     &  Past looking horizon windows \\
    $X_{n \times T}$        &  Time series matrix, where $X_{i}^{t}$ is the time series $i$ at time $t$ \\
    $X_{i}$                 &  Time series $i$ \\
    $Y_{i}^{z}$             &  STS from time series $i$ starting at $z$ \\ 
    $B_{n \times k}$        &  Loading matrix, where $B_{ij}$ is the exposure of time series $i$ to factor $j$ \\
    $f_{k \times 1}$        &  Vector of returns of the $k$ factors \\
    $f_{j}^{t}$             &  Value of returns of factor $j$ at time $t$ \\
    $\epsilon_{n \times 1}$ &  Vector of noise \\
    $L_{n \times k}$        &  Lag matrix, where $L_{ij}$ is the lag of time series $i$ to factor $j$ \\
    $U_{N \times q}$        & Universe matrix, where $U_{ij}$ is the STS $i$ at time $j$ \\
    $C_{N \times N}$        & Correlation matrix, where $C_{ij}$ is the correlation of STS $i$ and $j$ \\
    $P_{N \times N}$        & Sparse matrix, where $P_{ij}$ is assigned the weight of edge that connects STS $i$ and $j$ \\
    $G_{N \times N}$        & Similarity matrix, where $G_{ij}$ is the similarity of STS $i$ and $j$ \\
    $\phi_{d}$              & Cluster $d$ \\
    $\{X_{i}, X_{j}\}$      & Pair of time series $i$ and $j$ \\
    $\Delta_{d}\{X_{i}, X_{j}\}$   & Set of the relative lags between all pairs of time series $i$ and $j$ in cluster $d$ \\
    $V_{i \times j}$        & Voting matrix, where $v_{i \times j}$ is the number of lags between STS of $i$ and $j$ in the same cluster \\
    $\gamma\{X_{i}, X_{j}\}$   & The estimated value of the relative lags between all pairs of time series $i$ and $j$ in all clusters \\
    $\Gamma_{n \times n}$   & Lead-lag matrix, where $\Gamma_{ij}$ is the lead or lag value between time series $i$ and $j$ \\
    $E_{n \times n}$        & Error matrix, where $E_{ij}$ is the error of lead or lag value between time series $i$ and $j$ \\
    $\Psi_{n \times n}$     & Ground truth lead-lag matrix, where $\Psi_{ij}$ is the ground truth of lead or lag value between time series $i$ and $j$ \\
        \bottomrule
    \end{tabular}
\label{tab: notation}
\end{table}

\section{Methodology} \label{sec: methodology}

In this section, we present our methodology to infer unknown lags $L_{ij}$ in detail, before applying it to synthetic, financial, and environmental data set.

\begin{figure}[!htbp]
\centering
\includegraphics[width=\textwidth,trim=3cm 0cm 3cm 0cm,clip]{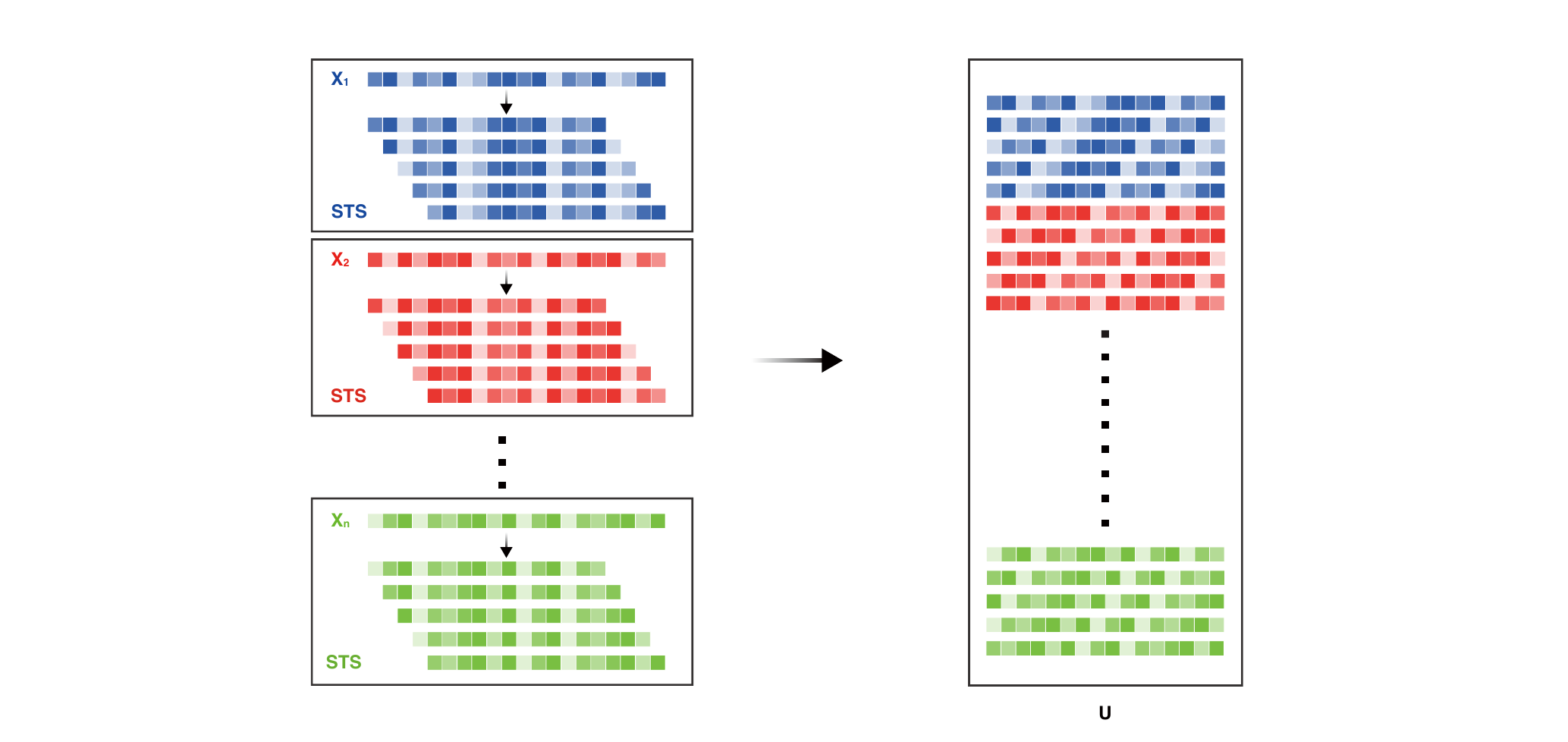}
\caption{Left: STS are extracted from each time series via a sliding window approach. Right: An enlarged universe matrix $U_{N \times q}$  is constructed by stacking all the extracted STS from the input set of time series $X_{n \times T}$.}
\label{fig:sts}
\end{figure}

We consider a set of time series $X_{n \times T}$ as our input. STS with length $q$ are extracted from each time series $X_i$ by a sliding window shifted by $s$. Therefore, the number of STS for each time series in the ensemble is $h = \frac{T-q}{s} + 1$, and the total number of STS from $X_{n \times T}$ is $N = n \cdot h$. An enlarged universe matrix $U_{N \times q}$ is constructed by collecting all STS, which is shown in Figure \ref{fig:sts}. Note that $U_{N \times q}$ contains exactly the same information as $X_{n \times T}$.

After the STS extraction step, we employ various clustering techniques to group similar STS in the same cluster. One way is to apply K-means++ (KM) clustering [\cite{arthur2006k}]. We follow this approach in Algorithm \ref{K-means++}, which initializes the cluster centers before proceeding with the standard K-means algorithm. With the KM initialization, a solution that is within  $O(\text{log}(k))$ of the optimal standard K-means solution is guaranteed.

\begin{algorithm}[!htbp] 
\caption{\textbf{\small : K-means++ Clustering}}
\label{K-means++}
\hspace*{\algorithmicindent} \textbf{Input:} Universe matrix $U_{N \times q}$, the total number of clusters $K$. \\
\hspace*{\algorithmicindent} \textbf{Output:} Clusters $\phi_1, \phi_2, ..., \phi_K$. 
\begin{algorithmic}[1]
\State Randomly select an initial center $\phi_1$ from $U$.
\State \textbf{Repeat} for $i \in 1,2, \ldots, K-1, K$. 
Select the next center $\phi_i=x \in U$ with the probability
$$
P(x)=\frac{D(x)^2}{\sum_{x^{\prime} \in U} D\left(x^{\prime}\right)^2},
$$
where $x^{\prime}$ is the closest center that has already been chosen and $D\left(x^{\prime}\right)$ is the distance to that center.
\State Continue with the standard K-means algorithm.
\end{algorithmic}
\end{algorithm}

\begin{algorithm}[!htbp] 
\caption{\textbf{\small : Spectral Clustering}}
\label{spectral_clustering}
\hspace*{\algorithmicindent} \textbf{Input:} Similarity matrix $G_{N \times N}$, the total number of clusters $K$. \\
\hspace*{\algorithmicindent} \textbf{Output:} Clusters $\phi_1, \phi_2, ..., \phi_K$.
\begin{algorithmic}[1] 
\State Compute normalized Laplacian $L$.
\State Compute the eigenvectors $v_1, v_2, ..., v_K$ corresponding to K smallest eigenvalues of $L$. 
\State Construct matrix $M \in \mathbb{R}^{N \times K}$ with $v_1, v_2, ..., v_K$ as columns.
\State Form matrix $\Tilde{M} \in \mathbb{R}^{N \times K}$ by normalizing row vectors of $M$ to norm 1.
\State Apply K-means clustering to assign rows of $\Tilde{M}$ to clusters $\phi_1, \phi_2, ..., \phi_K$. 
\end{algorithmic}
\end{algorithm}

An alternative method we considered is that of spectral (SP) clustering, described in Algorithm~\ref{spectral_clustering}. Note that if the size $U_{N \times q}$ is prohibitively large due to a large number $n$ of input time series, in order to speed up the computation of eigenvalues and eigenvectors, one could apply the K-nearest neighbours algorithm (KNN) on $U_{N \times q}$, which leads to the sparse matrix $P_{N \times N}$. The entries of the similarity matrix $G_{N \times N}$ are computed using a Gaussian kernel between neighbours from $P_{N \times N}$.

\begin{equation} \label{eq:Gaussian_kernel}
G_{ij} = \exp \left(-\left\|Y_i^a-Y_j^b\right\|^2 /\left(2 \sigma^2\right)\right),
\end{equation} 

\noindent
where the parameter $\sigma = 1 / N$ by default. Finally, we apply SP clustering on $G_{N \times N}$.
 
We denote time series $i$ and $j$ by $X_i$ and $X_j$ respectively, and a pair consisting of them by $\{X_{i}, X_{j}\}$. For each cluster $\phi_{d}$ $(d = 1,\ldots, K)$, we consider all possible pairwise relative lags of $X_i$ and $X_j$. We denote this set by $\Delta_{d}\{X_{i}, X_{j}\}$ and obtain it by computing the difference between starting indices of STS from $X_i$ and $X_j$ appearing in cluster $d$. Let $S_{i} = \{s_{i}^{1}, \ldots, s_{i}^{k}\}$ and $S_{j} = \{s_{j}^{1}, \ldots, s_{j}^{l}\}$ be the sets of STS starting indices of, respectively, $X_i$ and $X_j$. To avoid double counting, we consider only $i < j$. This amount to the following
\begin{equation}
\Delta_{d}\{X_{i}, X_{j}\} = \Delta_{d}\{(Y_{i}^{s_{i}^{1}}, Y_{j}^{s_{j}^{1}}), \ldots,(Y_{i}^{s_{i}^{k}}, Y_{j}^{s_{j}^{l}})\} = \{s_{i}^{1} - s_{j}^{1},\ldots, s_{i}^{k} - s_{j}^{l}\}.
\end{equation}

We then aggregate $\Delta_{d}\{X_{i}, X_{j}\}$ across all clusters and estimate the relative lag of $\{X_{i}, X_{j}\}$ by considering the mode or median of the resulting set

\begin{equation}
\gamma\{X_{i}, X_{j}\} = \left\{
     \begin{array}{@{}l@{\thinspace}l}
        \text{Mode}(\bigcup_{d=1}^{K}\Delta_{d}\{X_{i}, X_{j}\}) \hspace{0.6cm} \text{Mode
estimation} \\
        \text{Median}(\bigcup_{d=1}^{K}\Delta_{d}\{X_{i},X_{j}\}) \hspace{0.3cm} \text{Median
estimation}\\
     \end{array}
   \right.
\end{equation}

Table \ref{tab: algorithm} shows the example of two time series $X_1$ and $X_2$, for which the ground truth lag value is $3$. We extract eleven STS from each time series by a sliding window, and then perform KM clustering on these STS, setting the value of $K$ to 11. In each cluster, we calculate the relative lags of STS $\Delta_{d}\{X_1, X_2\}$ $(d = 1,\ldots, 11)$. After that, we aggregate lags across all clusters together, as $\bigcup_{d=1}^{11}\Delta_{d}\{X_{1}, X_{2}\} = \{-7,3,3,3,3,3,3,3,3,-10,-9\}$. Figure \ref{fig:hist_lags} displays the histogram of the relative lags of STS from $X_1$ and $X_2$.

\vspace{0.5cm}
\begin{minipage}{\textwidth}
\begin{minipage}[b]{0.5\textwidth}
  \centering
      \captionof{table}{Example of calculating the relative lags of STS in each cluster from two time series.}
    \begin{tabular}{p{1.5cm}p{3.5cm}p{1.5cm}}
        \toprule
        \textbf{Cluster} & \textbf{Subsequence} & \textbf{Lag}  \\
        \midrule
            $\phi_1$ & $(Y_1^9, Y_2^{2})$ & -7 \\
            $\phi_2$ & $(Y_1^7, Y_2^{10})$ & 3 \\
            $\phi_3$ & $(Y_1^1, Y_2^{4})$ & 3 \\
            $\phi_4$ & $(Y_1^2, Y_2^{5})$ & 3 \\
            $\phi_5$ & $(Y_1^0, Y_2^{3})$ & 3 \\
            $\phi_6$ & $(Y_1^6, Y_2^{9})$ & 3 \\
            $\phi_7$ & $(Y_1^3, Y_2^{6})$ & 3 \\
            $\phi_8$ & $(Y_1^4, Y_2^{7})$ & 3 \\
            $\phi_9$ & $(Y_1^5, Y_2^{8})$ & 3 \\
            $\phi_{10}$ & $(Y_1^10, Y_2^{0})$, $(Y_1^10, Y_2^{1})$ & -10, -9 \\
            $\phi_{11}$ & $Y_1^8$ & NaN \\
        \bottomrule
    \end{tabular}
\label{tab: algorithm}

\end{minipage}
\hfill
\begin{minipage}[b]{0.47\textwidth}
\centering
\includegraphics[width=\textwidth,trim=0cm 0cm 0cm 0cm,clip]
{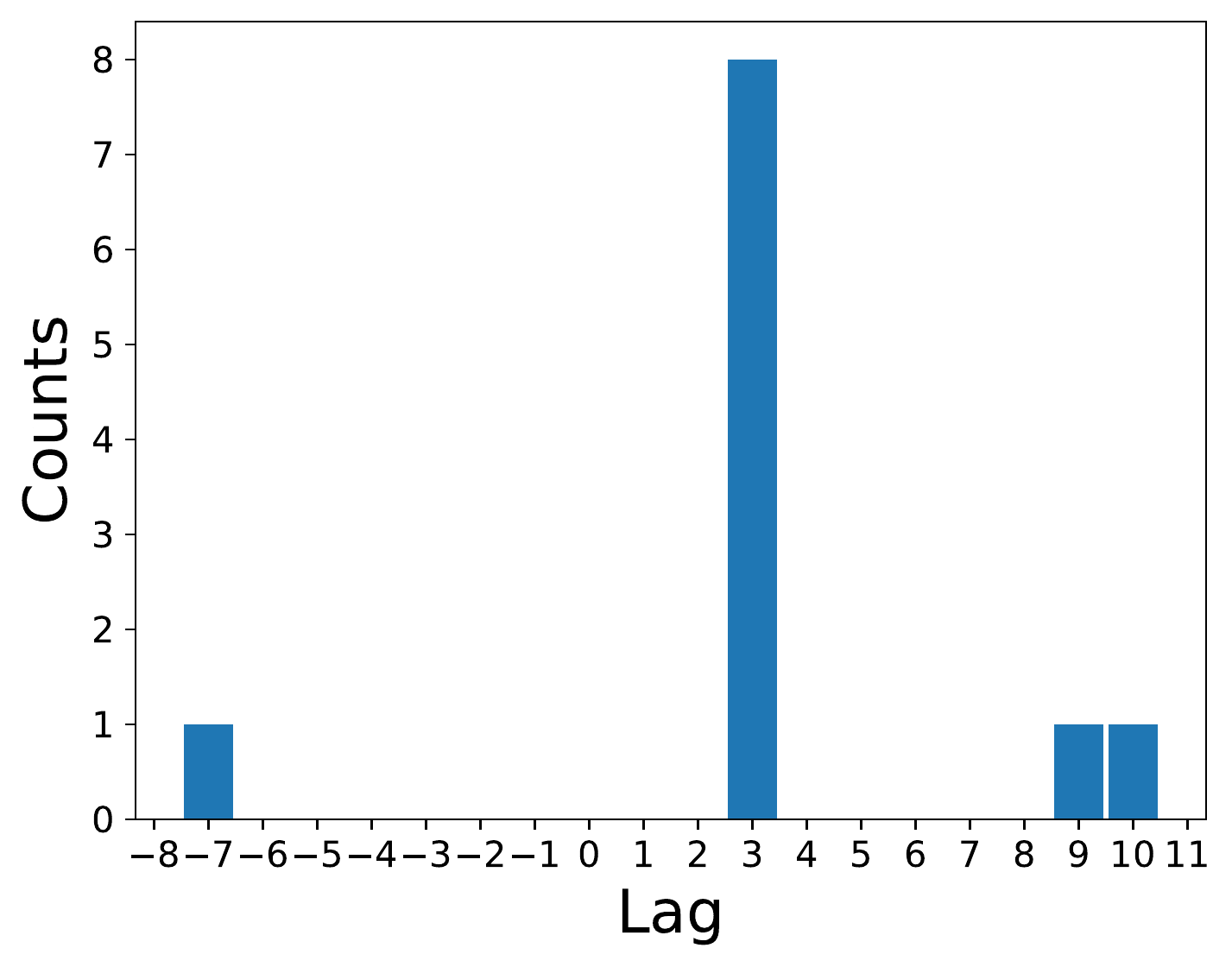}
\captionof{figure}{Histogram of the relative lags of STS from two time series.}
\label{fig:hist_lags}
\end{minipage}
\end{minipage}
\vspace{0.5cm}

By considering the mode or median of $\bigcup_{d=1}^{11}\Delta_{d}\{X_{1}, X_{2}\}$, we arrive at $\gamma\{X_{1}, X_{2}\}=3$. From Table \ref{tab: algorithm} and Figure \ref{fig:hist_lags}, we observe that even though there are outliers $\{-7,-10,-9\}$ in $\bigcup_{d=1}^{11}\Delta_{d}\{X_{1}, X_{2}\}$, we are still able to correctly recover the ground truth value of $3$.

A refinement of the algorithm is to calculate a {\em voting matrix} $V_{n \times n}$ by counting the sum of the number of lags between STS from $X_i$ and $X_j$ in the same clusters. We can then set a voting threshold denoted by $\theta$ to filter out a small number of counts. This is motivated by the reasoning that unless STS corresponding to a pair of time series are consistently clustered together across different clusters, the resulting lead-lag estimate is unlikely to be accurate. This altogether amounts to 

\begin{equation}
V_{ij} = \left\{
     \begin{array}{@{}l@{\thinspace}l}
       |\bigcup_{d=1}^{K}\Delta_{d}\{X_{i}, X_{j}\}|  & \hspace{1cm}\text{if}\hspace{0.3cm} |\bigcup_{d=1}^{K}\Delta_{d}\{X_{i}, X_{j}\}| \geq \theta \\
       0  & \hspace{1cm}\text{otherwise}\\
     \end{array}
   \right.
\end{equation}

\noindent
where $|\cdot|$ counts the number of elements in the $\bigcup_{d=1}^{K}\Delta_{d}\{X_{i}, X_{j}\}$.

Finally, the lead-lag matrix $\Gamma_{n \times n}$ is built by 

\begin{equation}
\Gamma_{ij} = \left\{
     \begin{array}{@{}l@{\thinspace}l}
       \gamma\{X_{i}, X_{j}\}  & \hspace{1cm}\text{if}\hspace{0.3cm} V_{ij} \neq 0\\
       0  & \hspace{1cm}\text{otherwise}\\
     \end{array}
   \right.
\end{equation}

We summarize the above procedures in Algorithm \ref{lead-lag_detection}, which is our main algorithm.

\begin{algorithm}[htp] 
\caption{\textbf{\small: Lead-lag Relationship Detection Algorithm}}
\label{lead-lag_detection}
\hspace*{\algorithmicindent} \textbf{Input:} Time series matrix $X_{n \times T}$. \\
\hspace*{\algorithmicindent} \textbf{Output:} Lead-lag matrix $\Gamma_{n \times n}$.
\begin{algorithmic}[1]
\State STS $Y_{i}^{p}$ are extracted from each time series $X_i$ by a sliding window.
\State An enlarged universe matrix $U_{N \times q}$ is created in Step 2.
\State Apply KNN to create the sparse matrix $P_{N \times N}$, and the similarity matrix $G_{N \times N}$ is then only computed using the Gaussian kernel between neighbours from $P_{N \times N}$.
\State Clusters are extracted by performing KM clustering on $U_{N \times q}$ or SP clustering on $G_{N \times N}$. For each cluster, record lags between every pair of time series $\{X_{i}, X_{j}\}$.
\State For each pair of time series $\{X_{i}, X_{j}\}$, we calculate the voting matrix $V_{ij}$ by counting lags between STS from $X_i$ and $X_j$ across all clusters. We use a voting threshold $\theta$ to filter small counts.
\State Calculate the lead-lag matrix $\Gamma_{n \times n}$ by considering \textit{mode} or \textit{median} of lags based on the voting matrix $V_{n \times n}$.
\end{algorithmic}
\end{algorithm}

\newpage
\section{Synthetic data experiments}
\label{sec: synthetic}

The purpose of our synthetic data experiments is to assume that our data is generated by a multi-factor model with known ground truth lead-lag matrix $L$, and then validate the performance of our proposed algorithms under different scenarios.

\subsection{Setup}

As noted earlier, our focus is the single membership setting. We generate synthetic data from the lagged multi-factor model (\ref{eq:lagged_multi-factor_model}) with $k = \{1,2,3\}$ factors. 
We let $M=5$, $T=100$ and $n=6$. The factors $f$ and errors $\epsilon$ are assumed to be i.i.d. $\mathcal{N}(0,1)$. We define $B$ and $L$ as follows:





\begin{table}[htbp]
  \centering
  \captionof{table}{Top row: Loading matrix $B$. Bottom row: Lag matrix $L$.}
    \begin{tabular}{p{0.3cm}p{4.5cm}|p{4.5cm}p{4.5cm}}
      \multicolumn{1}{c}{} &
      \multicolumn{1}{c}{\textbf{Homogeneous Setting}}    &  \multicolumn{2}{c}{\textbf{Heterogeneous Setting}}  \\

\multirow{8}{*}{\textit{B}}
\multirow{24}{*}{}
\multirow{24}{*}{}
\multirow{24}{*}{}
\multirow{18}{*}{\textit{L}}
    
    &

    \begin{equation*}
    \left[\begin{array}{ccc} 
    	1 \\
    	1 \\
    	1 \\
    	1 \\
    	1 \\
    	1 \\
    \end{array}\right] 
    \end{equation*}
    
    \begin{equation*}
    \left[\begin{array}{ccc} 
    	0 \\
    	1 \\
    	2 \\
    	3 \\
    	4 \\
    	5 \\
    \end{array}\right] 
    \end{equation*}

    &
    \begin{equation*}
    \left[\begin{array}{ccc} 
    	1 & 0 \\
    	1 & 0 \\
    	1 & 0 \\
    	0 & 1 \\
    	0 & 1 \\
    	0 & 1 \\
    \end{array}\right] 
    \end{equation*}
    
    \begin{equation*}
    \left[\begin{array}{ccc} 
    	0 & 0 \\
    	2 & 0 \\
    	4 & 0 \\
    	0 & 0 \\
    	0 & 2 \\
    	0 & 4 \\
    \end{array}\right] 
    \end{equation*}

    &
    \begin{equation*}
    \left[\begin{array}{ccc} 
    	1 & 0 & 0\\
    	1 & 0 & 0 \\
    	0 & 1 & 0\\
    	0 & 1 & 0 \\
    	0 & 0 & 1\\
    	0 & 0 & 1 \\
    \end{array}\right] 
    \end{equation*}
    
    \begin{equation*}
    \left[\begin{array}{ccc} 
    	0 & 0 & 0\\
    	3 & 0 & 0 \\
    	0 & 0 & 0\\
    	0 & 3 & 0 \\
    	0 & 0 & 0\\
    	0 & 0 & 3 \\
    \end{array}\right] 
    \end{equation*}
    \\
    \multicolumn{1}{c}{} &\multicolumn{1}{c}{$k=1$} & \multicolumn{1}{c}{$k=2$} & \multicolumn{1}{c}{$k=3$}  \\
    \end{tabular}
\end{table}

When estimating the lead-lag matrix, we use a sliding window of length $q=90$ and a shift of $s=1$. After estimating the lead-lag matrix, we calculate the error matrix $E$ to evaluate the performance of the method, which we denote as 
\begin{equation} \label{eq:error_matrix}
E_{n \times n} = \Gamma_{n \times n} - \Psi_{n \times n},
\end{equation}
where $\Gamma_{n \times n}$ is the estimated lead-lag matrix, and $\Psi_{n \times n}$ is the ground truth lead-lag matrix, which can be obtained from $L_{n \times k}$.

\subsection{Results}

We first explore the data by computing the Pearson and distance correlations between STS in the universe matrix $U_{N \times q}$, for scenarios with different $k$. Figure \ref{tab:heatmap} shows the correlation between STS in the two settings, homogeneous and heterogeneous. Each heatmap represents similarities between the STS, with a darker shade of red corresponding to a higher similarity. 

For the homogeneous setting ($k = 1$), we note that all $6 \times 6$ blocks, with each block corresponding to a different time series, have groups of red pixels stretching diagonally from the top left to the bottom right. This reflects a high similarity between pairs of STS starting at different initial time points. Thus, for example, time series 1 and time series 2 have highest correlation between STS at a lag of 1, which reflects the ground truth. For the heterogeneous setting ($k \in \{2, 3\}$), we note that only the diagonal blocks show high similarities between STS, which reflects the structure of $B$. Again, we note a high similarity between STS at the ground truth lag value. Our proposed method clusters these similar STS together to estimate the lag between the corresponding time series.

\begin{table}[htbp]
  \centering
    \begin{tabular}{p{4.5cm}|p{4.5cm}p{4.5cm}}
      \multicolumn{1}{c}{\textbf{Homogeneous Setting}}    &  \multicolumn{2}{c}{\textbf{Heterogeneous Setting}}  \\

      \includegraphics[width=\linewidth,trim=0cm 0cm 0.25cm 0cm,clip]{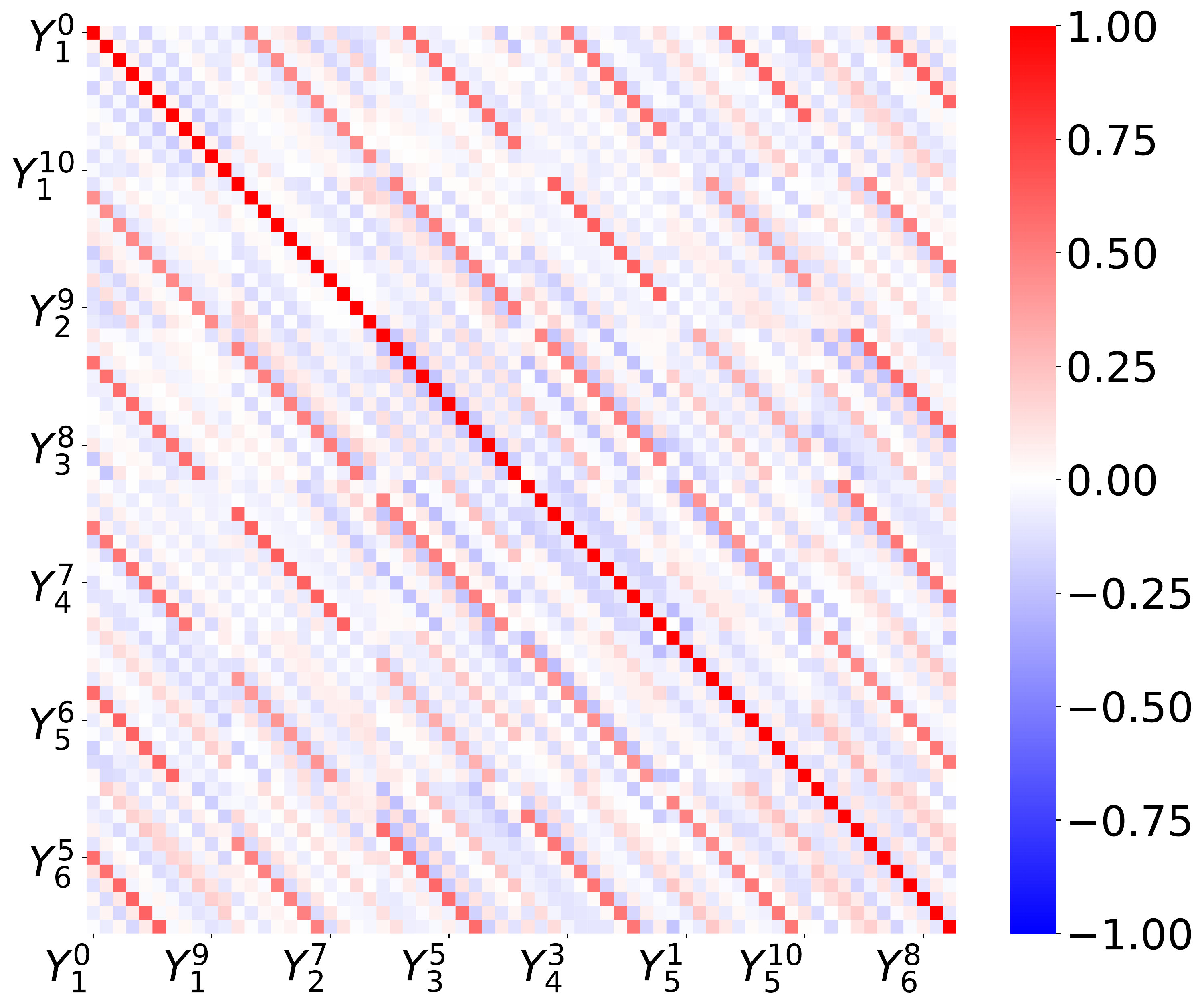}

    \includegraphics[width=\linewidth,trim=0cm 0cm 0.25cm 0cm,clip]{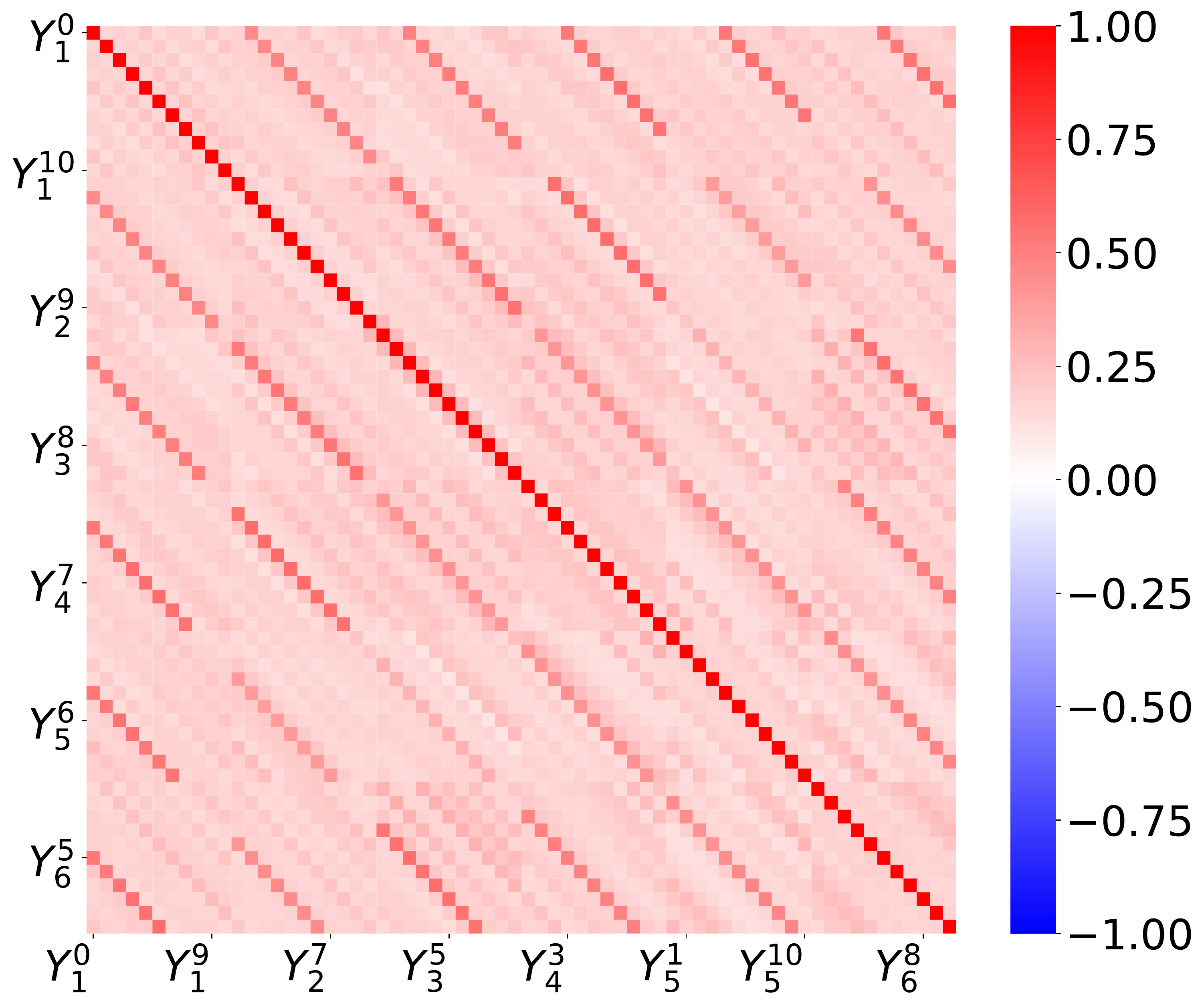}

    &
      \includegraphics[width=\linewidth,trim=0cm 0cm 0.25cm 0cm,clip]{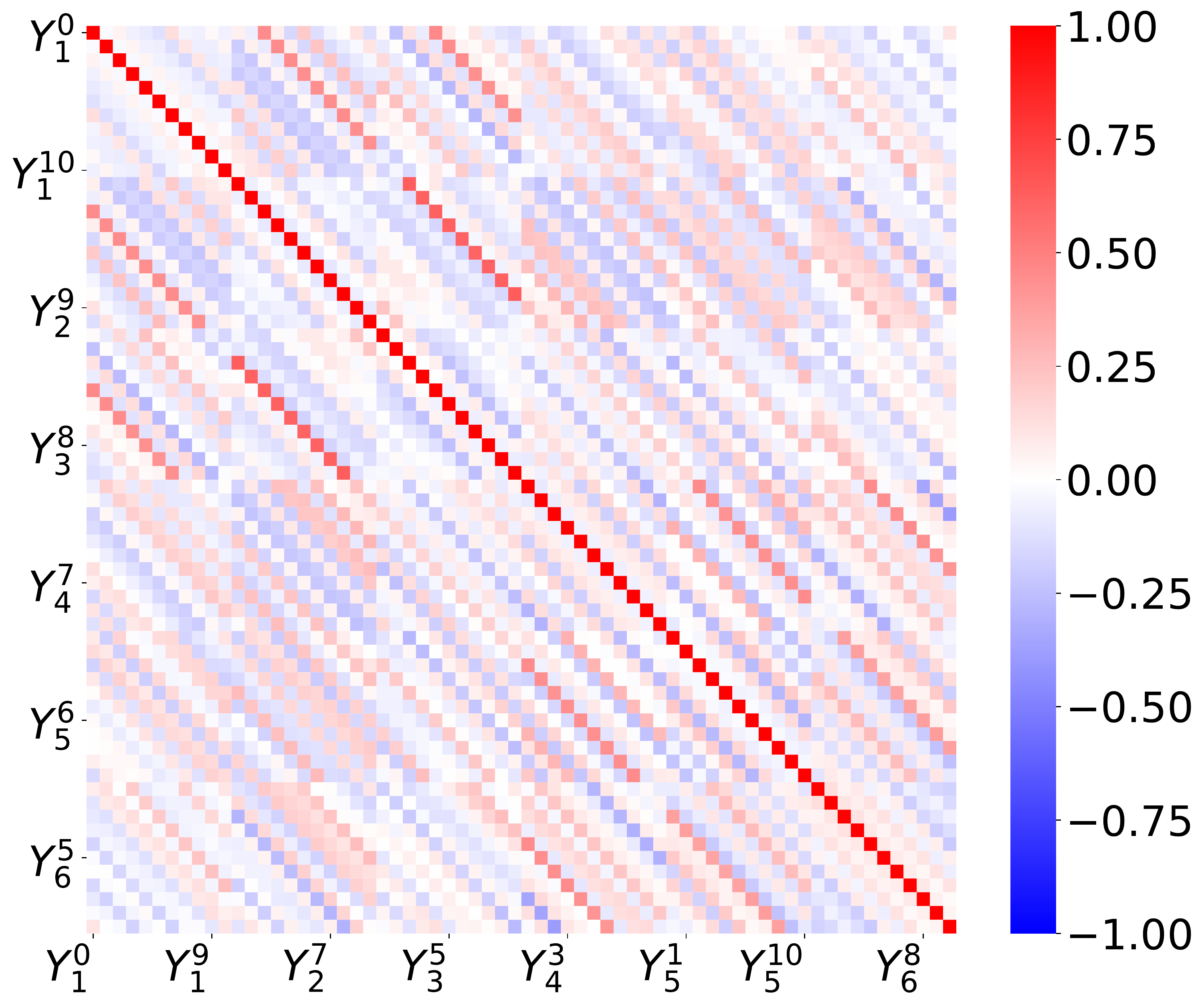}
     \includegraphics[width=\linewidth,trim=0cm 0cm 0.25cm 0cm,clip]{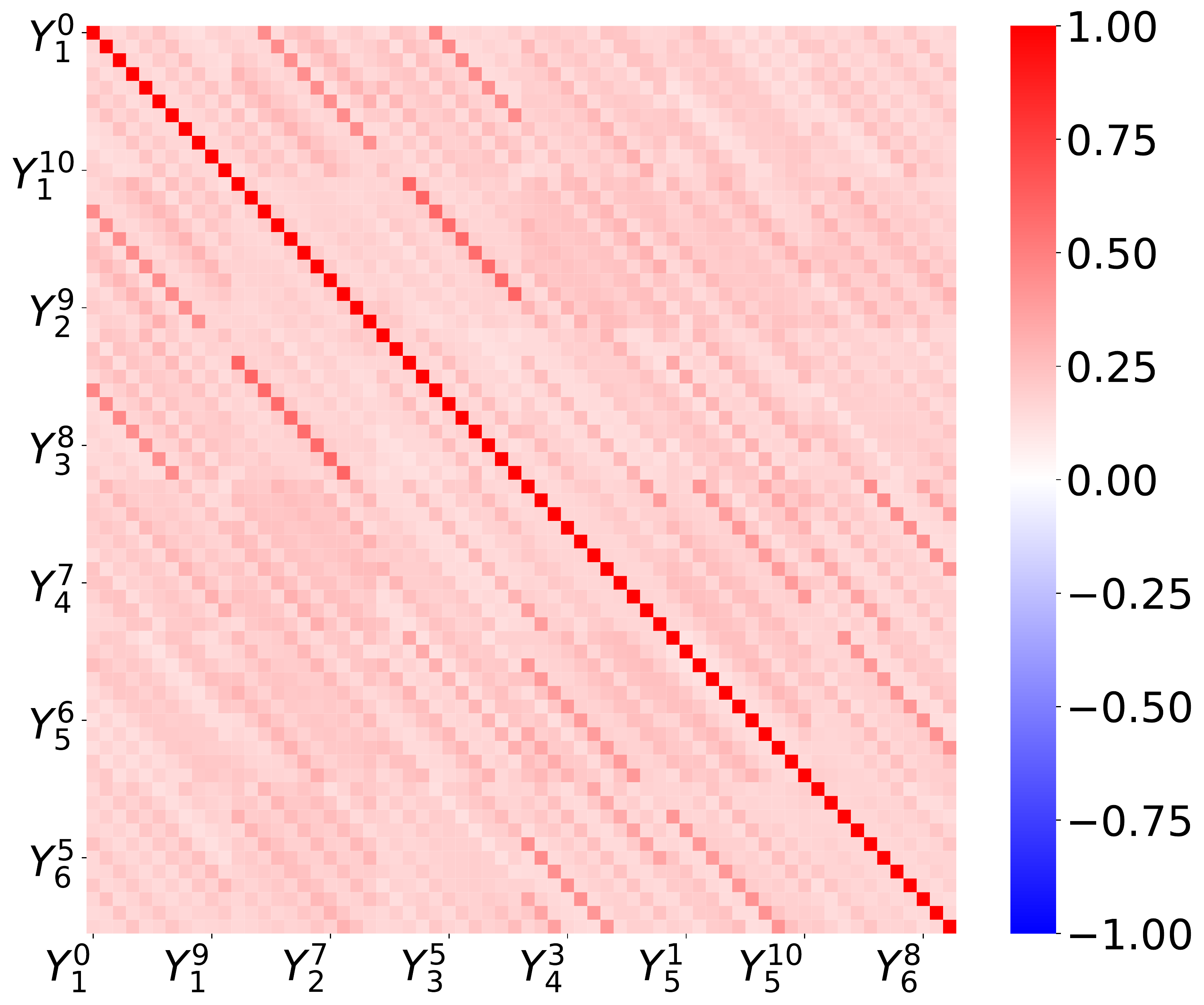}

    &
      \includegraphics[width=\linewidth,trim=0cm 0cm 0.25cm 0cm,clip]{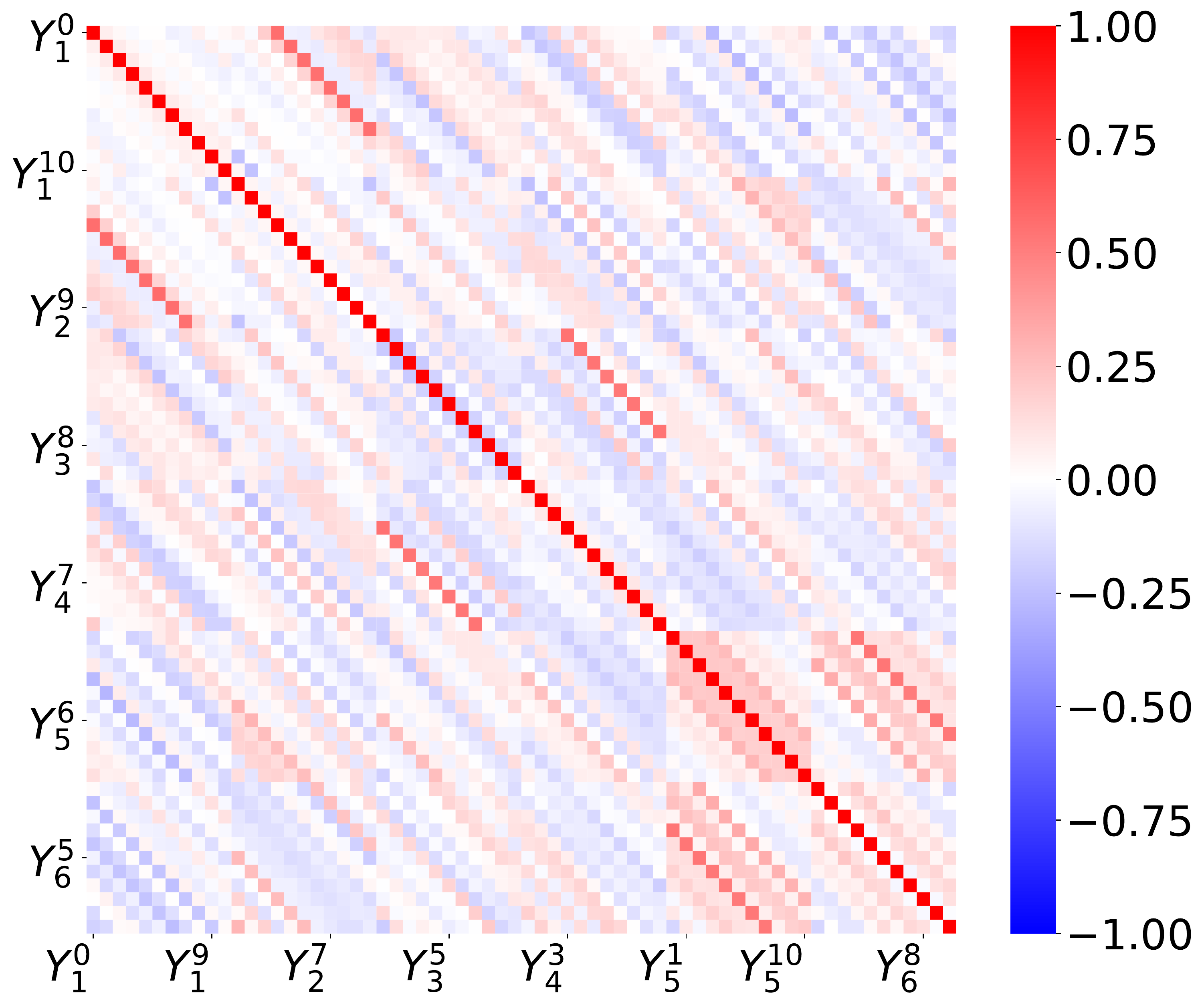}
      \includegraphics[width=\linewidth,trim=0cm 0cm 0.25cm 0cm,clip]{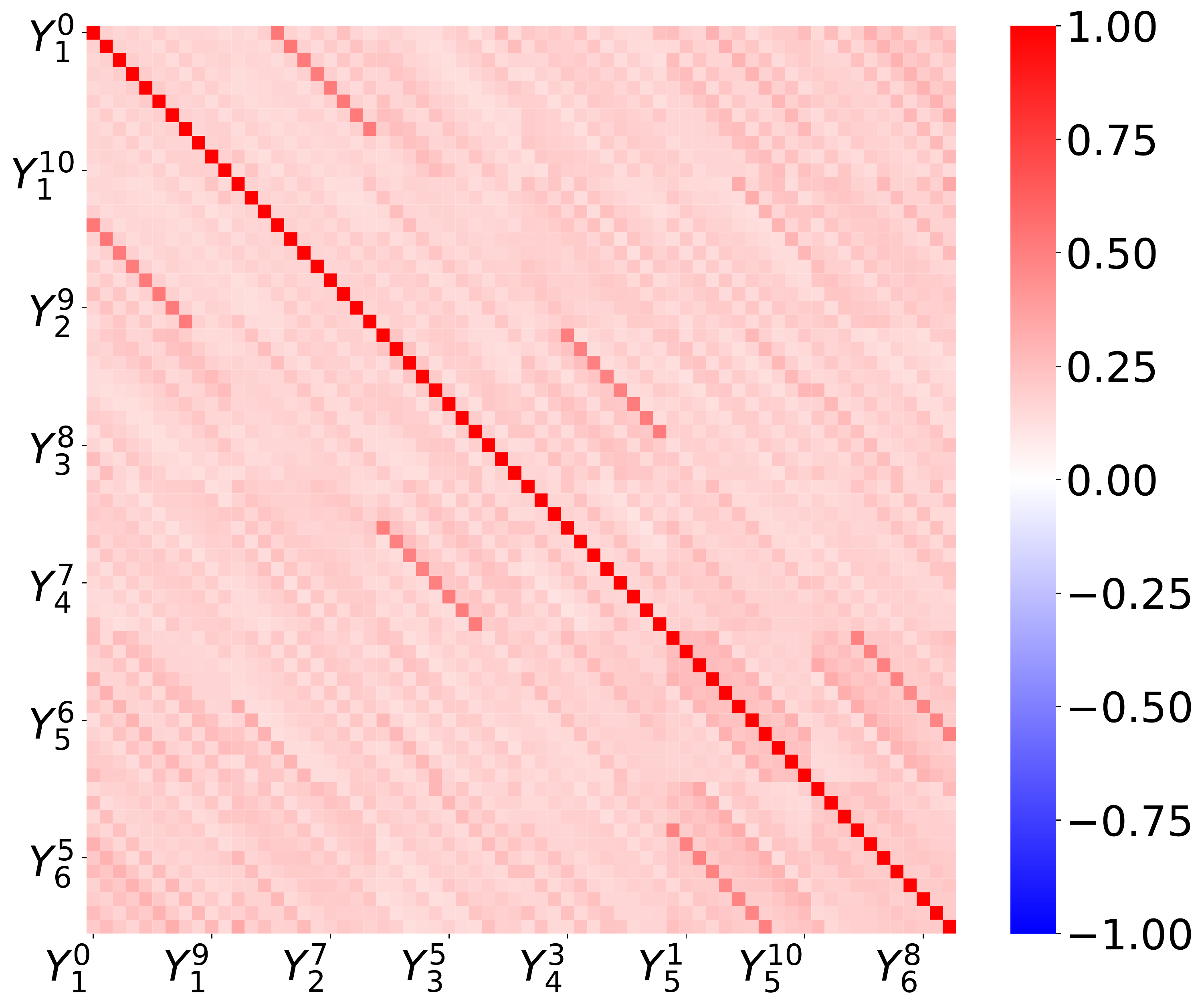}
    \\
    \multicolumn{1}{c}{$k=1$} & \multicolumn{1}{c}{$k=2$} & \multicolumn{1}{c}{$k=3$}  \\
    \end{tabular}
\captionof{figure}{Top panel: Heatmap of enlarged universe $U_{N \times q}$ by Pearson correlation. Bottom panel: Heatmap of enlarged universe $U_{N \times q}$ by distance correlation.}
\label{tab:heatmap}
\end{table}

\noindent

We next follow the methodology described in Section \ref{sec: methodology} to estimate the lead-lag matrix. In what follows, we use KM clustering to cluster STS. We set the number of clusters for a model with $k$ factors to $11 \cdot k$. A similar result, utilizing SP clustering, is shown in Appendix \ref{appendix_spectral_clustering} Figures [\ref{fig:spectral_vote_matrix_before_plus_error_matrix_befores}, \ref{fig:spectral_vote_matrix_after_plus_error_matrix_afters}]. We first consider not setting a voting threshold ($\theta=1$). 
Figure \ref{fig:kmean_vote_matrix_before_plus_error_matrix_befores} contains the number of votes associated with each time series pair. The top plots denote the voting matrix, the middle plots represent the error matrix for the mode estimation, and the bottom plots correspond to the median estimation. 
For the homogeneous case, we can observe that the voting matrix consists of all $0$s across the diagonal and is non-zero for all other entries. The error matrices based on mode and median estimation consist of all $0$s, which indicates our algorithm fully recovers the lead-lag relationships between these six time series. However, for the heterogeneous setting, we can observe that the error matrices have some non-zero values.

\begin{table}[htbp]
  \centering
    \begin{tabular}{p{4.5cm}|p{4.5cm}p{4.5cm}}
      \multicolumn{3}{c}{\textbf{K-means++ clustering}} \vspace{0.3cm}\\
      
      \multicolumn{1}{c}{\textbf{Homogeneous Setting}}    &  \multicolumn{2}{c}{\textbf{Heterogeneous Setting}}  \\

      \includegraphics[width=\linewidth,trim=0cm 0cm 1.5cm 1cm,clip]{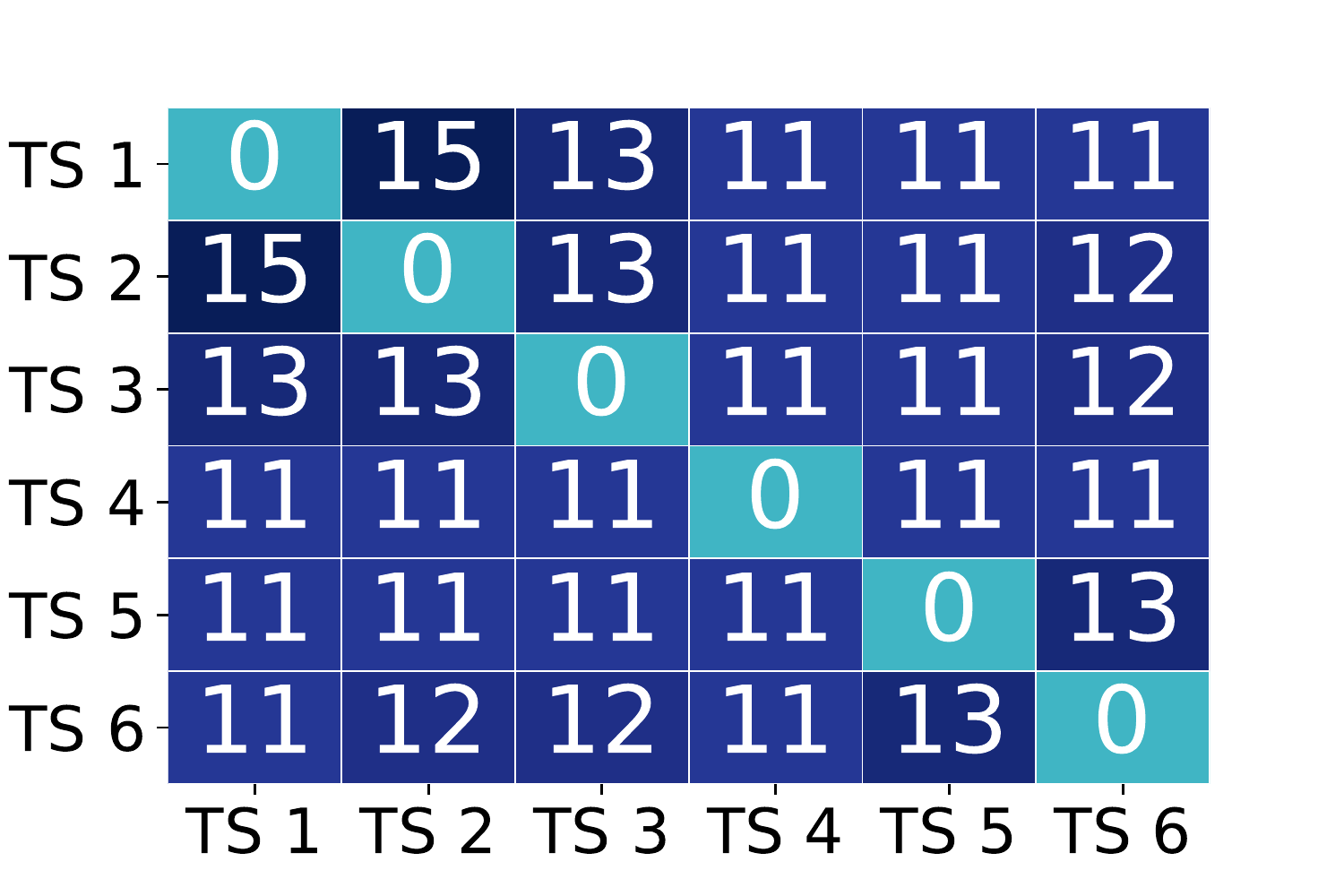}
      
      \includegraphics[width=\linewidth,trim=0cm 0cm 1.5cm 1cm,clip] {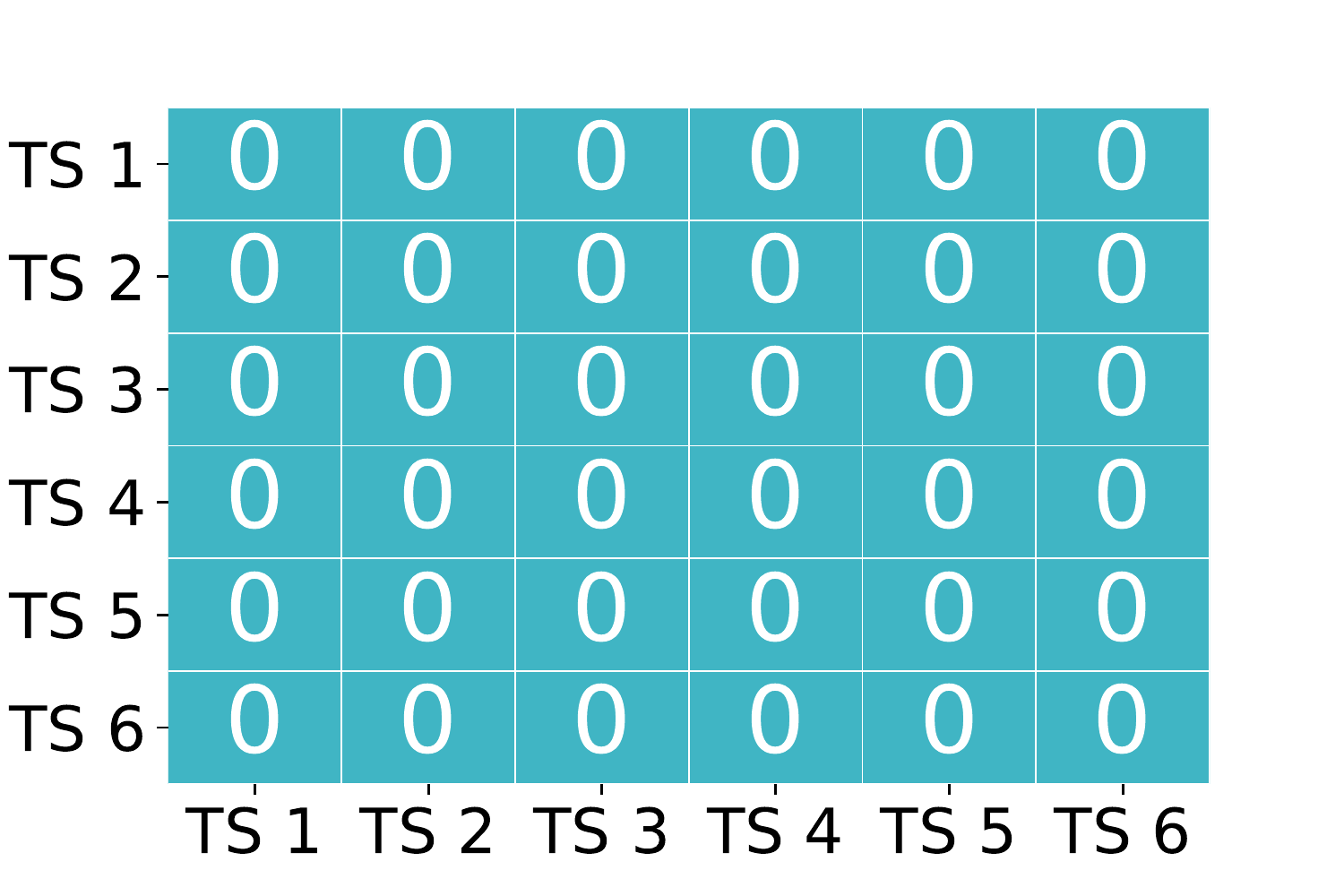}

    \includegraphics[width=\linewidth,trim=0cm 0cm 1.5cm 1cm,clip] {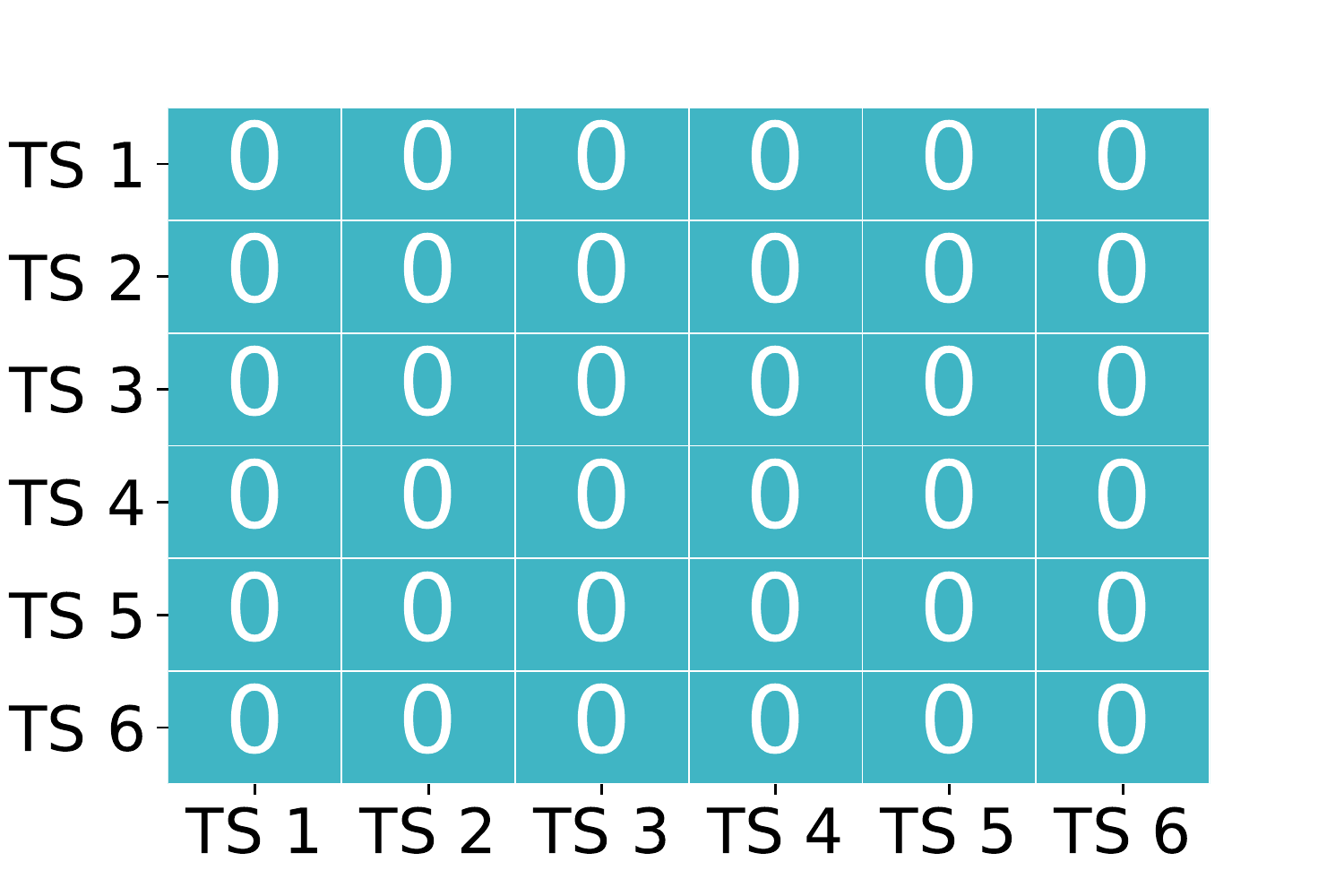}

    &
      \includegraphics[width=\linewidth,trim=0cm 0cm 1.5cm 1cm,clip]{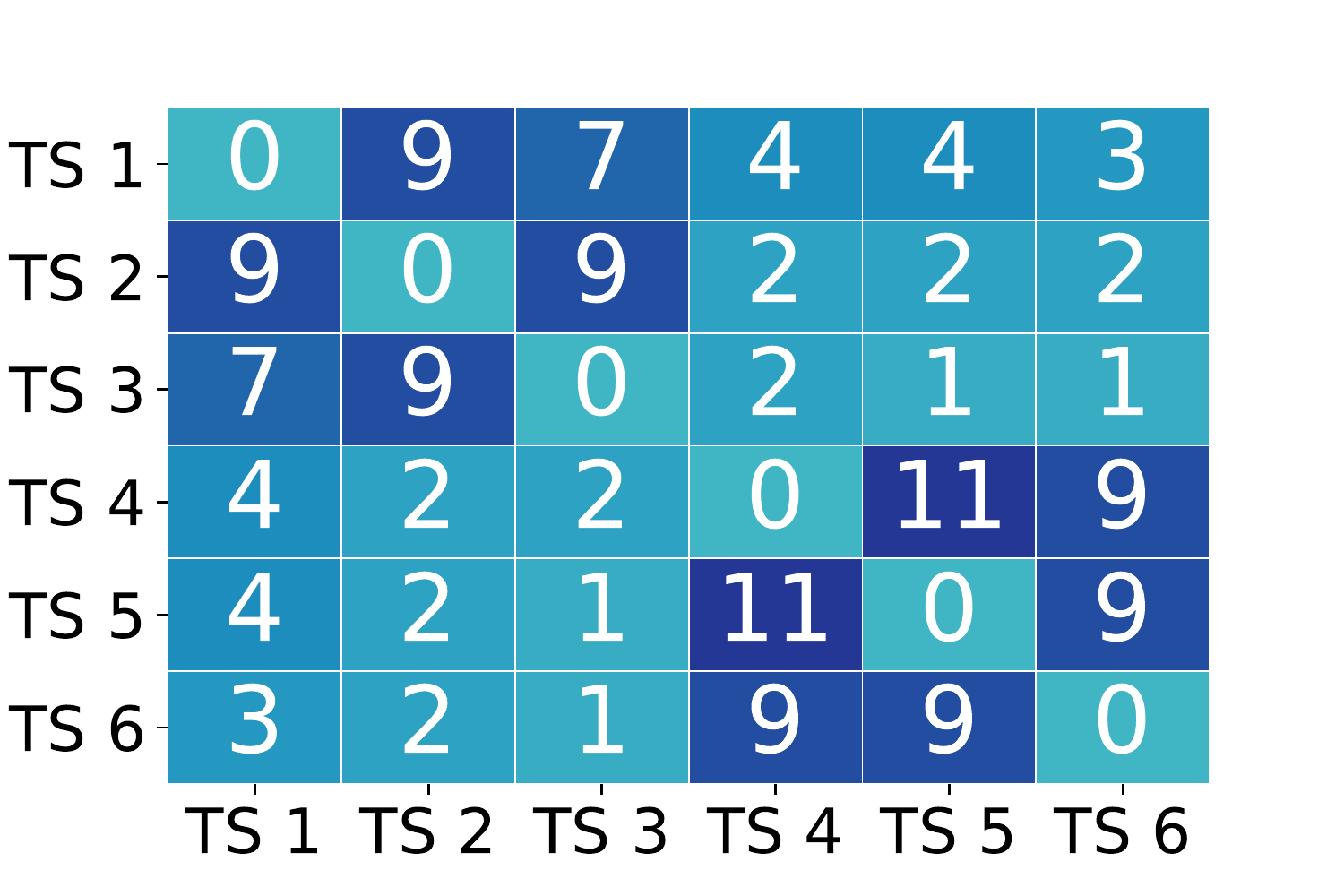}
      
      \includegraphics[width=\linewidth,trim=0cm 0cm 1.5cm 1cm,clip] {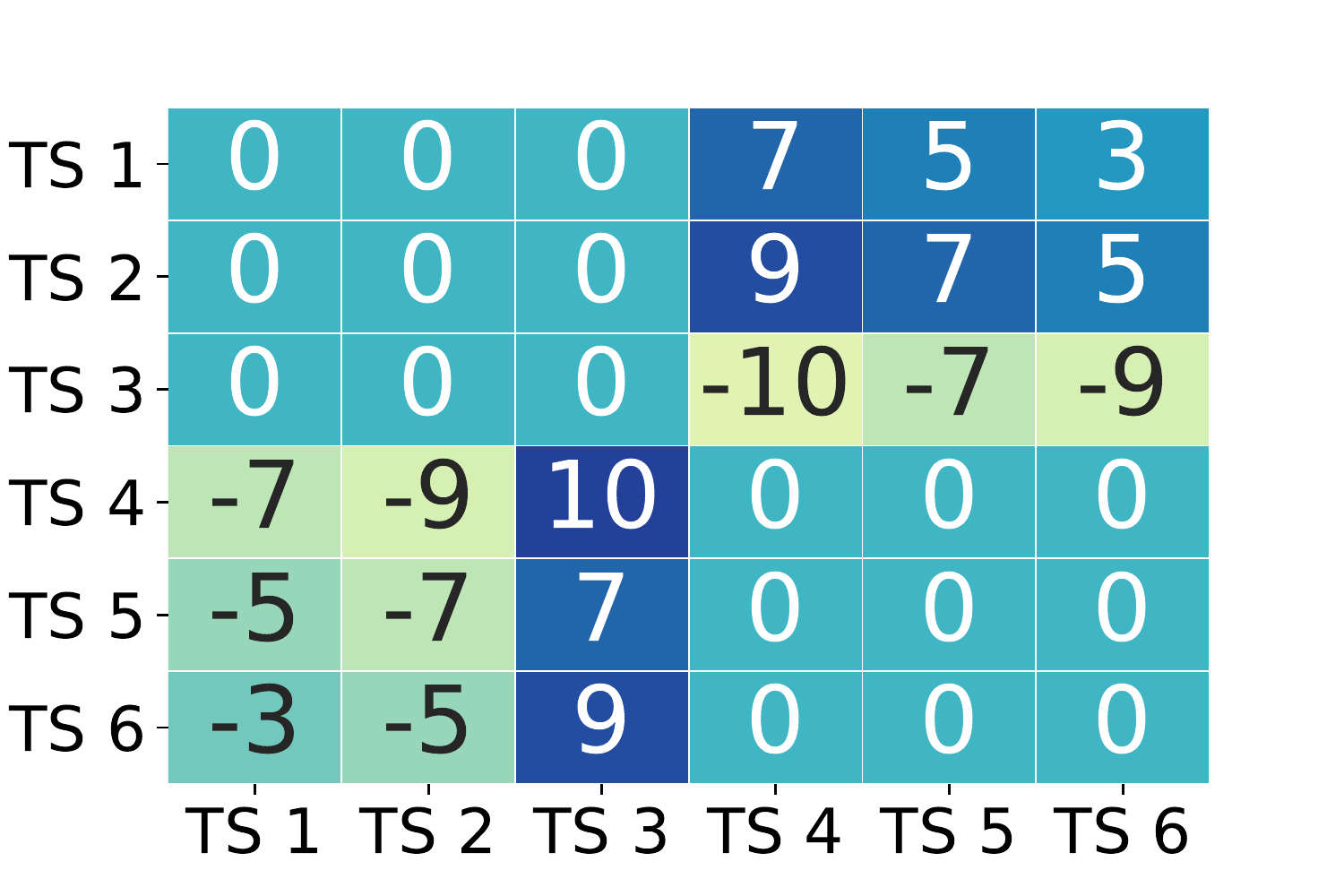}

    \includegraphics[width=\linewidth,trim=0cm 0cm 1.5cm 1cm,clip] {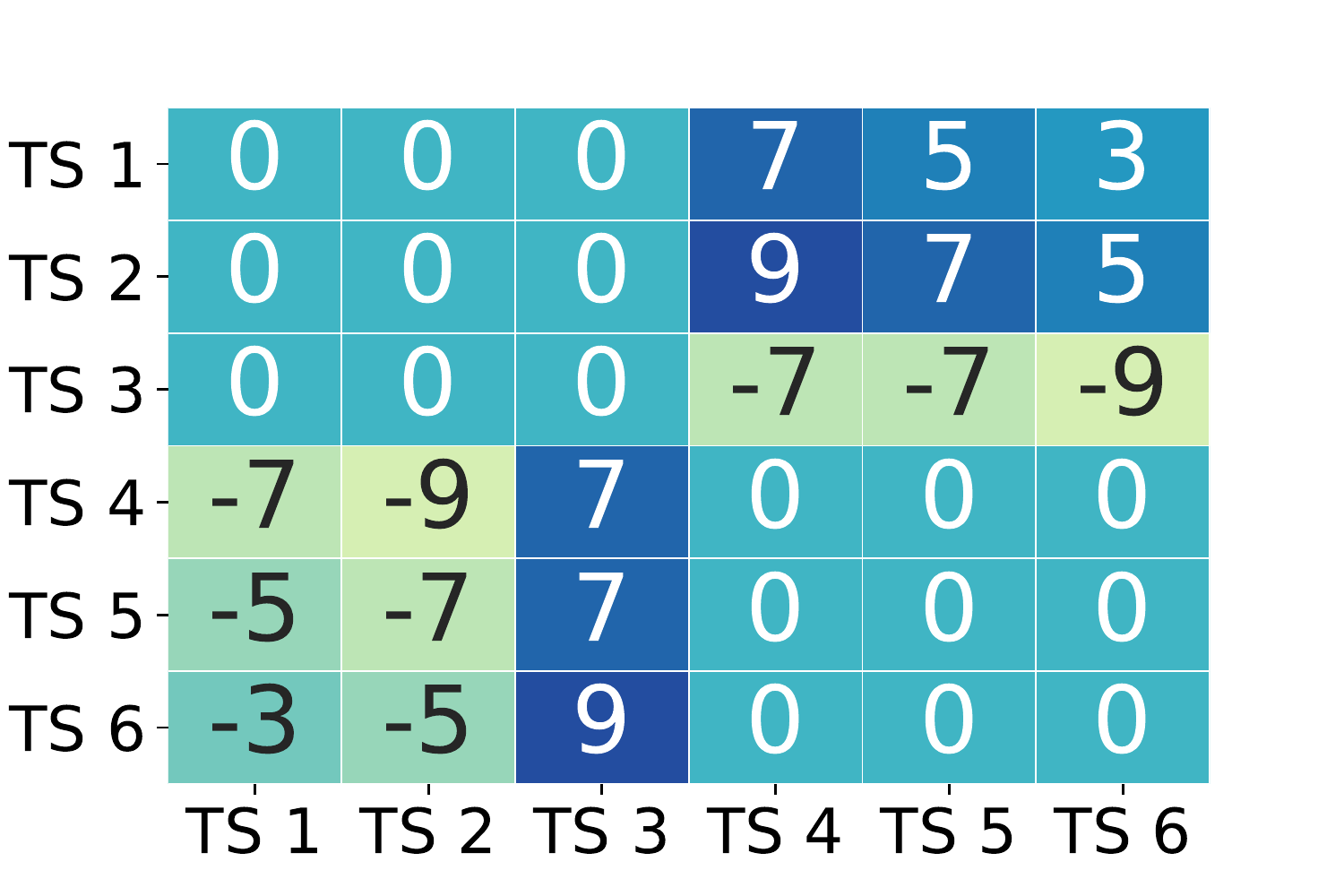}
    &
      \includegraphics[width=\linewidth,trim=0cm 0cm 1.5cm 1cm,clip]{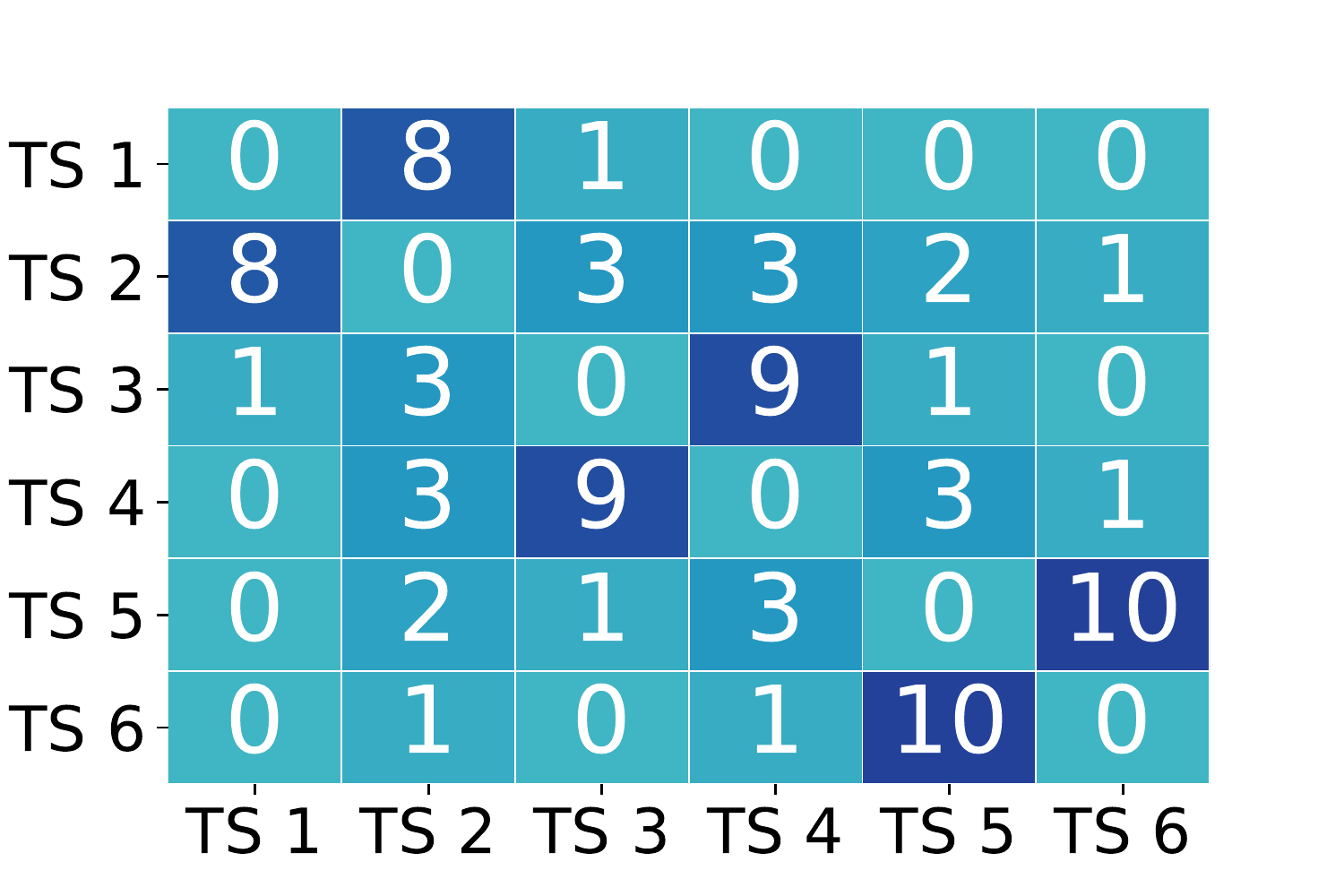}
      
      \includegraphics[width=\linewidth,trim=0cm 0cm 1.5cm 1cm,clip] {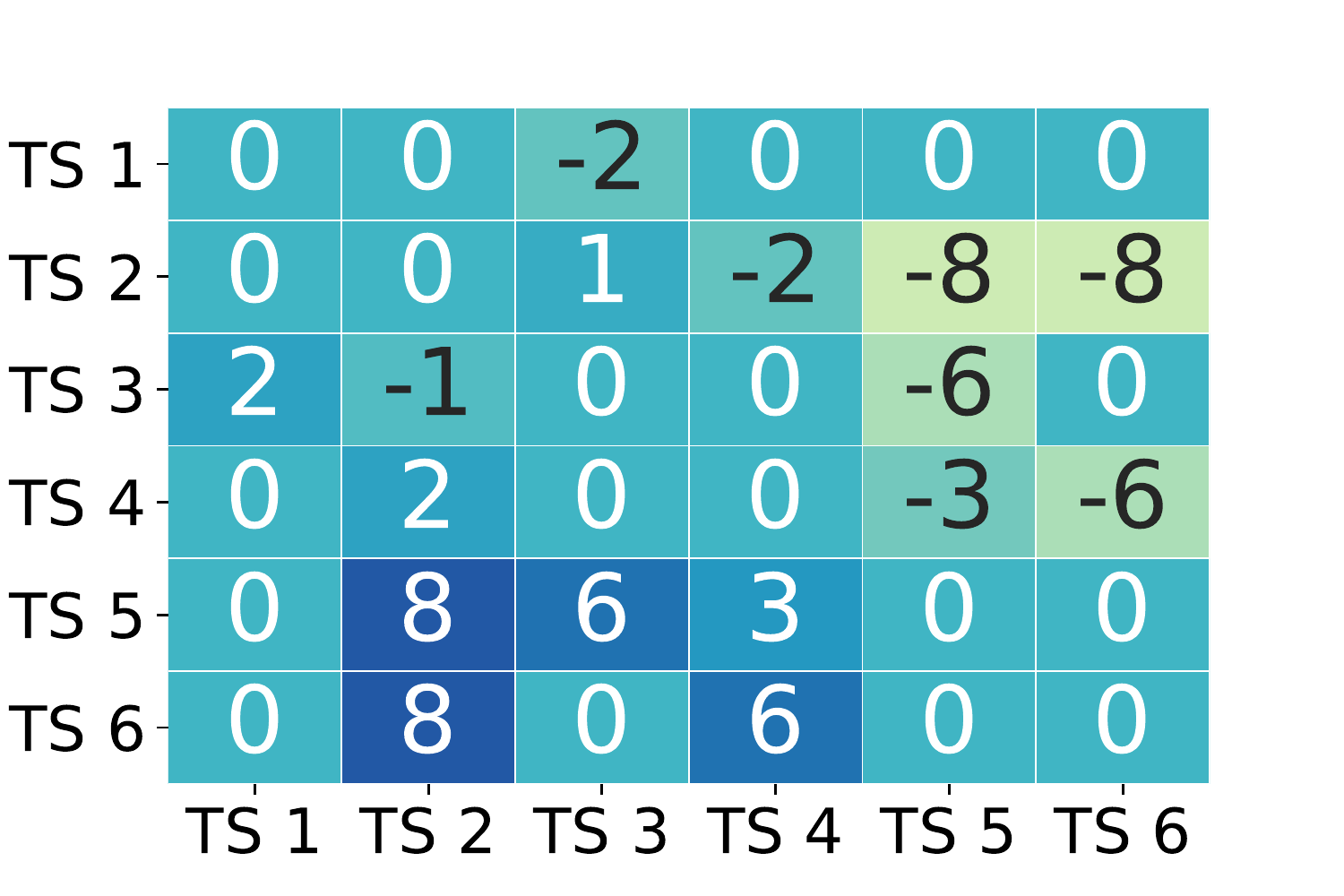}

      \includegraphics[width=\linewidth,trim=0cm 0cm 1.5cm 1cm,clip] {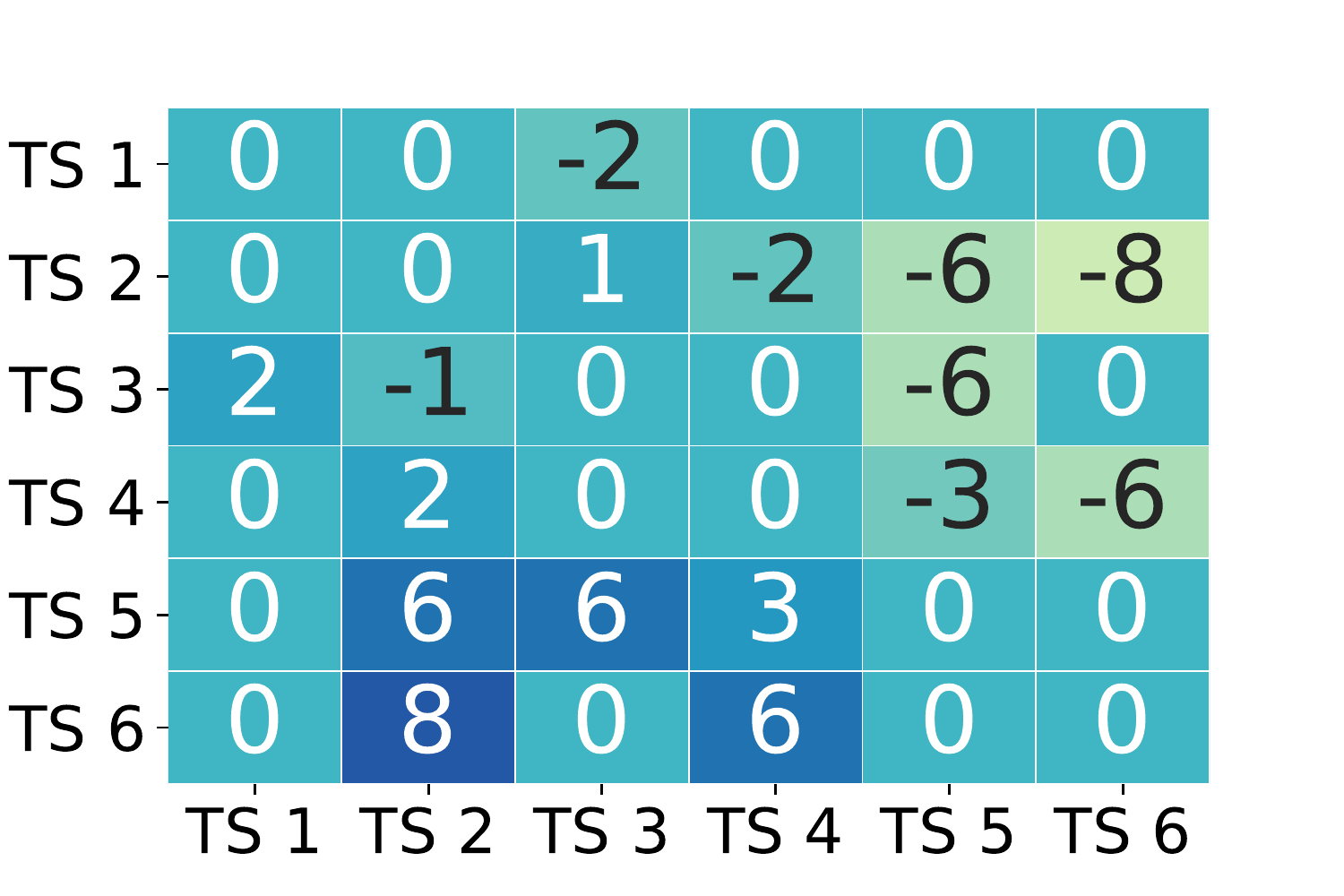}
    \\
    \multicolumn{1}{c}{$k=1$} & \multicolumn{1}{c}{$k=2$} & \multicolumn{1}{c}{$k=3$}  \\
    \end{tabular}
\captionof{figure}{Top panel: Voting matrix without voting threshold ($\theta = 1$). Middle panel: Error matrix based on mode estimation without the voting threshold ($\theta = 1$). Bottom panel: Error matrix based on median estimation without the voting threshold ($\theta = 1$).}
\label{fig:kmean_vote_matrix_before_plus_error_matrix_befores}
\end{table}

We extend our experiment by setting the voting threshold $\theta = 6$, for which results are shown in Figure \ref{fig:kmean_vote_matrix_after_plus_error_matrix_afters}. Thus, any value lower than $6$ will be replaced by $0$ in the voting matrix, and subsequently in the lead-lag matrix. We note that, for both the homogeneous and heterogeneous settings, the voting matrix consists of non-zeros across the diagonals for each block, and the other blocks consist only of $0$s. We conclude that our proposed method with a voting threshold has good performance on synthetic data, and fully recovers the lead-lag relationship in both the homogeneous and heterogeneous settings.

\begin{table}[htbp]
  \centering
    \begin{tabular}{p{4.5cm}|p{4.5cm}p{4.5cm}}
      \multicolumn{3}{c}{\textbf{K-means++ clustering}} \vspace{0.3cm}\\
      
      \multicolumn{1}{c}{\textbf{Homogeneous Setting}}    &  \multicolumn{2}{c}{\textbf{Heterogeneous Setting}}  \\

      \includegraphics[width=\linewidth,trim=0cm 0cm 1.5cm 1cm,clip]{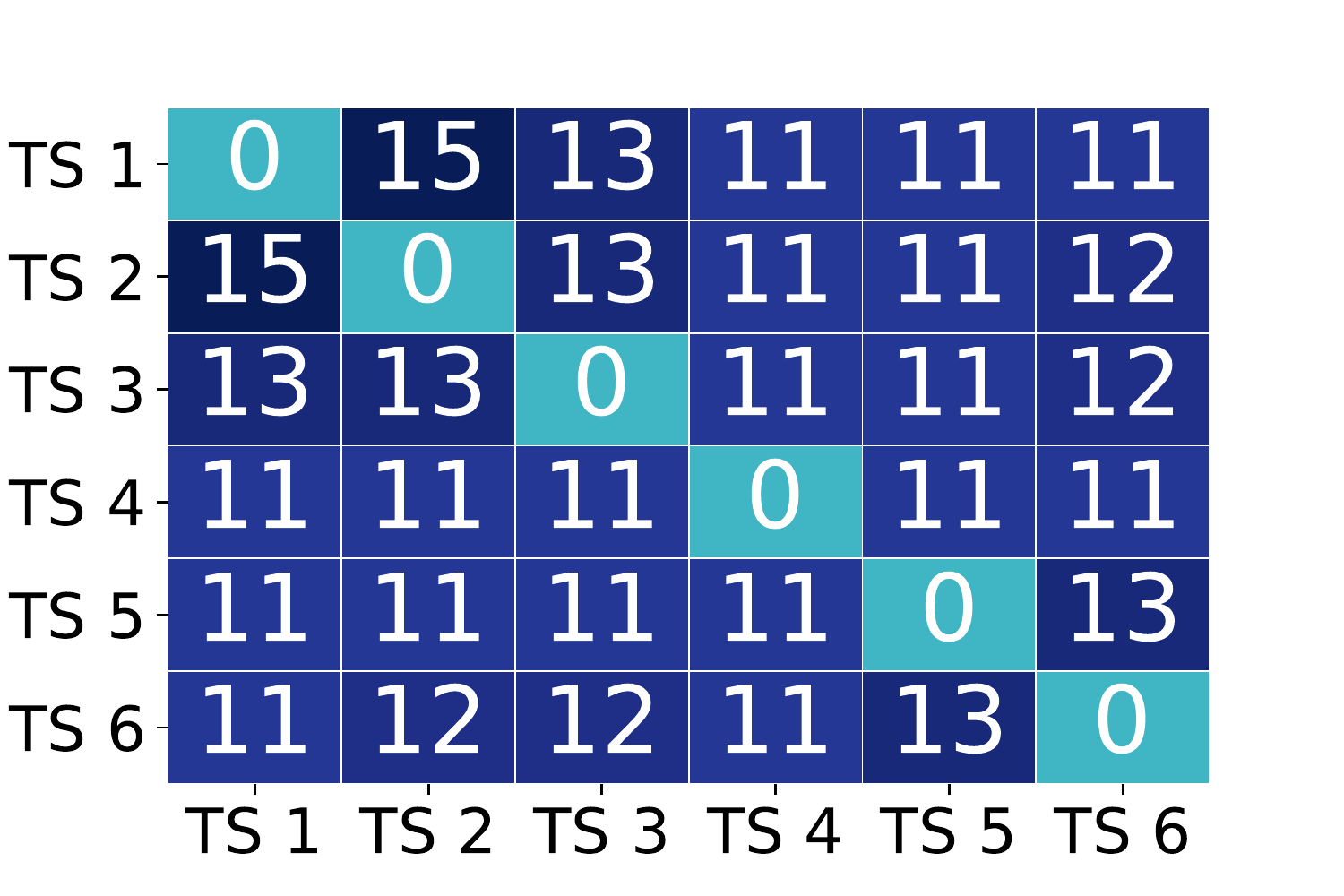}
      
      \includegraphics[width=\linewidth,trim=0cm 0cm 1.5cm 1cm,clip] {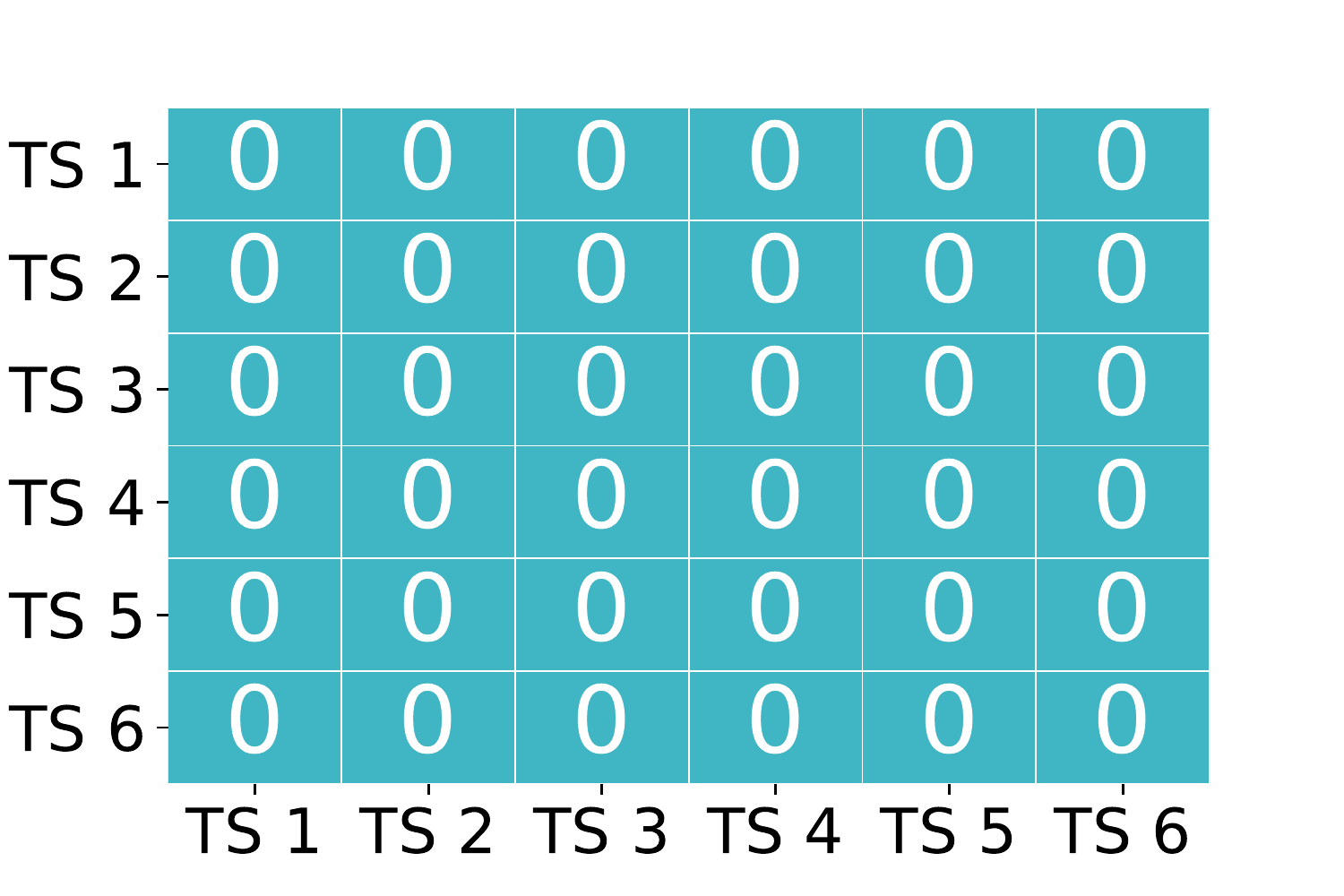}

    \includegraphics[width=\linewidth,trim=0cm 0cm 1.5cm 1cm,clip] {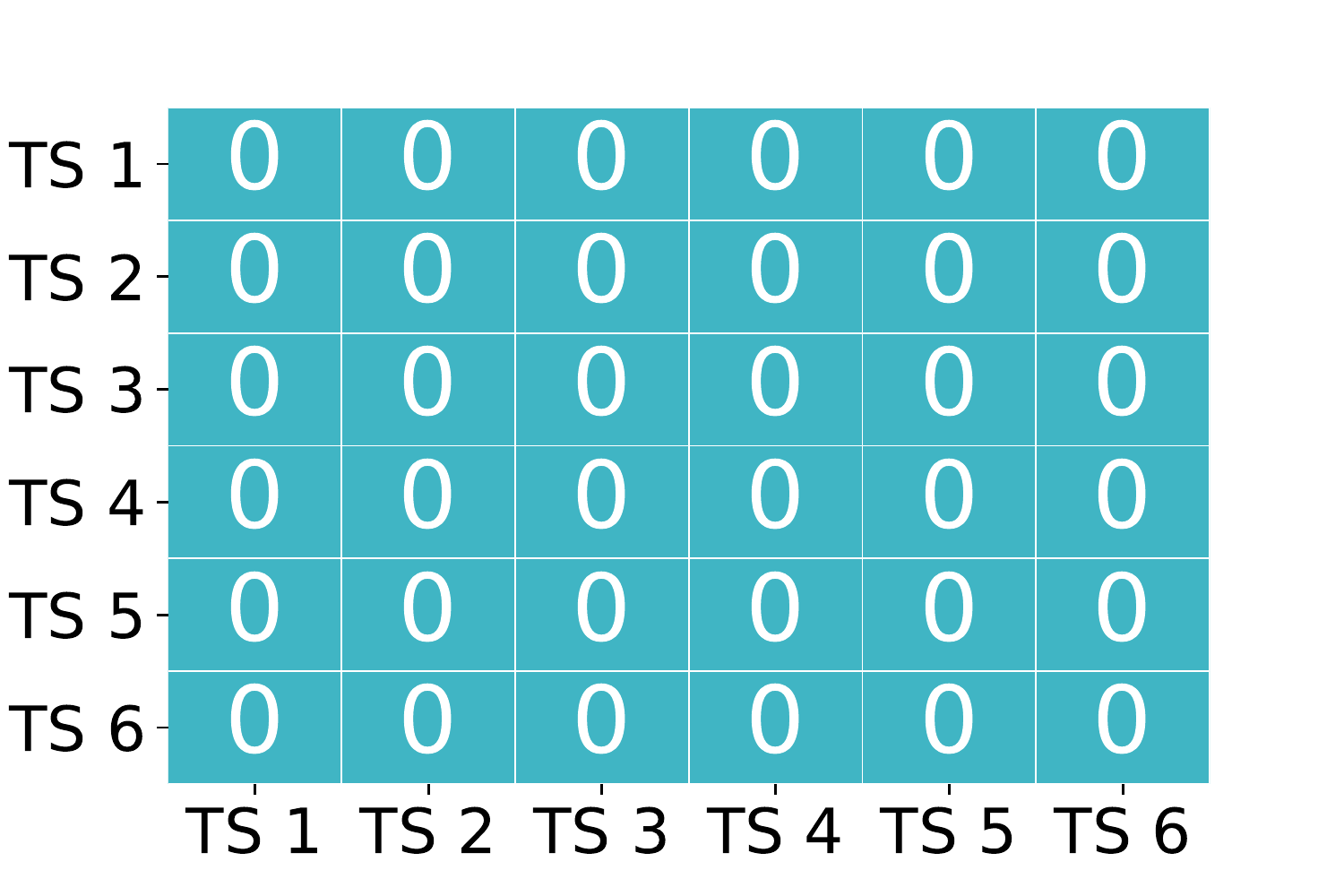}

    &
      \includegraphics[width=\linewidth,trim=0cm 0cm 1.5cm 1cm,clip]{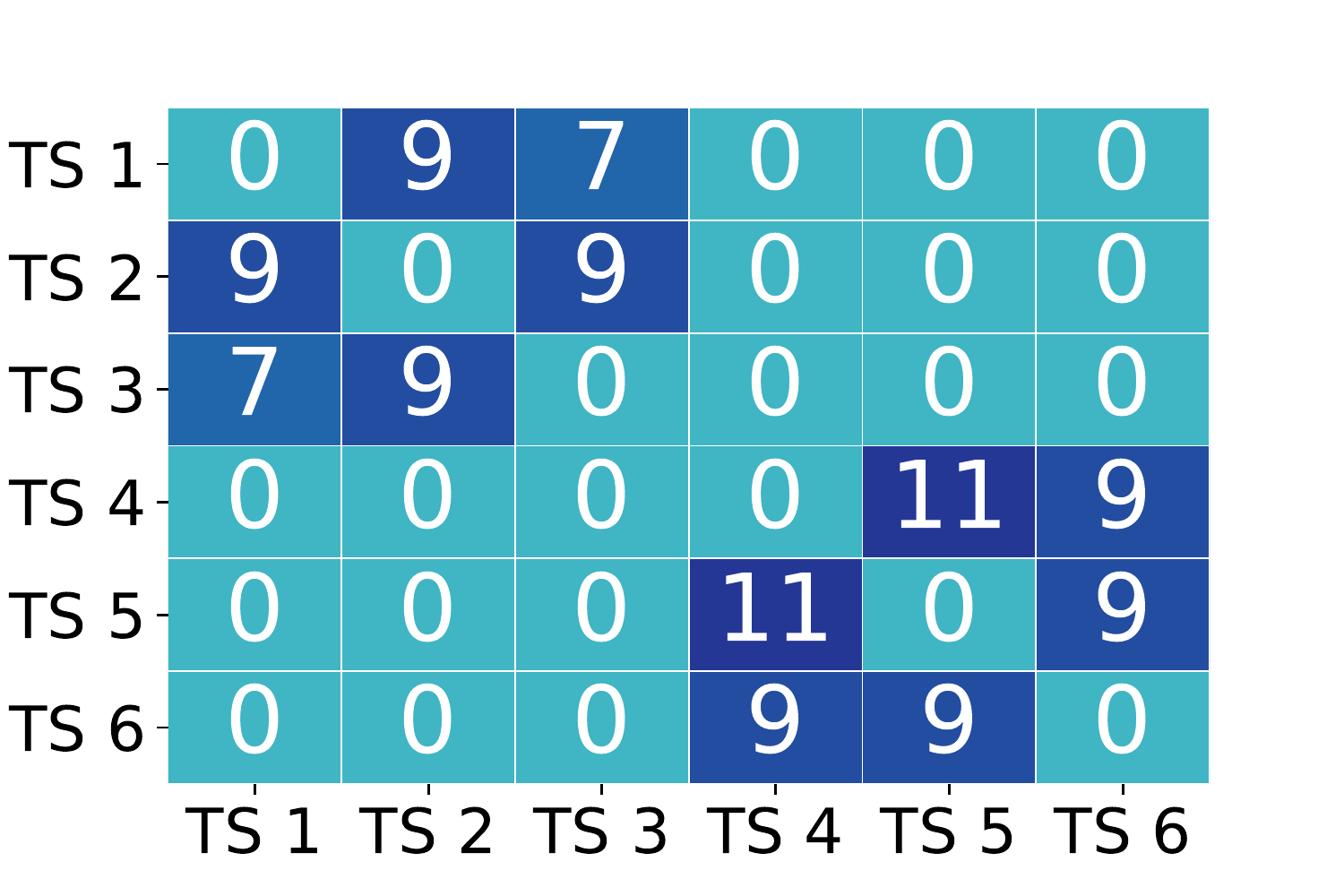}
      
      \includegraphics[width=\linewidth,trim=0cm 0cm 1.5cm 1cm,clip] {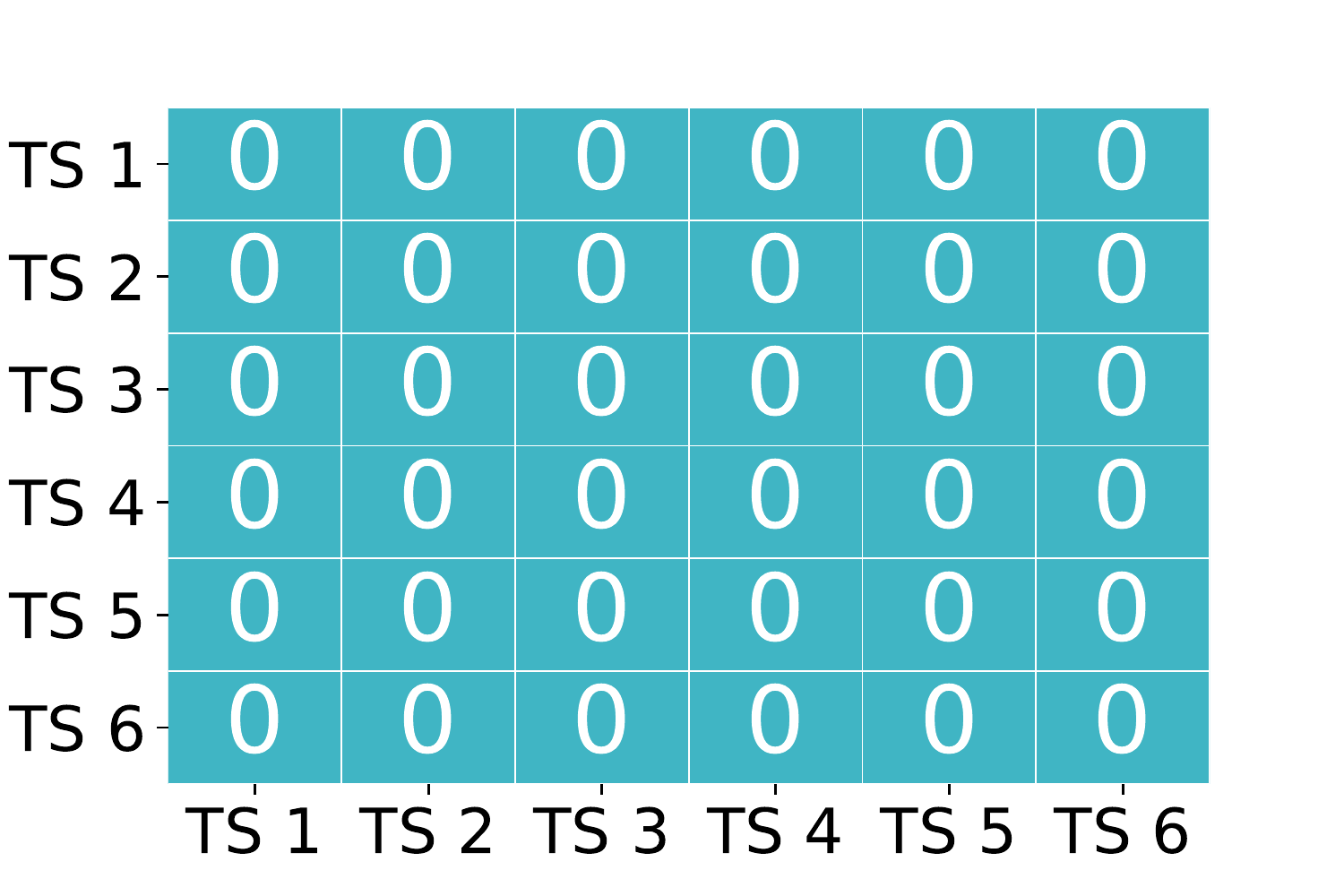}

    \includegraphics[width=\linewidth,trim=0cm 0cm 1.5cm 1cm,clip] {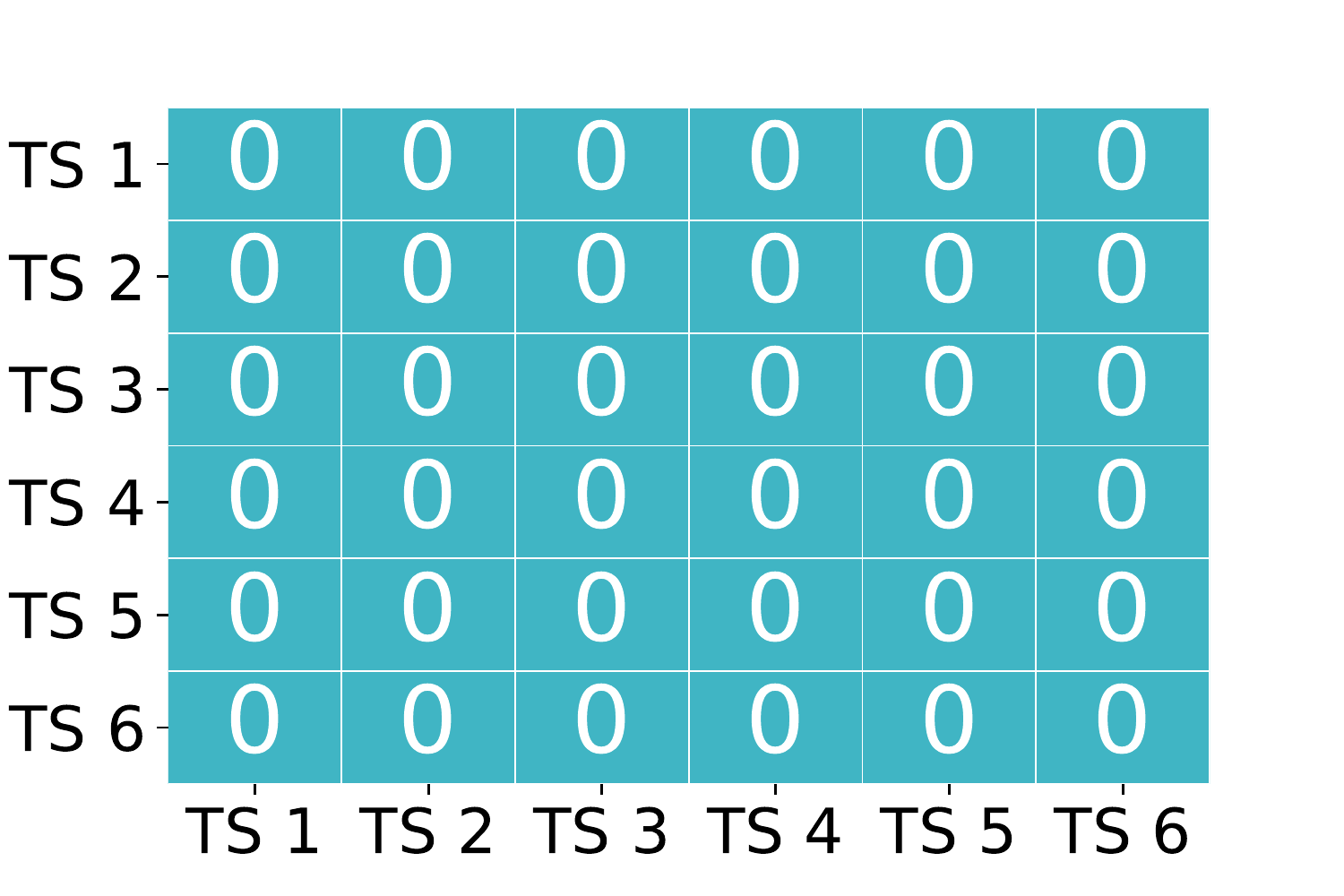}
    &
      \includegraphics[width=\linewidth,trim=0cm 0cm 1.5cm 1cm,clip]{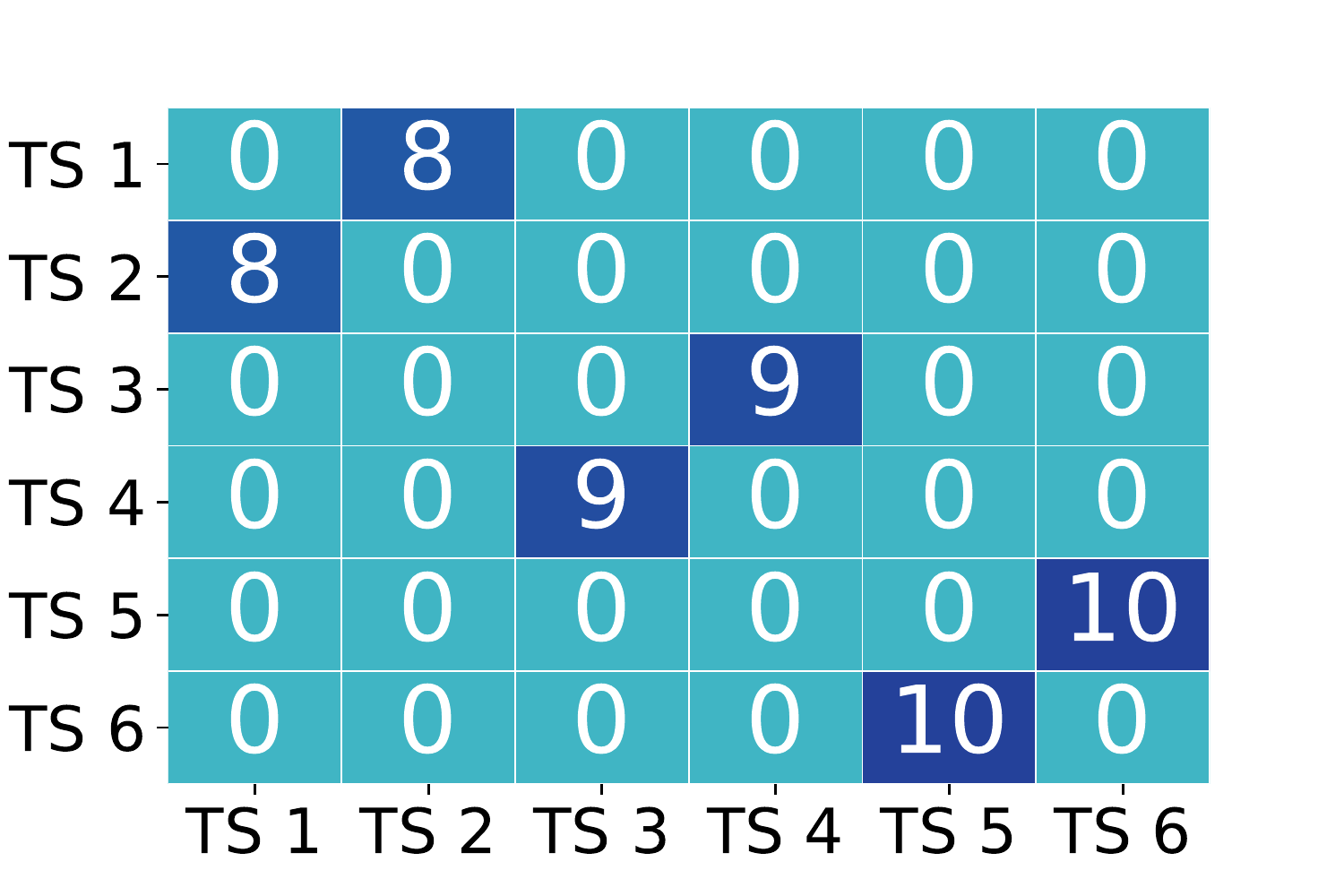}
      
      \includegraphics[width=\linewidth,trim=0cm 0cm 1.5cm 1cm,clip] {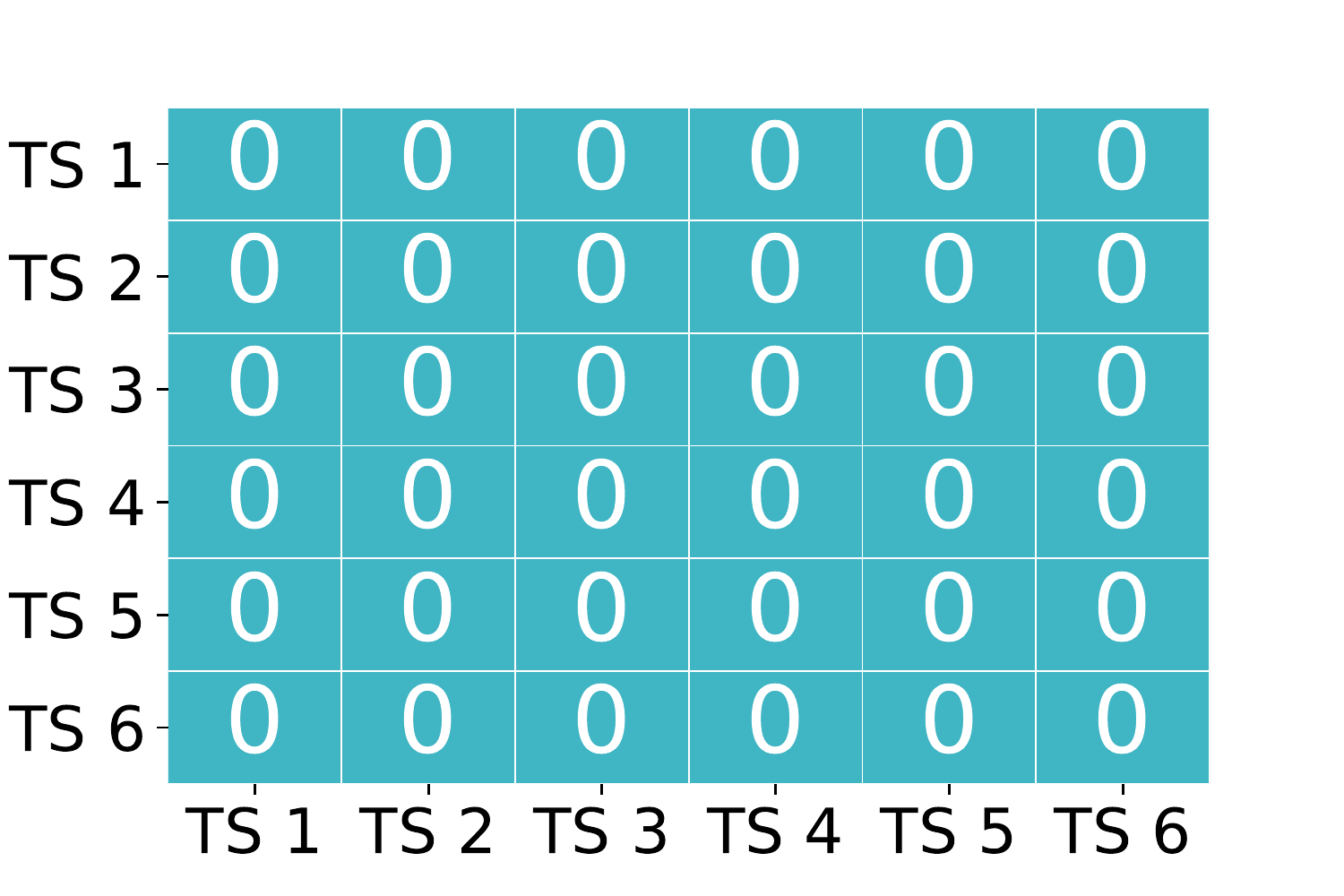}

        \includegraphics[width=\linewidth,trim=0cm 0cm 1.5cm 1cm,clip] {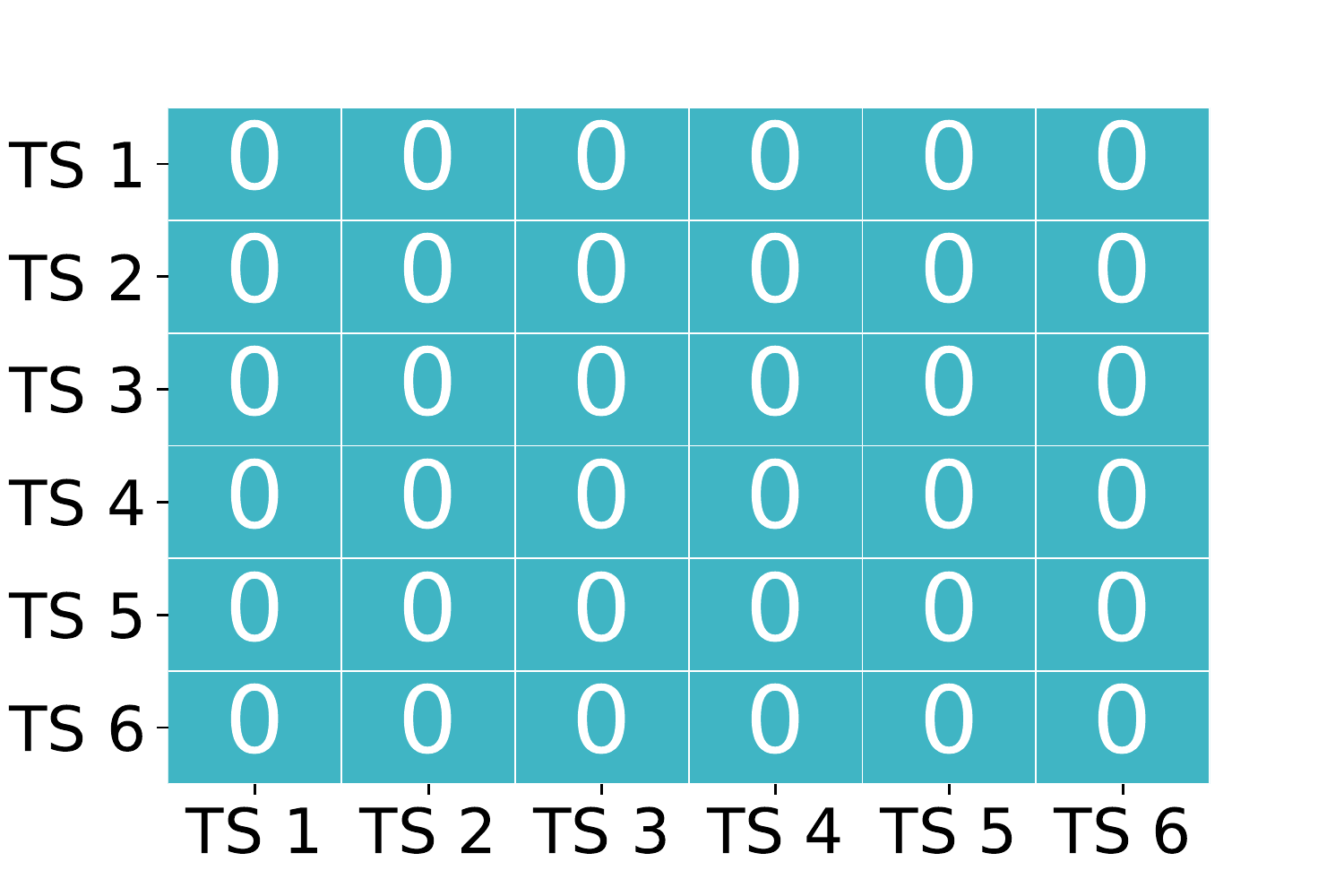}
    \\
    \multicolumn{1}{c}{$k=1$} & \multicolumn{1}{c}{$k=2$} & \multicolumn{1}{c}{$k=3$}  \\
    \end{tabular}
\captionof{figure}{Top panel: Voting matrix with voting threshold ($\theta = 6$). Middle panel: Error matrix based on mode estimation with the voting threshold ($\theta = 6$). Bottom panel: Error matrix based on median estimation with the voting threshold ($\theta = 6$).}
\label{fig:kmean_vote_matrix_after_plus_error_matrix_afters}
\end{table}

\newpage
\subsection{Simulation}


In the next set of synthetic experiments, we increase $n$ to $60$. The $B$ and $L$ matrices follow an analogous pattern to the $n = 6$ case. The Adjusted Rand Index (ARI) calculates a similarity metric between two clusterings by examining every sample pair and counting the number of pairs assigned to the same or different clusters in both the predicted and actual clusterings. The ARI is calculated by adjusting the Rand Index for chance agreement, and it ranges from -1 to 1, where a score of 1 indicates perfect agreement between the two clusterings, a score of 0 indicates random agreement, and a score less than 0 indicates disagreement between the two clusterings. As observed in Figure \ref{tab:ari_sigma}, it is evident that for both the homogeneous and heterogeneous settings, when the $\sigma$ ranges from $0$ to $1.5$, the KM clustering achieves a stable ARI around $0.8$, whereas the SP clustering fluctuates slightly in the earlier $\sigma$ values. However, when the $\sigma$ value exceeds $1.5$, the SP clustering gains a slightly higher performance than the KM clustering, and the ARI for both methods drops significantly.


\begin{table}[htbp]
  \centering
    \begin{tabular}{p{4.5cm}|p{4.5cm}p{4.5cm}}
      \multicolumn{1}{c}{\textbf{Homogeneous Setting}}    &  \multicolumn{2}{c}{\textbf{Heterogeneous Setting}}  \\

      \includegraphics[width=\linewidth,trim=0cm 0cm 0cm 0cm,clip]{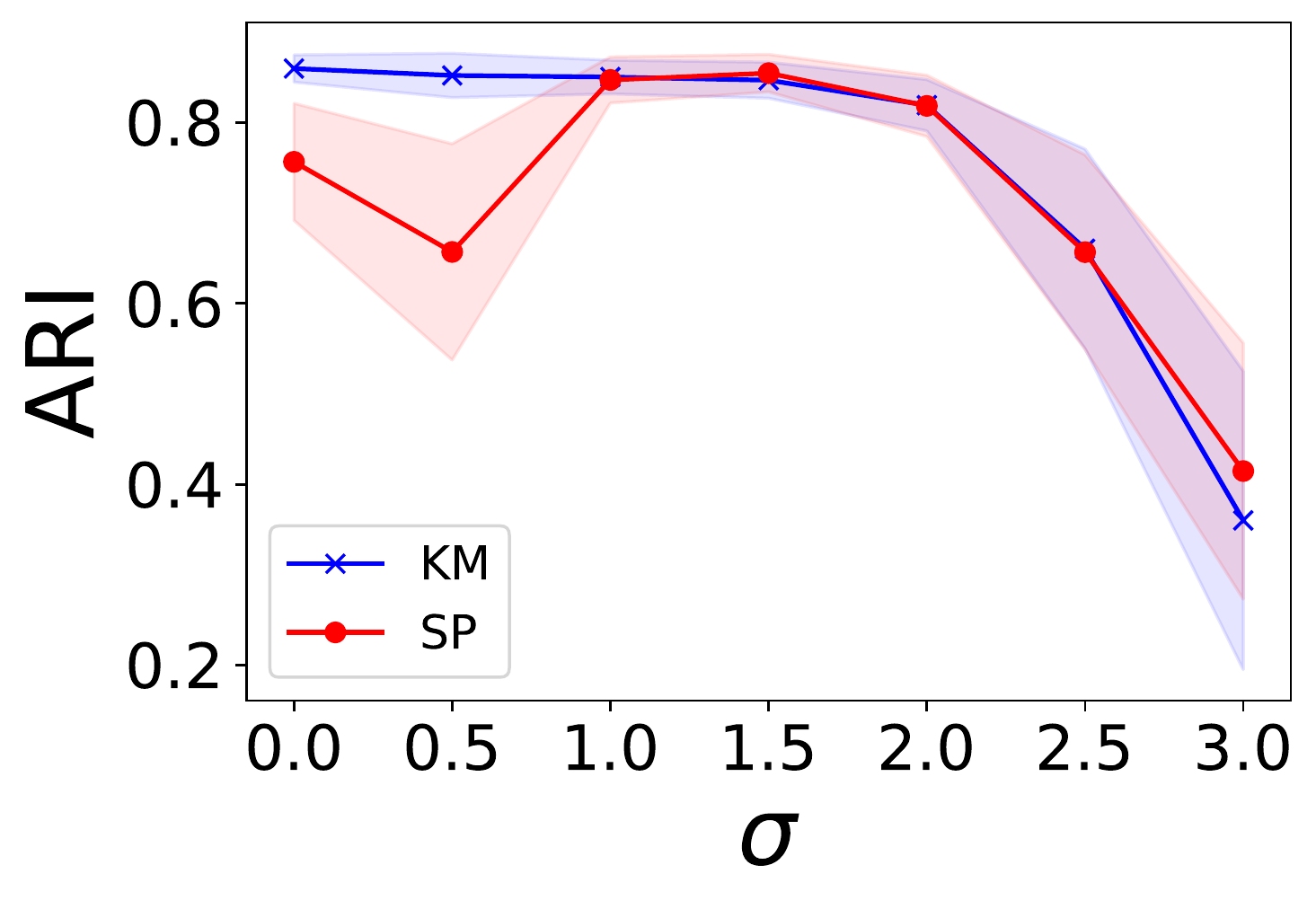}

    &
      \includegraphics[width=\linewidth,trim=0cm 0cm 0cm 0cm,clip]{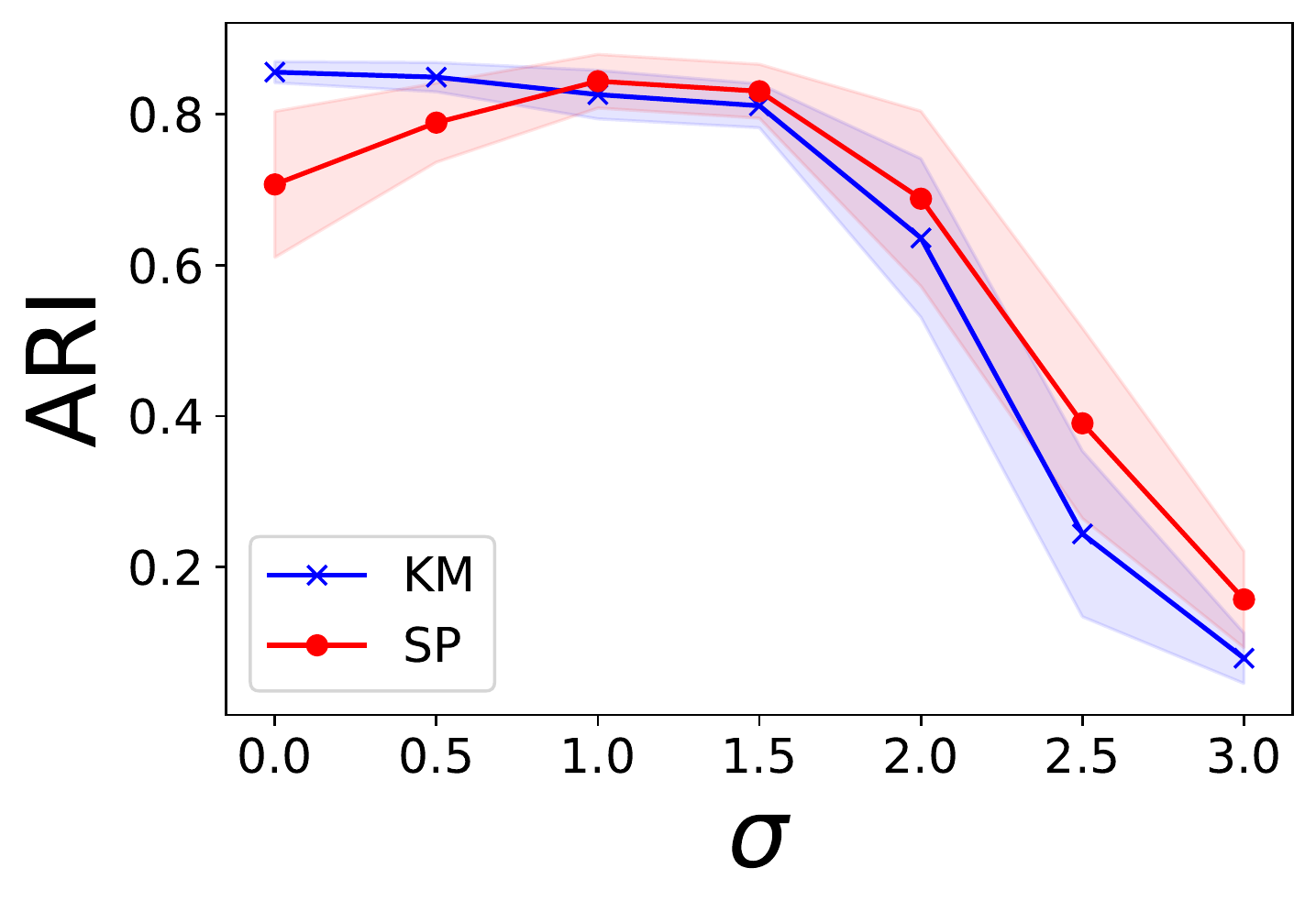}

    &
      \includegraphics[width=\linewidth,trim=0cm 0cm 0cm 0cm,clip]{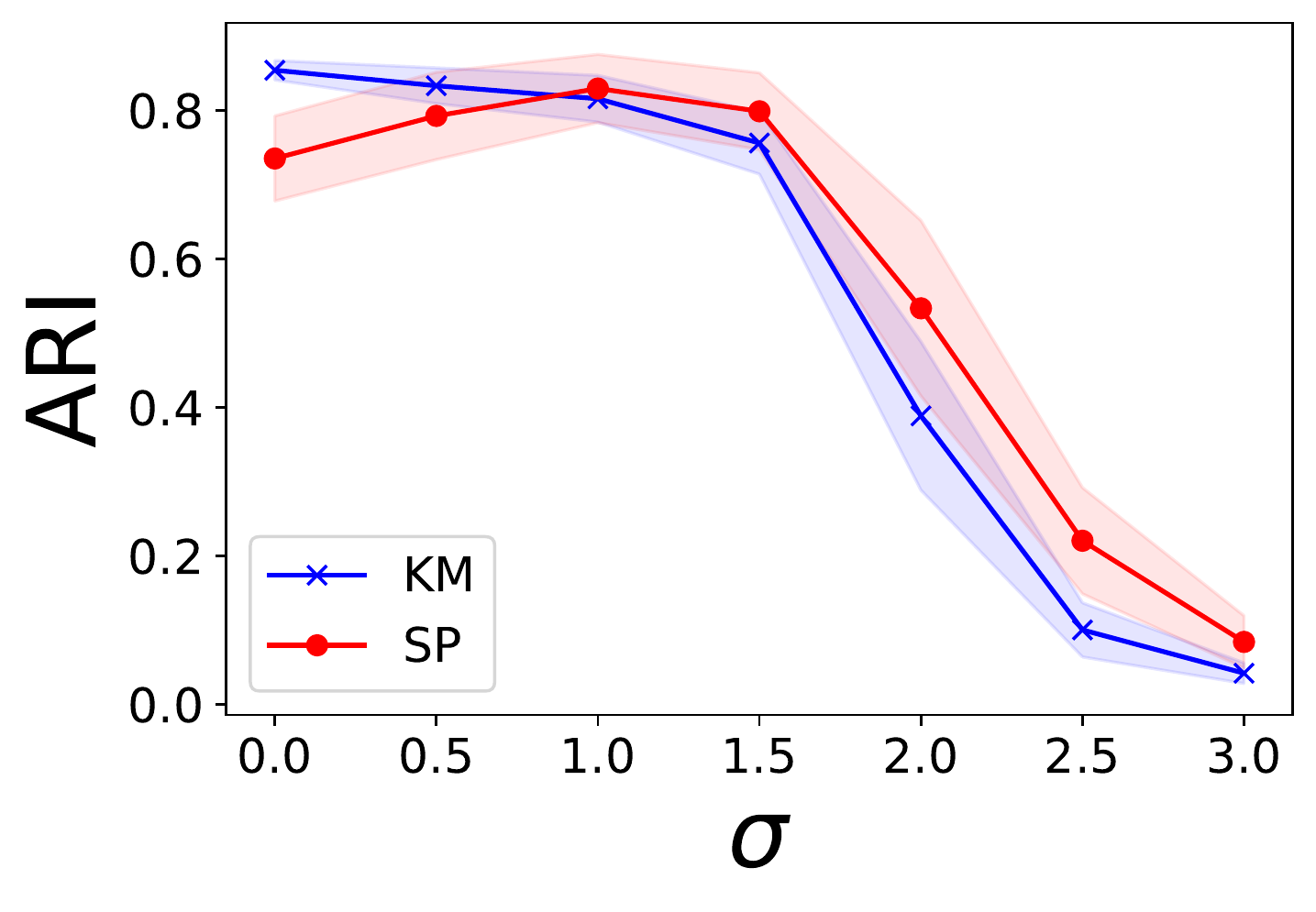}
    \\
    \multicolumn{1}{c}{$k=1$} & \multicolumn{1}{c}{$k=2$} & \multicolumn{1}{c}{$k=3$}  \\
    \end{tabular}
\captionof{figure}{Average and confidence interval for the ARI with different levels of $\sigma$ based on 60 time series and 100 simulations for every iteration.}
\label{tab:ari_sigma}
\end{table}

Figure \ref{tab:mse_vs_sigma} shows the average and confidence interval for the MSE when estimating the true lag with different $\sigma$ levels based on $100$ simulations without voting threshold ($\theta = 1$, top panel) and with voting threshold ($\theta = 6$, bottom panel). 

In the homogeneous setting ($k=1$), both plots follow the same trend. Specifically, for $\sigma$ ranging from $0$ to $1.5$, the MSE of the K-means++ mode (KM\_Mod) and spectral mode (SP\_Mod) is near $0$. In comparison, the MSE of K-means++ median (KM\_Med) and spectral median (SP\_Med) is higher. When $\sigma$ ranges from $1.5$ to $3.0$, all four methods’ MSE increases significantly due to the higher noise level. 

In the heterogeneous setting ($k \in \{2,3\}$), we note that for methods not using the voting mechanism, the MSE is significantly higher, having a value of around $12$. However, once the voting mechanism is added, the MSE becomes significantly lower. When the $\sigma$ values range between $0$ to $1.5$, the MSE of KM clustering goes down to near $0$, whereas SP clustering have a slightly higher MSE. After the value of $\sigma$ reaches $1.5$ and beyond, the MSE increases dramatically for all four methods.

\begin{table}[htbp]
  \centering
    \begin{tabular}{p{4.5cm}|p{4.5cm}p{4.5cm}}
      \multicolumn{1}{c}{\textbf{Homogeneous Setting}}    &  \multicolumn{2}{c}{\textbf{Heterogeneous Setting}}  \\

      \includegraphics[width=\linewidth,trim=0cm 0cm 0cm 0cm,clip]{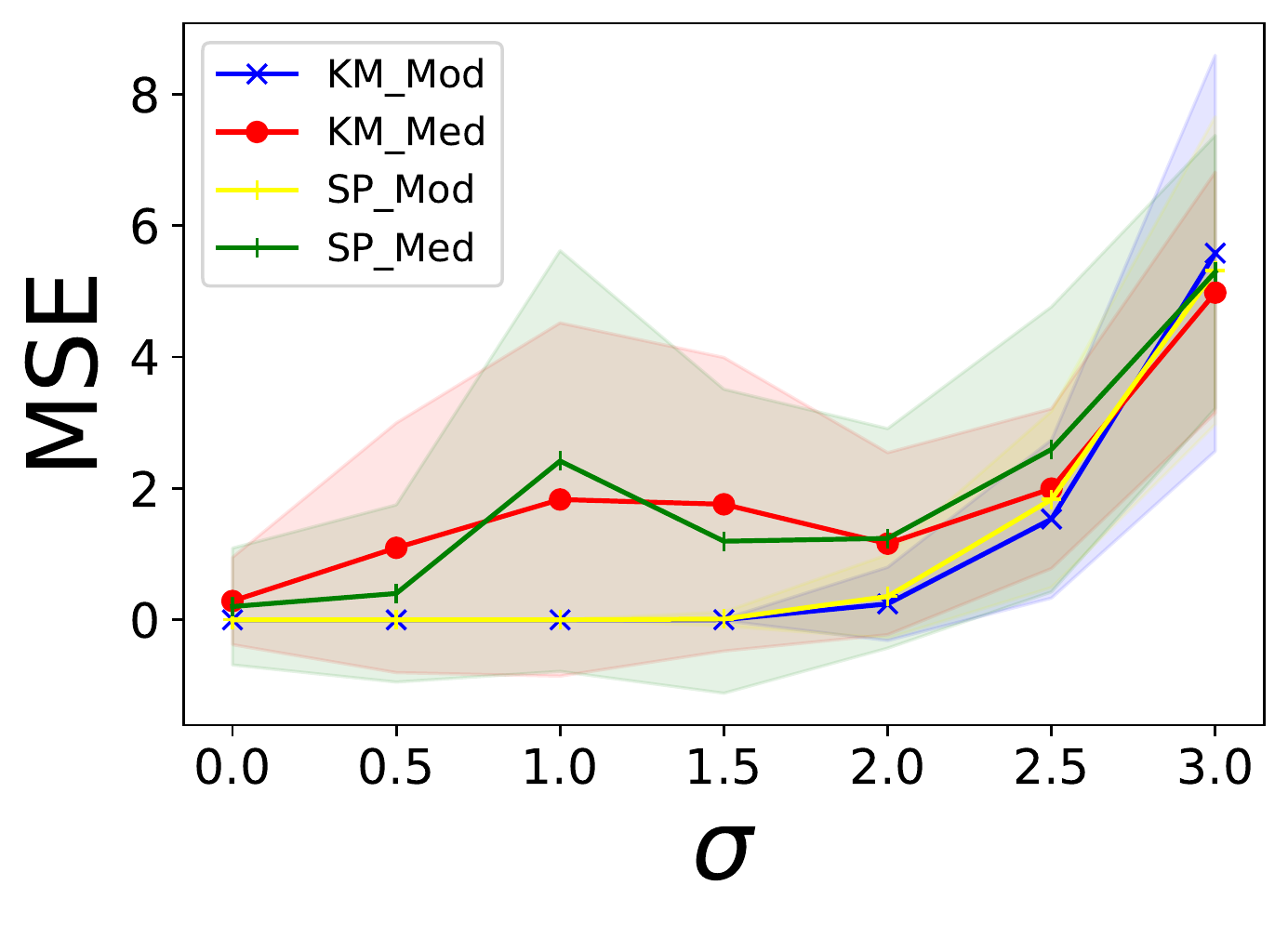}
      
      \includegraphics[width=\linewidth,trim=0cm 0cm 0cm 0cm,clip]{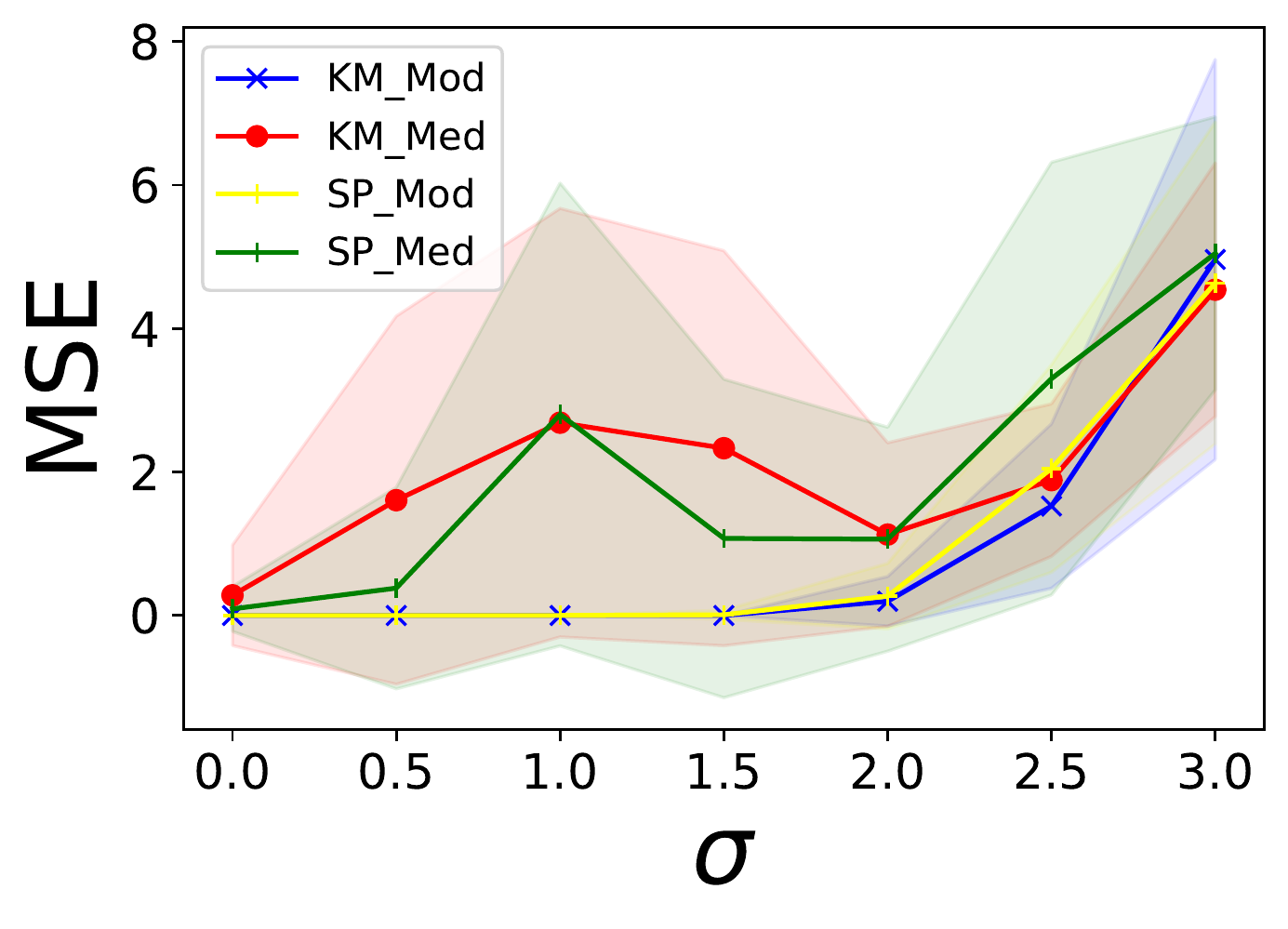}

    &
     \includegraphics[width=\linewidth,trim=0cm 0cm 0cm 0cm,clip]{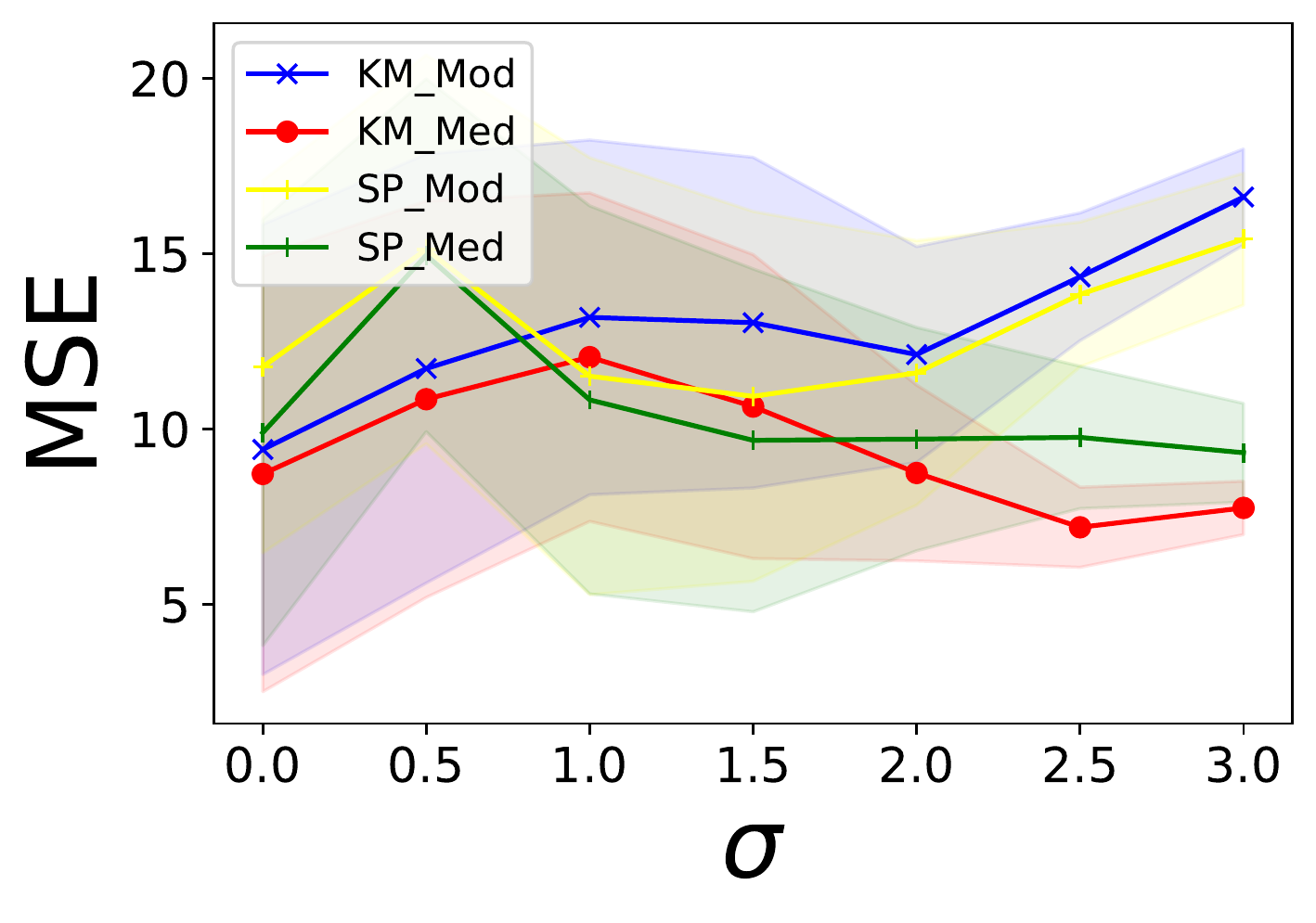}
      
    \includegraphics[width=\linewidth,trim=0cm 0cm 0cm 0cm,clip]{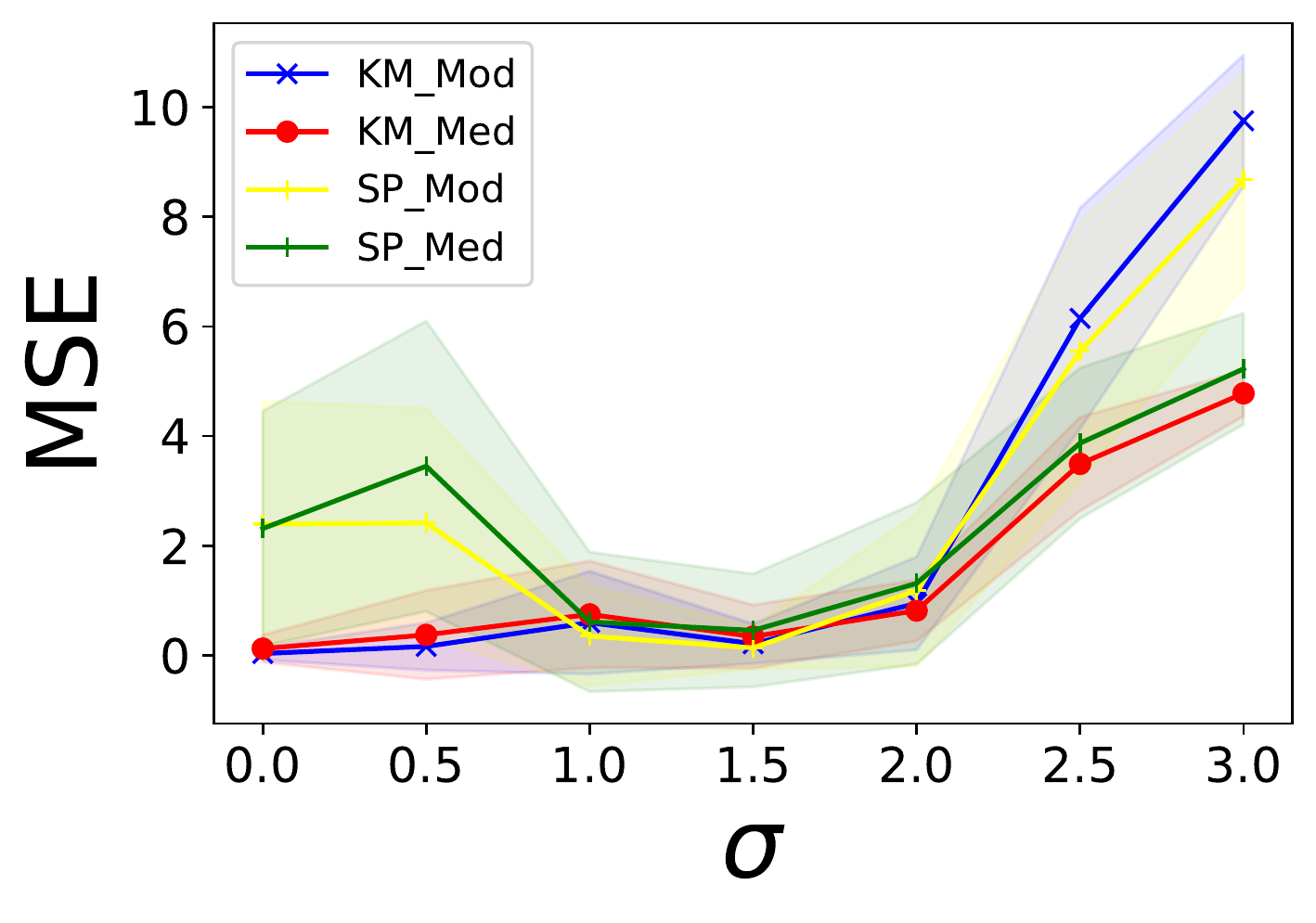}

    &
      \includegraphics[width=\linewidth,trim=0cm 0cm 0cm 0cm,clip]{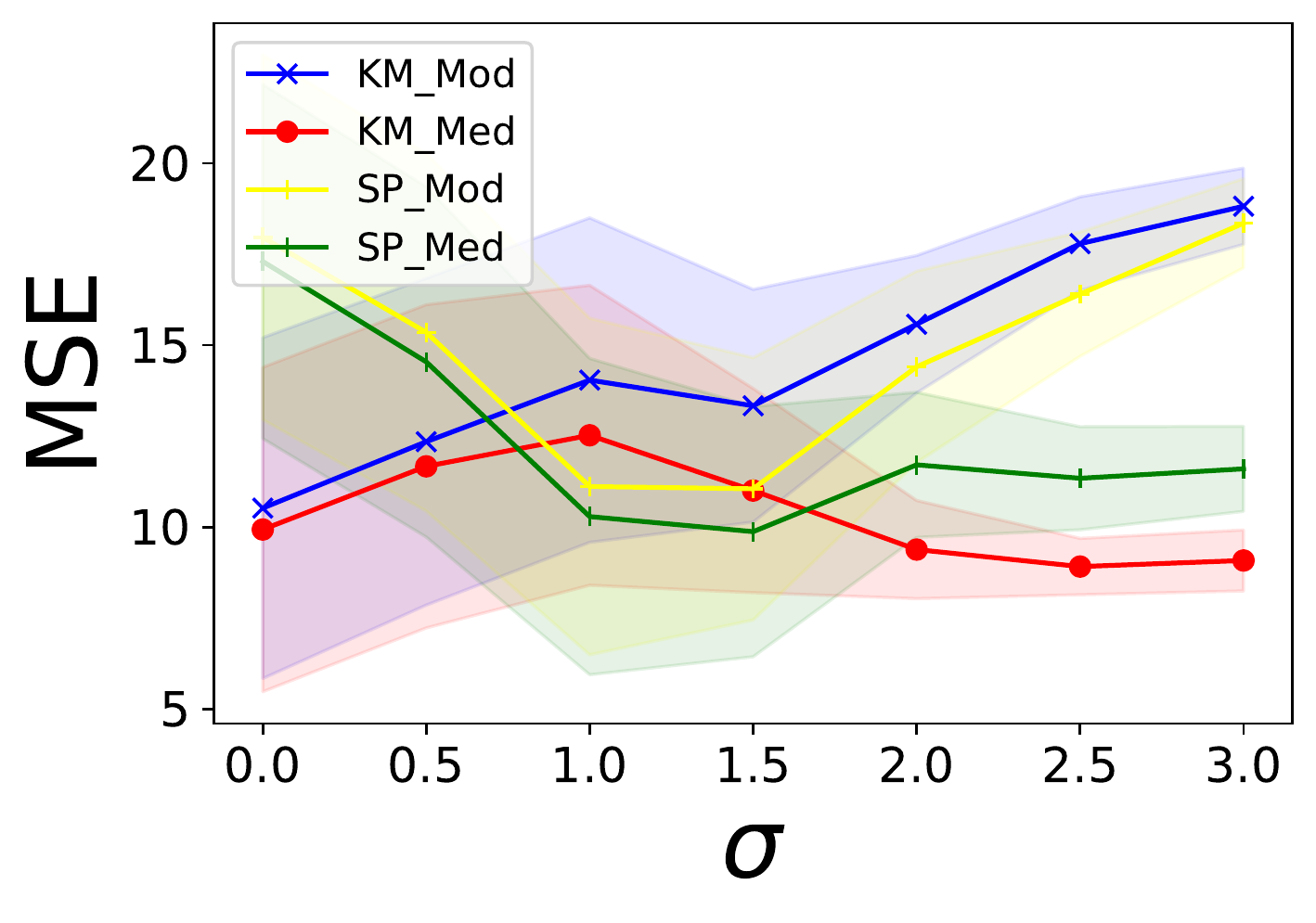}
      
      \includegraphics[width=\linewidth,trim=0cm 0cm 0cm 0cm,clip]{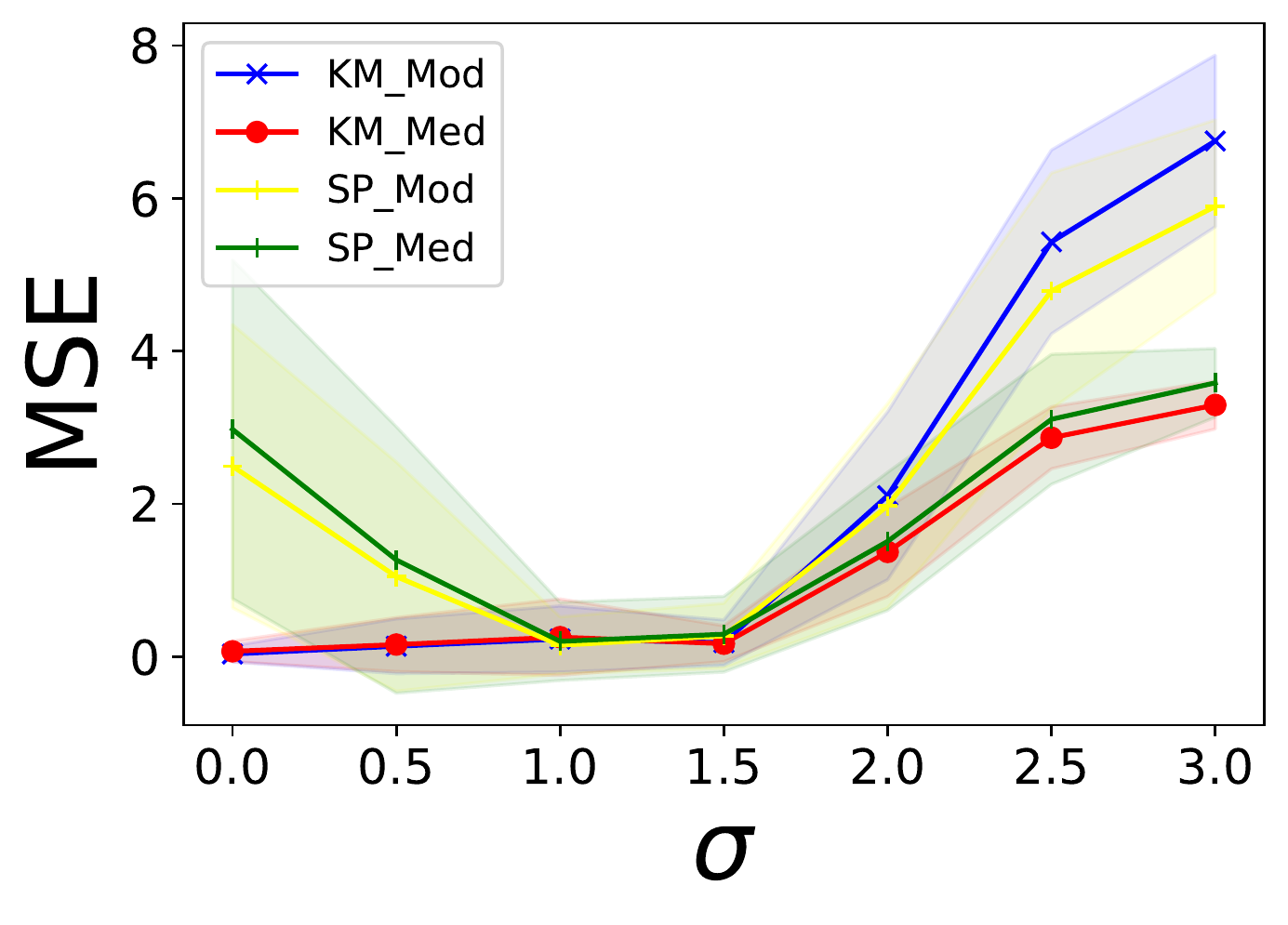}
    \\
    \multicolumn{1}{c}{$k=1$} & \multicolumn{1}{c}{$k=2$} & \multicolumn{1}{c}{$k=3$}  \\
    \end{tabular}
\captionof{figure}{Top panel: Average and confidence interval for the MSE with different levels of $\sigma$ based on $100$ simulations for every iteration without voting threshold ($\theta = 1$). Bottom panel: Average and confidence interval for the MSE with different $\sigma$ levels based on $100$ simulations for every iteration with voting threshold ($\theta = 6$).}
\label{tab:mse_vs_sigma}
\end{table}

We next study the sensitivity of our proposed method to the value of the voting threshold $\theta$. In Figure \ref{tab:mse_vs_threshold}, we first observe that in the homogeneous setting ($k = 1$), the four methods maintain a constant MSE as the voting threshold increases. This aligns with our expectations. Moreover, we also notice that mode estimation performs better than median estimation. This is evident throughout the plots in Figure \ref{tab:ari_sigma} and \ref{tab:mse_vs_sigma}. Thus, the MSE for median estimation lies between $2$ and $3$, whereas the MSE for the mode estimation drops to near $0$. For the heterogeneous setting ($k \in \{2, 3\}$), we note that as the voting threshold increases, the MSE decreases to near 0 for all methods and remains low thereafter. This altogether suggests that the proposed method is robust to the choice of the voting threshold $\theta$.

\begin{table}[htbp]
  \centering
    \begin{tabular}{p{4.5cm}|p{4.5cm}p{4.5cm}}
      \multicolumn{1}{c}{\textbf{Homogeneous Setting}}    &  \multicolumn{2}{c}{\textbf{Heterogeneous Setting}}  \\

      \includegraphics[width=\linewidth,trim=0cm 0cm 0cm 0cm,clip]{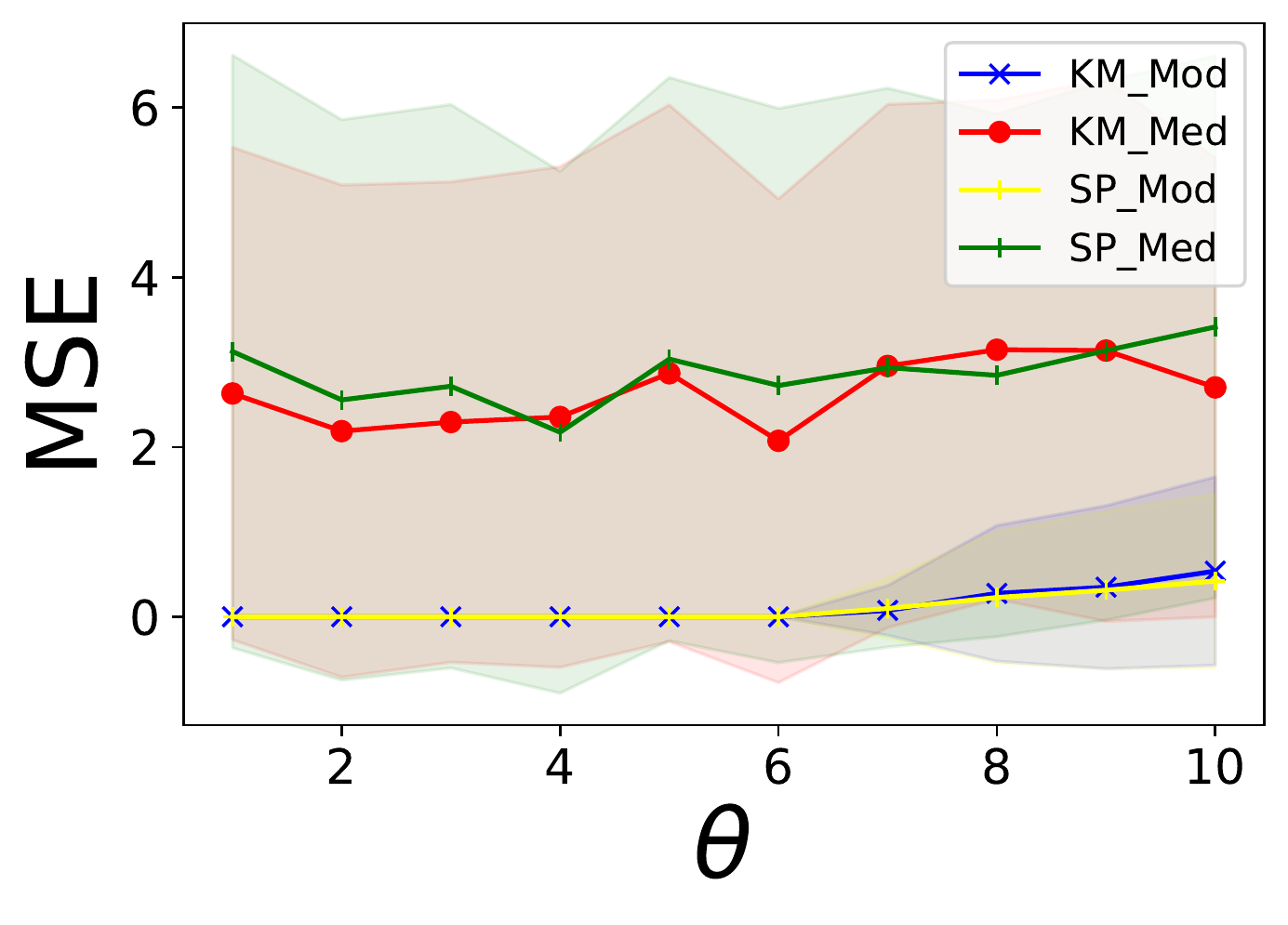}

    &
      \includegraphics[width=\linewidth,trim=0cm 0cm 0cm 0cm,clip]{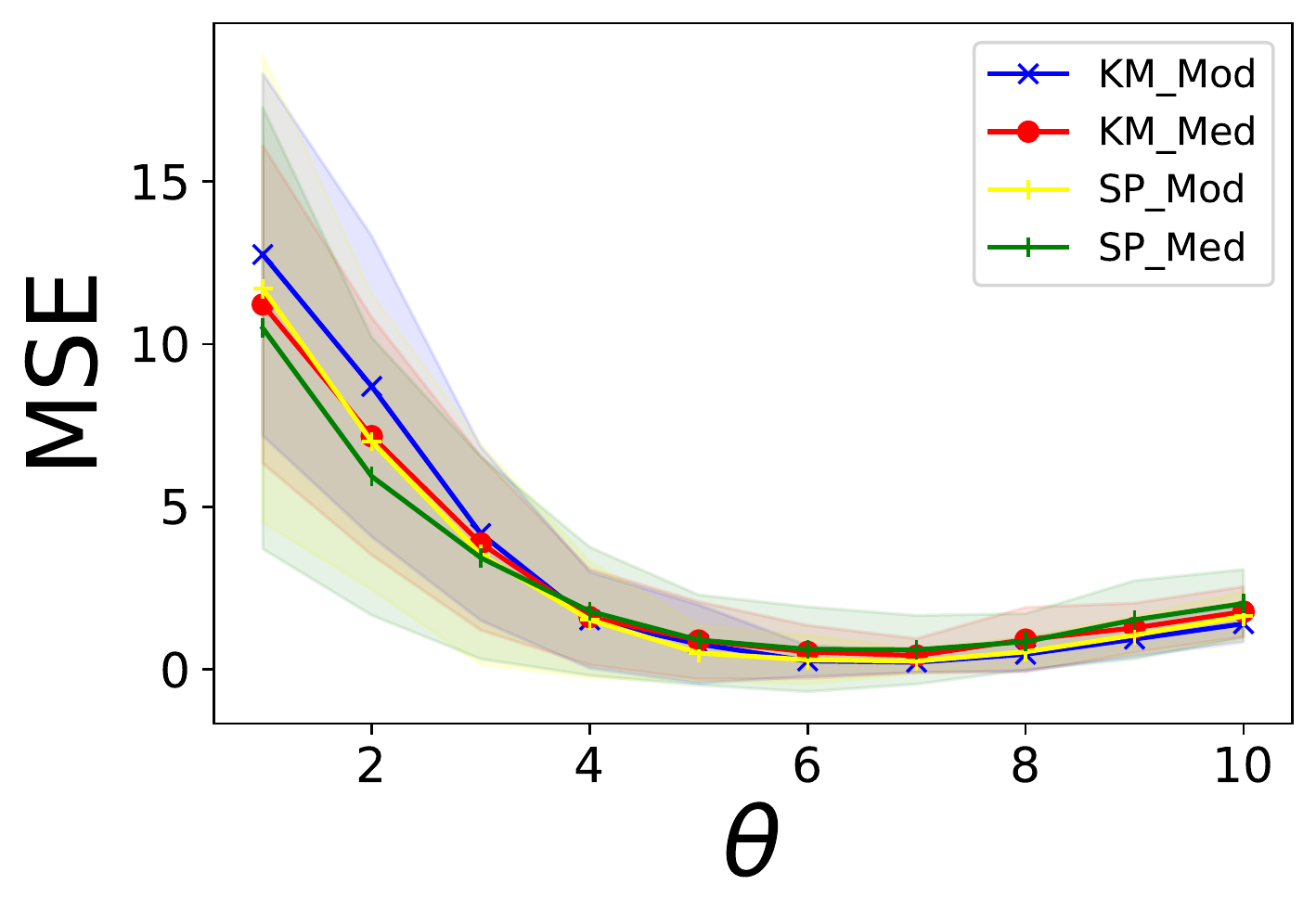}
      
    &
      \includegraphics[width=\linewidth,trim=0cm 0cm 0cm 0cm,clip]{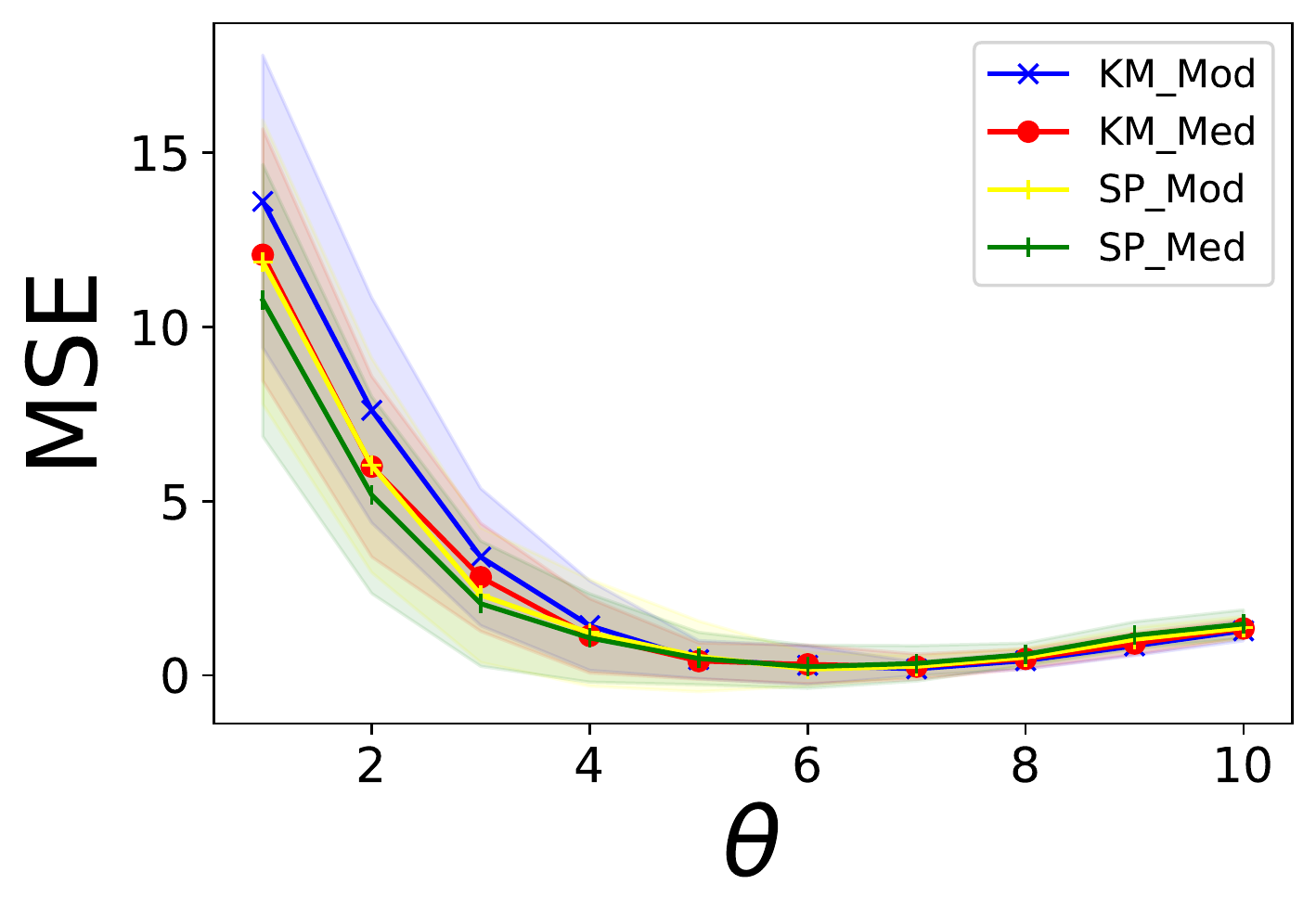}

    \\
    \multicolumn{1}{c}{$k=1$} & \multicolumn{1}{c}{$k=2$} & \multicolumn{1}{c}{$k=3$}  \\
    \end{tabular}
\captionof{figure}{Average and confidence interval for the MSE with different levels of $\theta$ based on 100 simulations when $\sigma = 1$.}
\label{tab:mse_vs_threshold}
\end{table}

Figure \ref{tab:mse_vs_sigma_threshold} shows the average MSE as a function of $\sigma$ and $\theta$. These are based on 100 simulations for every setting. The plots for each panel represent the methods, which are KM\_Mod, KM\_Med, SP\_Mod, and SP\_Med, from top to bottom, respectively. For the homogeneous setting ($k=1$), we can observe that for all four methods, the MSE is heavily influenced by the increase in $\sigma$, which is consistent with Figure \ref{tab:mse_vs_sigma}. We also note that when the voting threshold increases, there are no significant changes in the MSE which is consistent with Figure \ref{tab:mse_vs_threshold}. For the heterogeneous case, we note that, in general, the optimal MSE is achieved when the voting threshold is approximately $6$ across a wide range of values of $\sigma$. This is consistent with Figures \ref{tab:mse_vs_sigma} and \ref{tab:mse_vs_threshold}, and suggests that the choice of the voting threshold is robust to model noise.

\begin{table}[htbp]
  \centering
    \begin{tabular}{p{4.5cm}|p{4.5cm}p{4.5cm}}
      \multicolumn{1}{c}{\textbf{Homogeneous Setting}}    &  \multicolumn{2}{c}{\textbf{Heterogeneous Setting}}  \\

      \includegraphics[width=\linewidth,trim=0cm 0cm 0cm 0cm,clip]{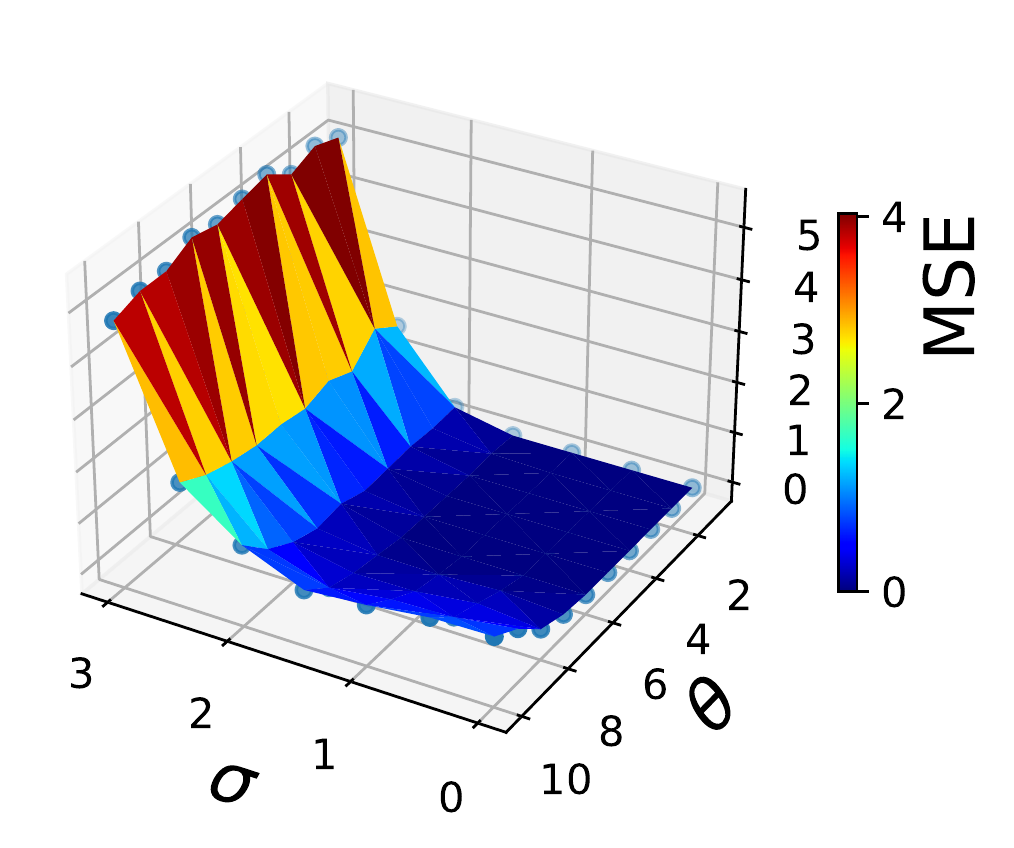}

      \includegraphics[width=\linewidth,trim=0cm 0cm 0cm 0cm,clip]{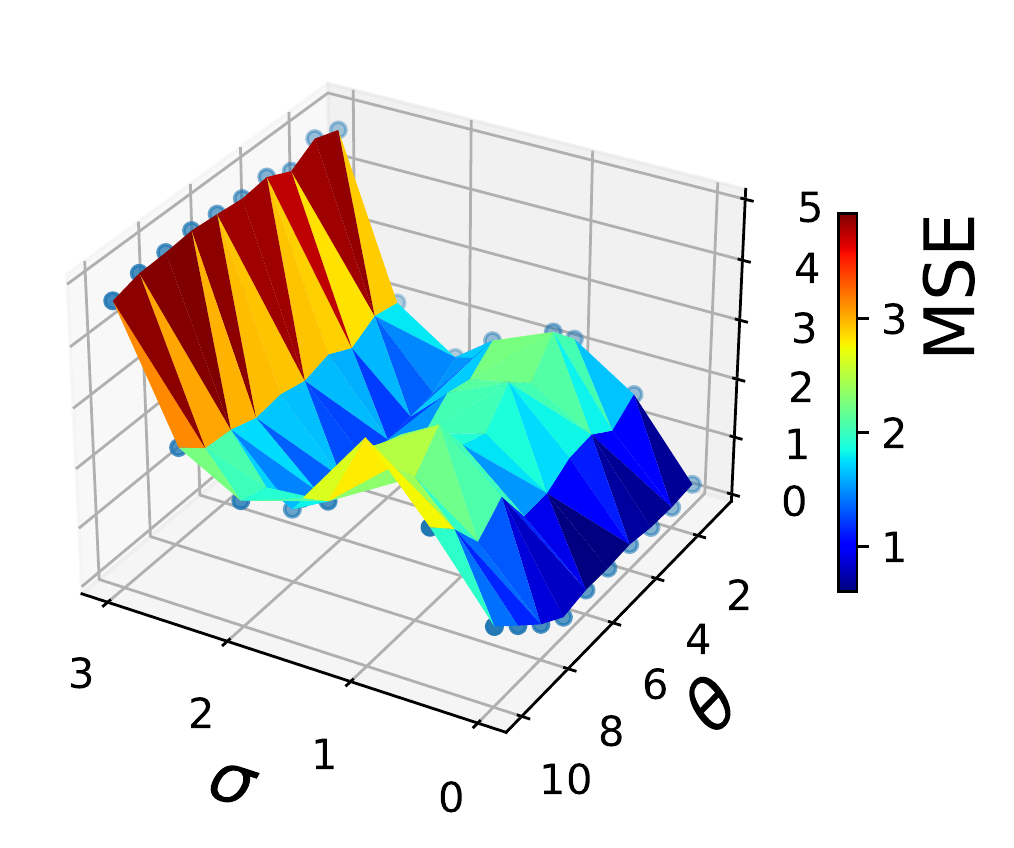}

    &
      \includegraphics[width=\linewidth,trim=0cm 0cm 0cm 0cm,clip]{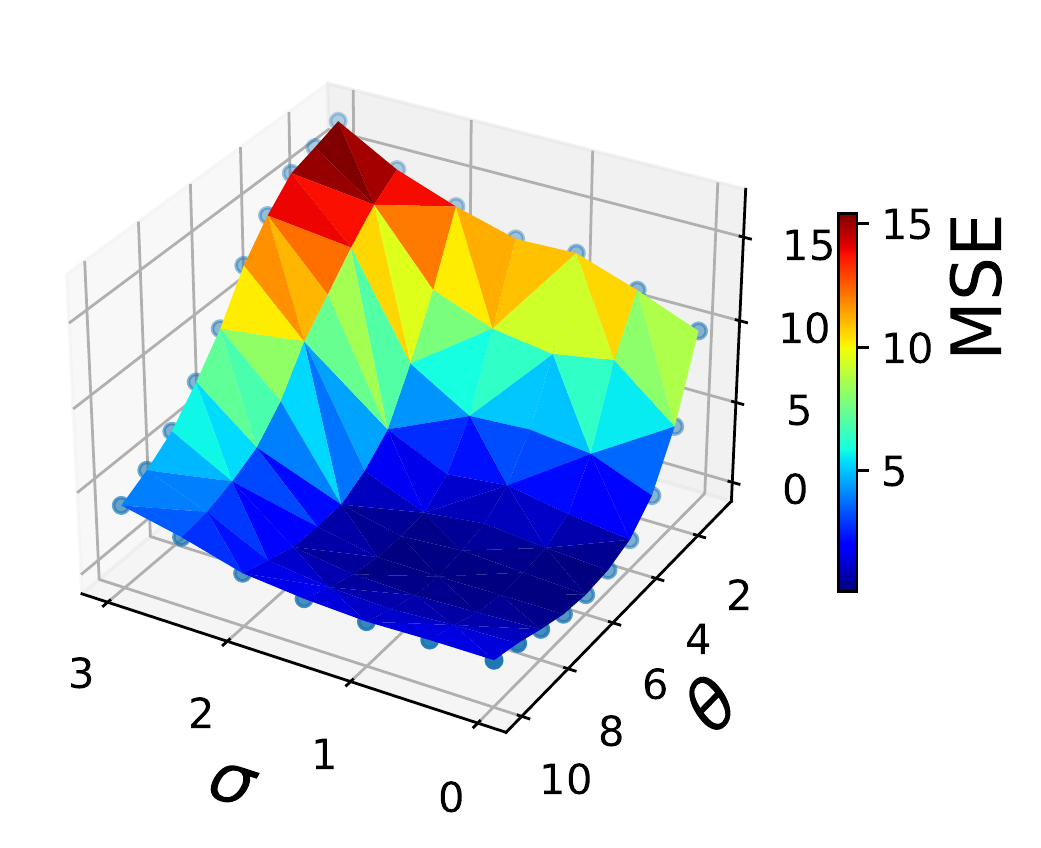}

      \includegraphics[width=\linewidth,trim=0cm 0cm 0cm 0cm,clip]{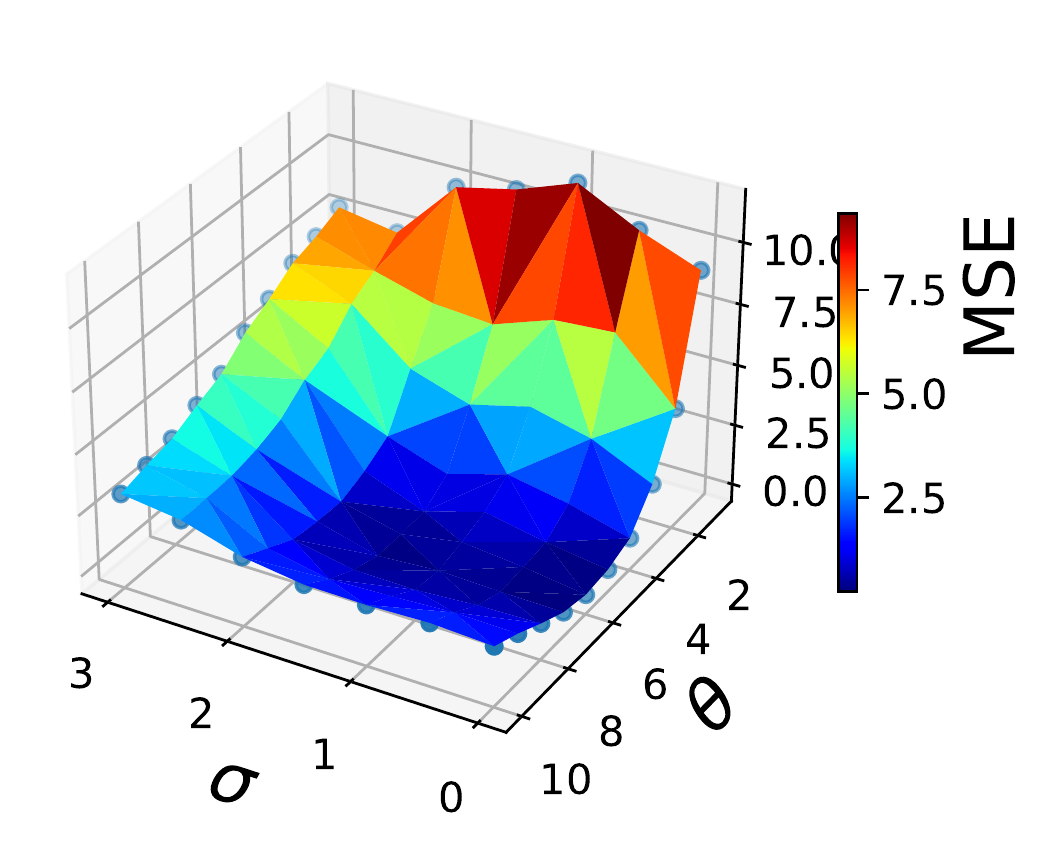}

    &
      \includegraphics[width=\linewidth,trim=0cm 0cm 0cm 0cm,clip]{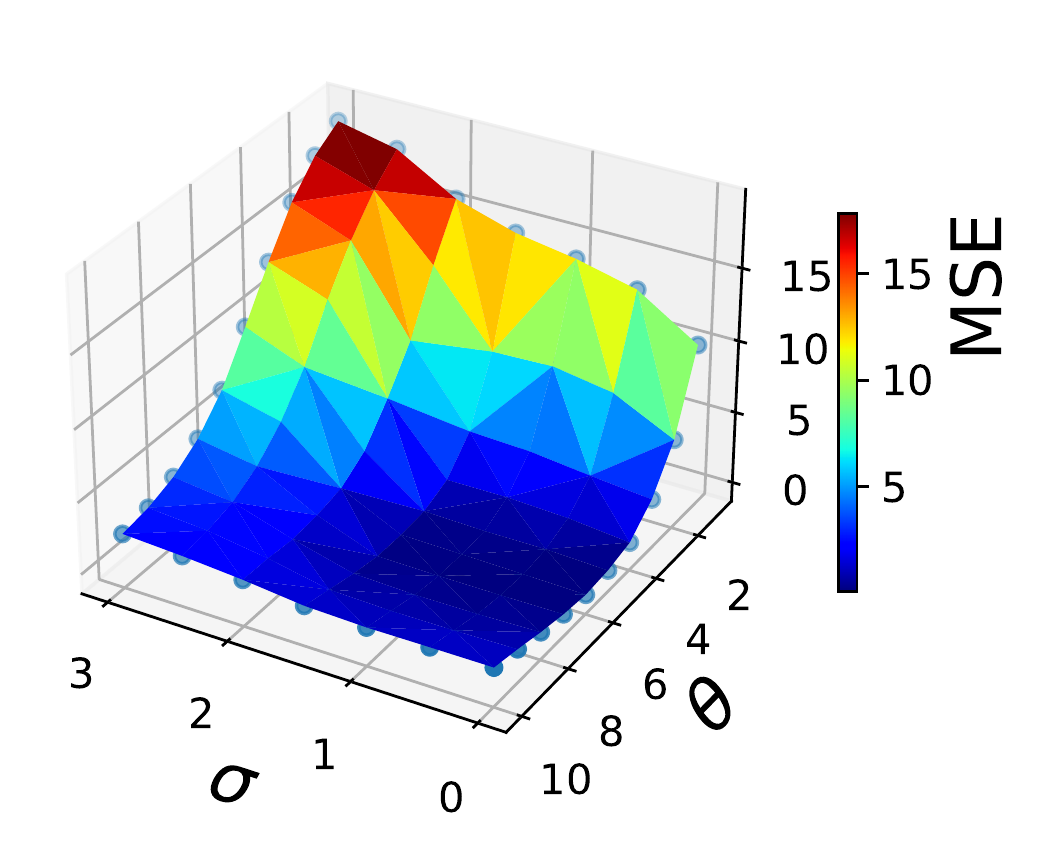}

    \includegraphics[width=\linewidth,trim=0cm 0cm 0cm 0cm,clip]{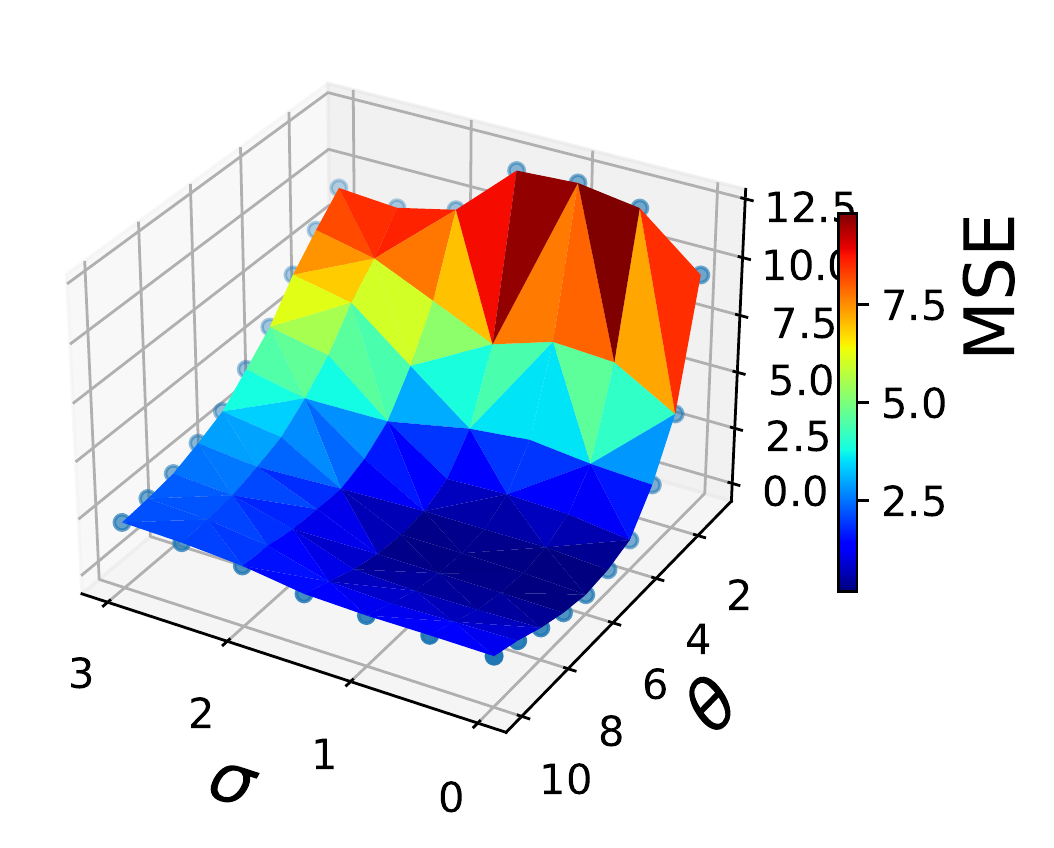}

    \\
    \end{tabular}

\end{table}

\begin{table}[htbp]
  \centering
    \begin{tabular}{p{4.5cm}|p{4.5cm}p{4.5cm}}



      \includegraphics[width=\linewidth,trim=0cm 0cm 0cm 0cm,clip]{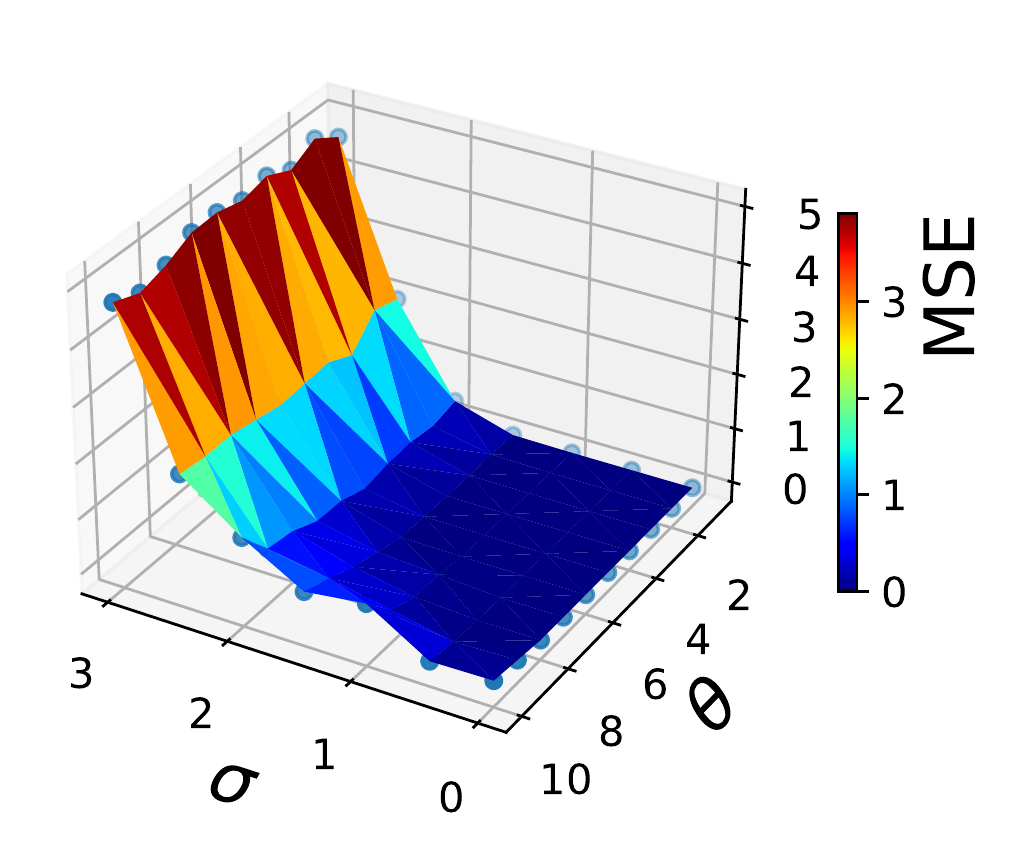}

      \includegraphics[width=\linewidth,trim=0cm 0cm 0cm 0cm,clip]{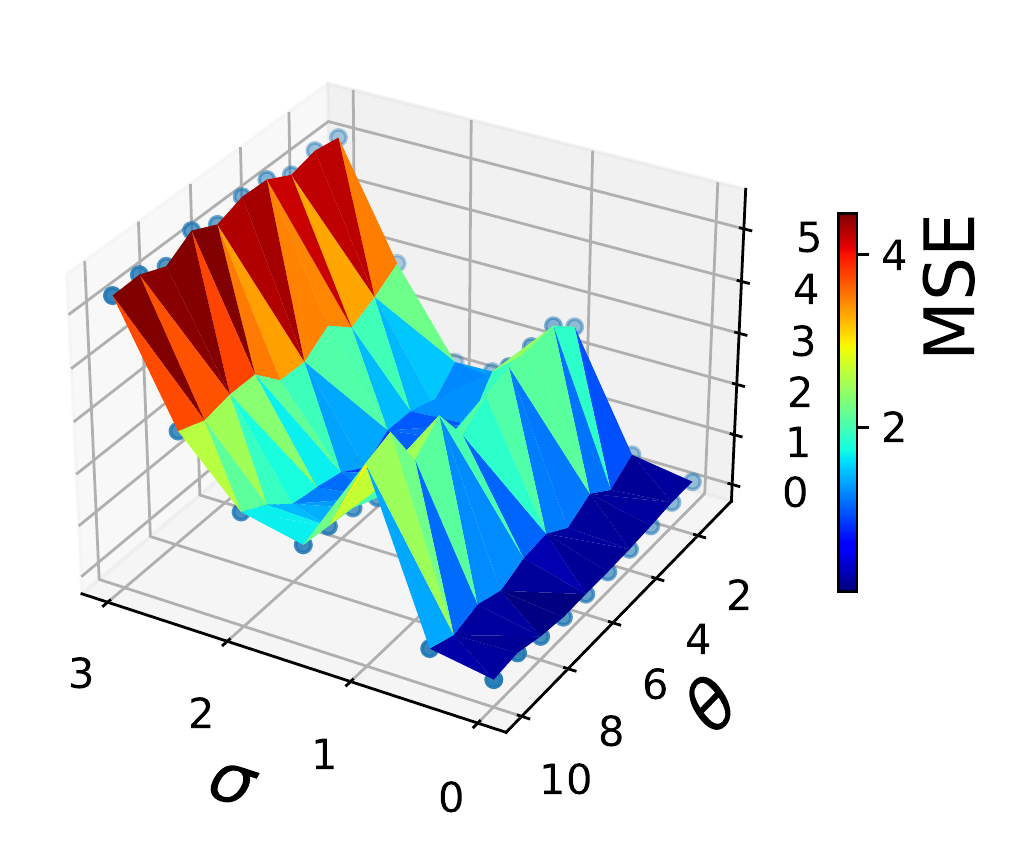}

    &

      \includegraphics[width=\linewidth,trim=0cm 0cm 0cm 0cm,clip]{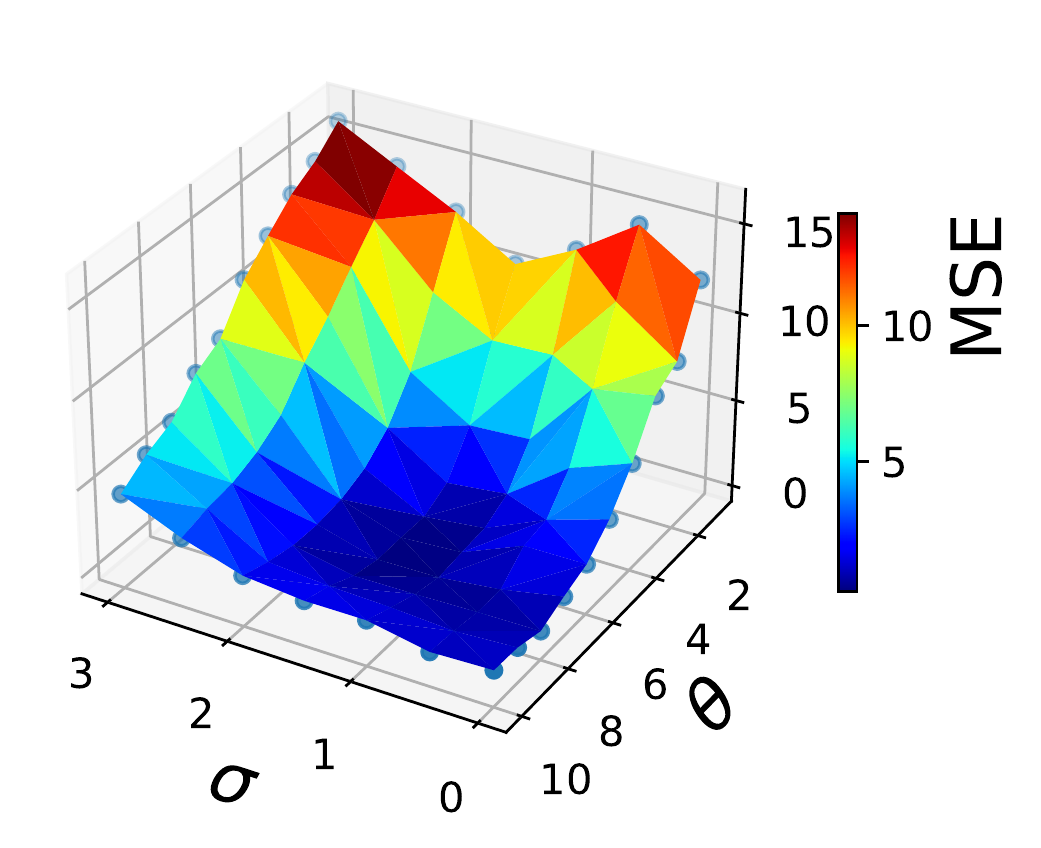}

      \includegraphics[width=\linewidth,trim=0cm 0cm 0cm 0cm,clip]{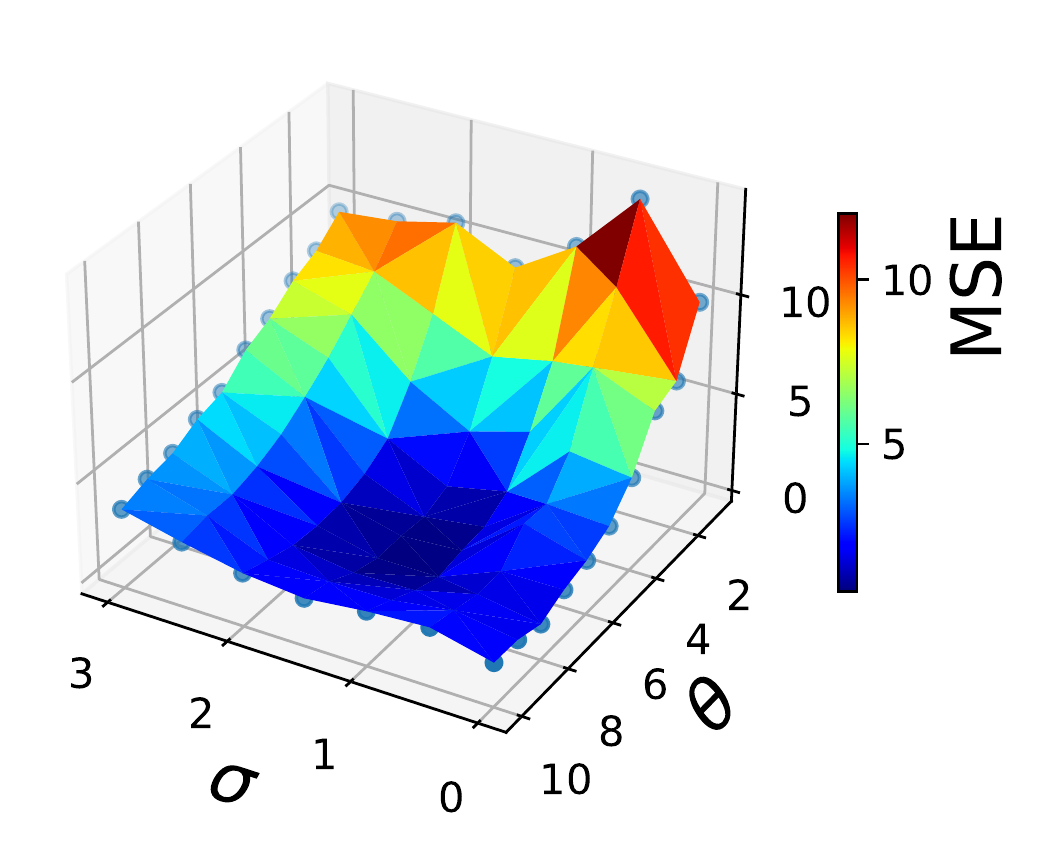}
      
    &

      \includegraphics[width=\linewidth,trim=0cm 0cm 0cm 0cm,clip]{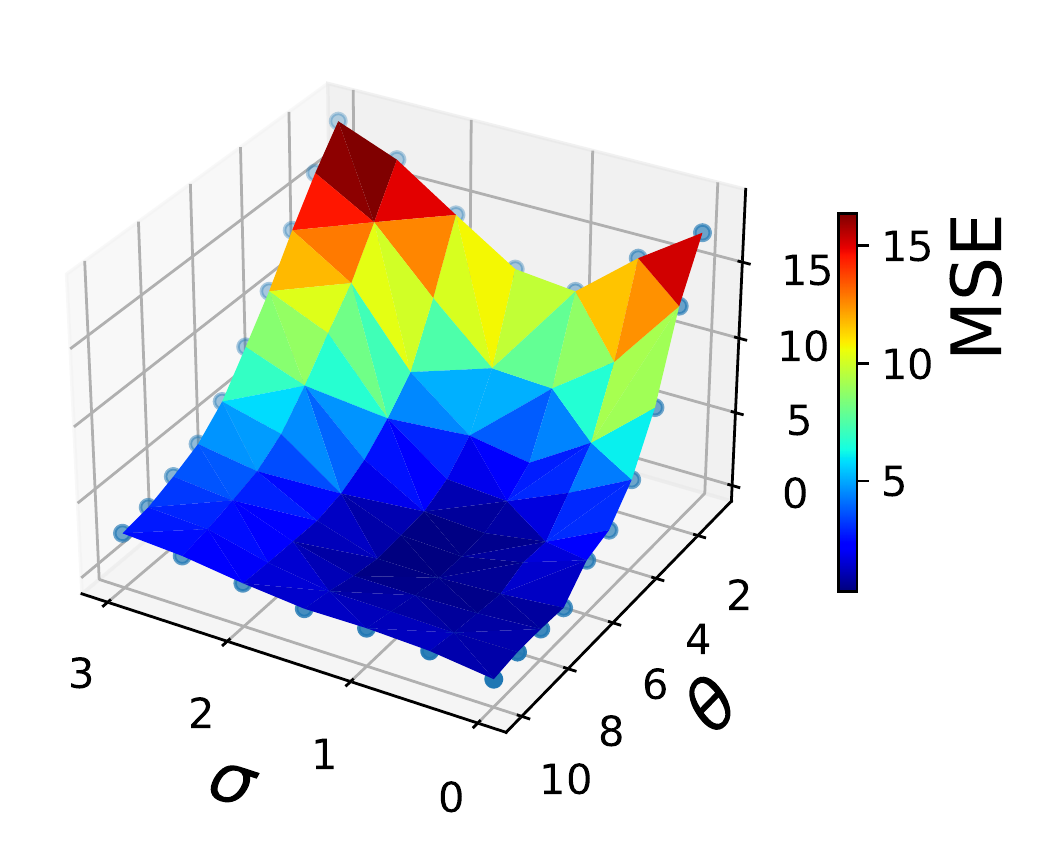}

      \includegraphics[width=\linewidth,trim=0cm 0cm 0cm 0cm,clip]{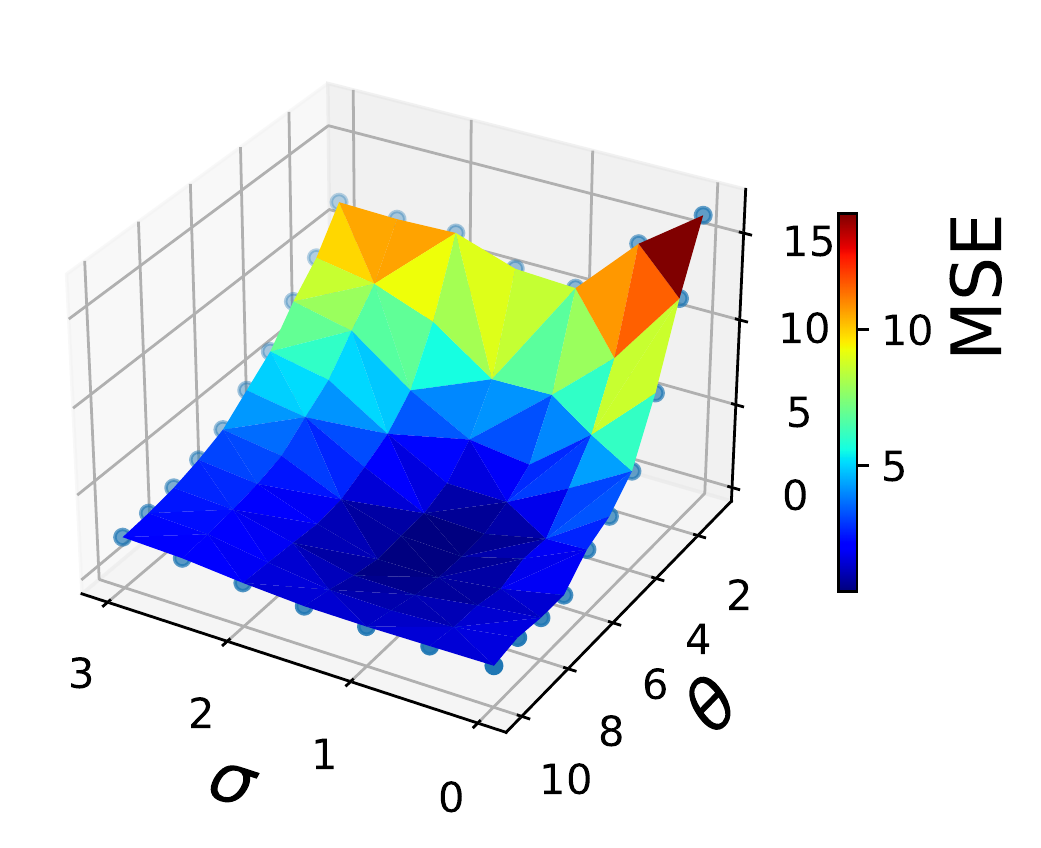}

    \\
    \multicolumn{1}{c}{$k=1$} & \multicolumn{1}{c}{$k=2$} & \multicolumn{1}{c}{$k=3$}  \\
    \end{tabular}
\captionof{figure}{Average MSE as a function of $\sigma$ and $\theta$ based on 100 simulations. From top panel to bottom panel, KM\_Mod, KM\_Med, SP\_Mod, and SP\_Med.}
\label{tab:mse_vs_sigma_threshold}
\end{table}

\section{Financial data experiments}
\label{sec: financial}

\subsection{Data description}

In this section, we apply our methods to a large-scale experiment using financial data. As mentioned earlier, this is a context where lead-lag relationships naturally occur. For the financial data experiments, we consider three data sets, which vary in terms of the number and type of assets and the number of days.

\begin{itemize}
\item For the first data set, we look at US equities from \href{https://wrds-www.wharton.upenn.edu/pages/about/data-vendors/center-for-research-in-security-prices-crsp/}{Wharton’s CRSP} data set. There are a total of $679$ equities within this data set, over a range of $5211$ trading days from 2000/01/03 to 2020/12/31. 
\item The second data set is derived from the same \href{https://wrds-www.wharton.upenn.edu/pages/about/data-vendors/center-for-research-in-security-prices-crsp/}{Wharton’s CRSP} data set., but looks at Exchange Traded Funds (ETFs). It consists of $14$ ETFs with $3324$ trading days from 2006/04/12 to 2019/07/01. 
\item The third data set is the \href{https://pinnacledata2.com/clc.html}{Pinnacle Data Corp CLC}, which includes $52$ futures contracts with $5166$ trading days from 2000/01/05 to 2020/10/16. All futures contracts are adjusted for rolling effects.  
This data set spans multiple asset classes, which include commodities, fixed income, and currency futures. 
\end{itemize}
All data sets are considered at daily frequency. We summarize all of the above in Table \ref{tab: dataset}, with additional details about Pinnacle Data Corp CLC available in Appendix \ref{sec:pinnacle} Tables [\ref{tab: Grains}, \ref{tab: Meats}, \ref{tab: Foodfibr}, \ref{tab: Metals}, \ref{tab: Indexes}, \ref{tab: Bonds}, \ref{tab: Currency}, \ref{tab: Oils}].

\begin{table}[htbp]
  \centering
  \caption{Summary of the three financial data sets considered in the numerical experiments.}
    \begin{tabular}{p{3cm}p{1cm}p{1cm}p{1.9cm}p{1.8cm}p{1.8cm}p{1.7cm}}
        \toprule
            \textbf{Data source} & \textbf{Type} & \textbf{Freq} & \textbf{\#\ of assets} & \textbf{Start date} & \textbf{End date} & \textbf{\#\ of days}\\
        \midrule
            Wharton’s CRSP     & Equity  & Daily & 679 & 2000/01/03 & 2020/12/31 & 5211\\
            Wharton’s CRSP     & ETF     & Daily & 14  & 2006/04/12 & 2019/07/01 & 3324\\
            Pinnacle Data Corp & Futures  & Daily & 52  & 2000/01/05 & 2020/10/16 & 5166\\ 
        \bottomrule
    \end{tabular}
\label{tab: dataset}
\end{table}

\subsection{Data pre-processing}

With regard to the US equity and ETF data sets, we download the close-to-close adjusted daily returns from Wharton’s CRSP. Due to the large number of NaNs in the equity data set, we drop the days for which more than 10\% of the equities have zero returns as well as the equities for which more than 50\% of days have zero returns. Instead of working with raw returns, we consider the market excess returns, a standard measure of how well each equity performed relative to the broader market. For both of these data sets, the return of the S\&P Composite Index is selected to compute the market excess returns by subtracting it from the return of each asset (i.e., for simplicity, we assume each asset has $\beta=1$ exposure to the market). Also, we winsorize the extreme value of excess returns for which any value is larger than 0.15 or smaller than -0.15.

For the futures data set, we download the close-to-close price series from the Pinnacle Data Corp CLC data set, and discard the days for which more than 10\% of the futures have zero prices in the respective dates, and drop the futures for which more than 160 days have zero prices. Afterwards, we first use forward-fill, then backward-fill to fill out the zero prices. Lastly, we compute the log-return from the close-to-close price. The remainder of the data pre-processing is the same as above.

\subsection{Benchmark}

In order to evaluate our proposed methodology, we also introduce a benchmark to detect lead-lag relationships without the use of clustering. It is very common to compute a sample cross-correlation function (CCF) between two time series. A CCF between time series $X_i$ and $X_j$ evaluated at lag $m$ is given by

\begin{equation} \label{eq:CCF}
\textrm{CCF}^{ij}(m) = \text{CORR}(\{X_{i}^{t-m}\},\{X_{j}^{t}\}),
\end{equation}

\noindent
where CORR() denotes a choice of the CCF. The corresponding lead-lag matrix $\Gamma_{n \times n}$ is estimated by computing the signed normalized area under the curve of CCF, given by

\begin{equation} \label{eq:CCF-auc}
\Gamma_{ij} =\frac{\text{MAX} (I(i, j), I(j, i)) \cdot \text{SIGN}(I(i, j)-I(j, i))}{I(i, j)+I(j, i)},
\end{equation}

\noindent
where $I(i, j)=\sum_{m=1}^M\left|\textrm{CCF}^{ij}(m)\right|$ for a user-specified maximum lag $M$.

We summarize the benchmark procedures in Algorithm \ref{benchmark}.

\begin{algorithm}[htp] 
\caption{\textbf{\small: CCF Algorithm}}
\label{benchmark}
\hspace*{\algorithmicindent} \textbf{Input:} Time series matrix $X_{n \times T}$. \\
\hspace*{\algorithmicindent} \textbf{Output:} Lead-lag matrix $\Gamma_{n \times n}$.
\begin{algorithmic}[1]
\State Calculate CCF for every pair of time series $\{X_{i}, X_{j}\}$.
\State Calculate the lead-lag matrix $\Gamma_{n \times n}$ by computing the signed normalized area under the curve of CCF.
\end{algorithmic}
\end{algorithm}

\subsection{Trading strategies}

Given $n$ time series each of length $T$, we first extract data by a sliding window with length $l = 21$. Next, we begin by extracting the STS with length $q = 10$ via a sliding window shifted by $s=1$ from the data, and cluster the STS to calculate the estimated lead-lag matrix $\Gamma_{n \times n}$ after applying the voting threshold $\theta = 6$, which is validated in the synthetic data experiment. We then utilize the lead-lag matrix to rank the time series from the most leading to the most lagging using the \textit{RowSum} ranking [\cite{huber1963pairwise}, \cite{gleich2011rank}], in order to then group the time series into leaders and laggers, with the former used to predict the latter.

\begin{table}[htbp]
  \centering
    \begin{tabular}{p{7cm}|p{7cm}}

      \includegraphics[width=\linewidth,trim=2cm 5.5cm 2cm 5.3cm,clip]{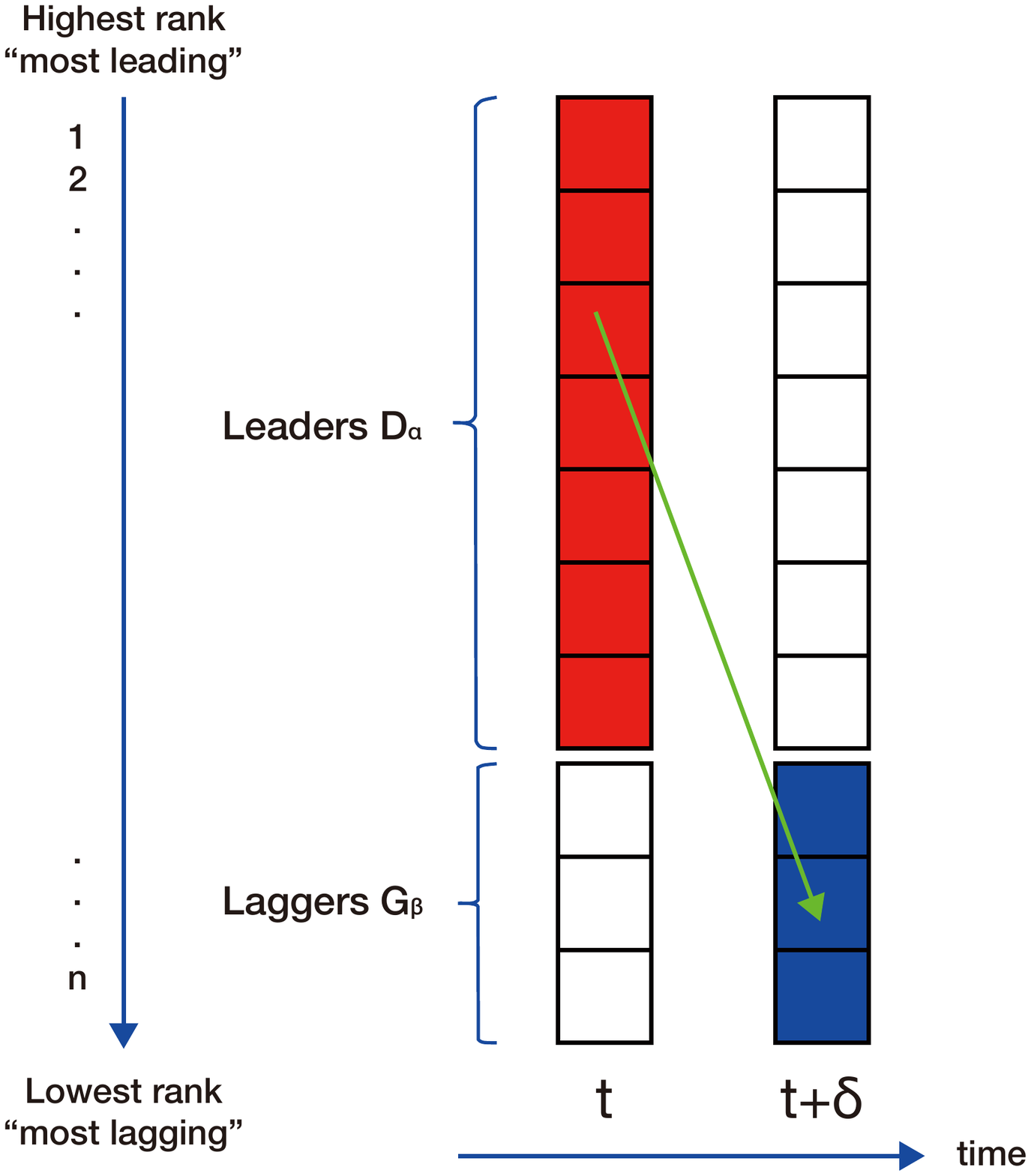}

    &
      \includegraphics[width=\linewidth,trim=2cm 5.5cm 2cm 5.3cm,clip]{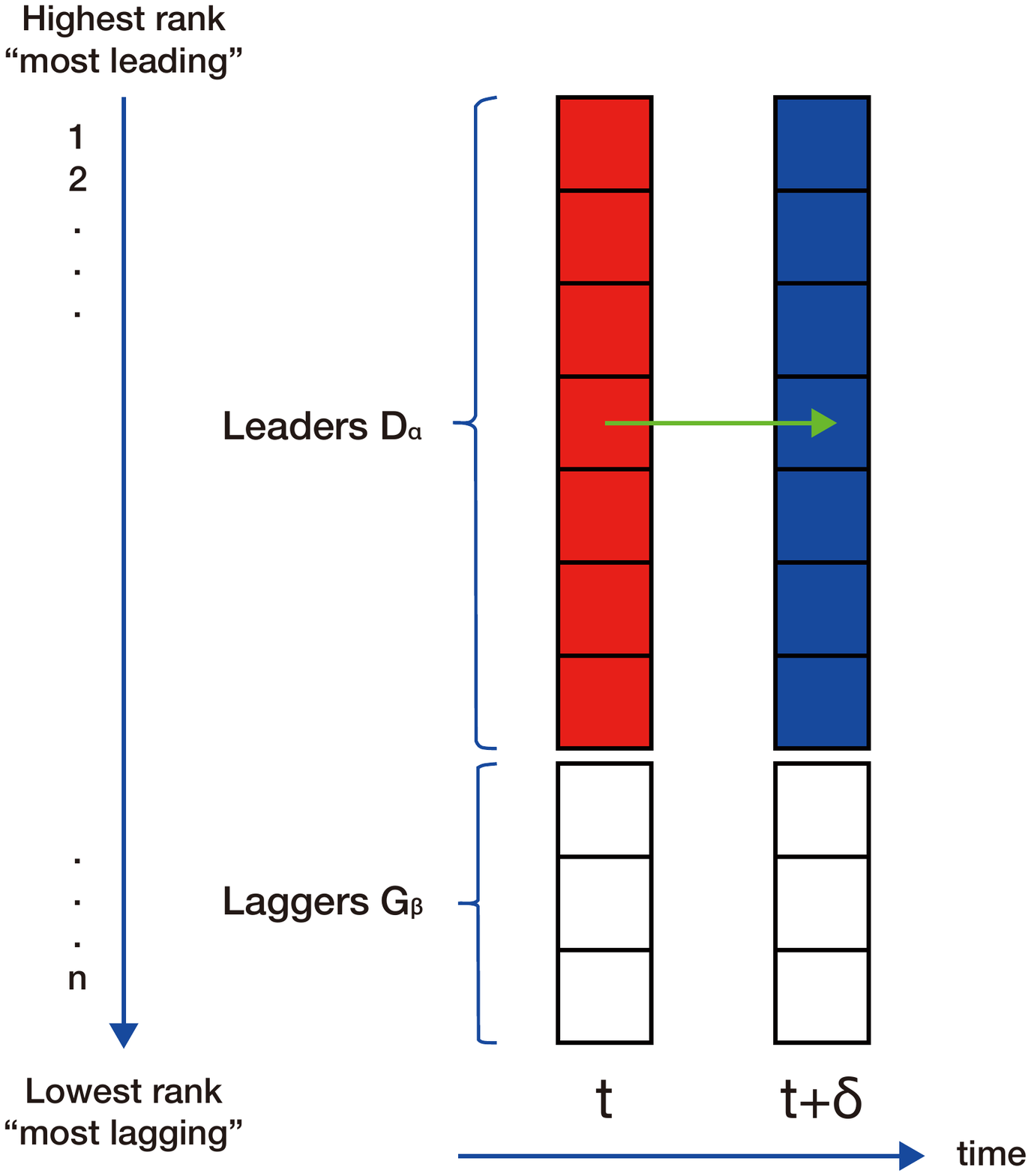}

    \\
    \multicolumn{1}{c}{$G_\beta$} & \multicolumn{1}{c}{$D_\alpha$} \\
    \end{tabular}
\captionof{figure}{$G_\beta$ strategy: Use $D_\alpha$ predict $G_\beta$ (left). $D_\alpha$ strategy: Use $D_\alpha$ predict $D_\alpha$ (right).}
\label{tab:ranking}
\end{table}

\begin{figure}[!htbp]
\centering
\includegraphics[width=0.8\textwidth,trim=0.5cm 0.5cm 0.5cm 0.5cm,clip]{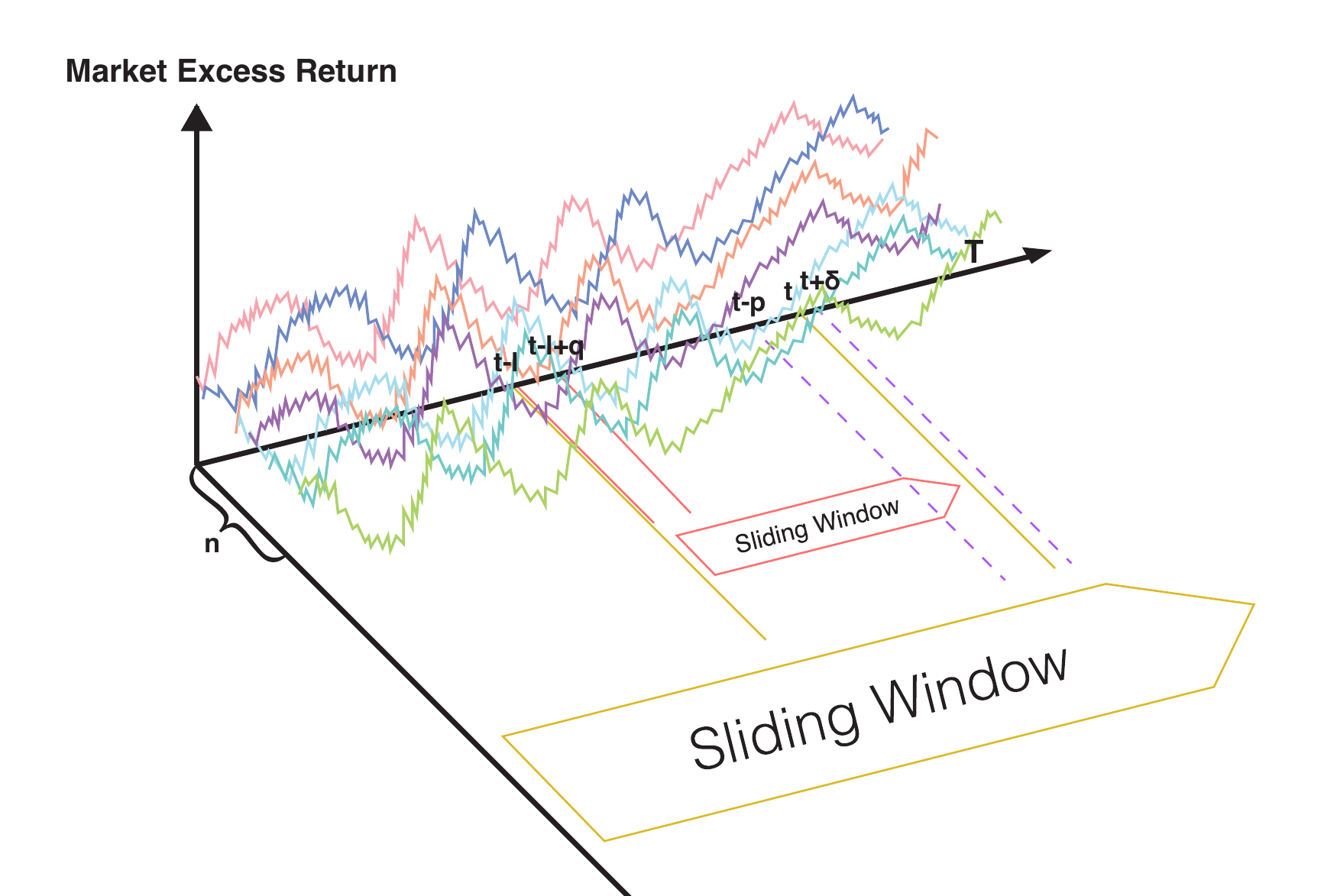}
\caption{Trading pipeline.}
\label{fig:sliding_window}
\end{figure}

Momentum is a widely studied phenomenon in the finance literature [\cite{jegadeesh2022momentum}, \cite{jegadeesh2001profitability}, \cite{tan2023spatio}, \cite{poh2022transfer}, \cite{wood2021trading}, \cite{wood2021slow}, \cite{lim2019enhancing}], which refers to the tendency of assets that have performed well in the recent past to continue to perform well in the near future, and vice versa. We denote the top $\alpha = \{0.7,0.75,0.8,0.85\}$ fraction of the time series as Leaders $D_\alpha$, and the bottom $\beta = 1 - \alpha$ as Laggers $G_\beta$. We use the exponentially weighted moving average (EWMA) signal on the past $p = \{1,3,5,7\}$ days of the average winsorized time series excess returns of the $D_\alpha$ to predict the average future $\delta = \{1,3,5,7\}$ days of the excess return of the $G_\beta$ and $D_\alpha$. We assume the $G_\beta$ can catch up with the $D_\alpha$, and the $D_\alpha$ provides the necessary momentum to maintain the trend in $\delta$ days, respectively. This is depicted in Figure \ref{tab:ranking}. Afterwards, we shift the sliding window by $s=1$, and re-apply the method to calculate the lead-lag matrix $\Gamma_{n \times n}$ and rank the time series until the end of the time series. For clarity, Figure \ref{fig:sliding_window} reflects our trading pipeline at time $t$, and we summarize the trading strategy in Algorithm \ref{strategy}.

\begin{algorithm}[htp] 
\caption{\textbf{\small: Trading strategy}}
\label{strategy}
\hspace*{\algorithmicindent} \textbf{Input:} Time series matrix $X_{n \times T}$. 
\begin{algorithmic}[1]
\State Build  $X_{n \times T}$ by applying a sliding window of length $l$ to obtain $X_{n \times l}$.
\State Use the \textbf{\small Lead-lag Relationship Detection Algorithm} on  $X_{n \times l}$ to obtain the lead-lag matrix $\Gamma_{n \times n}$. 
\State Based on $\Gamma_{n \times n}$, rank the time series from the most leading to the most lagging using the \textit{RowSum} ranking.
\State Pick the top $\alpha$ fraction of the time series as Leaders $D_\alpha$, and the bottom $\beta = 1 - \alpha$ as Laggers $G_\beta$.
\State Apply the EWMA on the past $p$ days of the average winsorized time series excess returns of the $D_\alpha$ to predict the average future $\delta$ days of the excess return of the $G_\beta$ and $D_\alpha$.
\State Shift the sliding window by $s$, and re-apply Steps 1 - 5 until the end of the time series.
\end{algorithmic}
\end{algorithm}


\subsection{Performance evaluation}

When assessing the effectiveness of various trading strategies, we rely on the following metrics to evaluate their performance:

We compute the Profit and Loss (PnL) of $G_\beta$ on a given day $t+\delta$ as 

    \begin{equation}
    \text{PnL}_{G_\beta}^{t+\delta} = \text{sign}(\text{EWMA}(ret_{D_\alpha}^{t-p}:ret_{D_\alpha}^{t}))\cdot\overline{ret_{G_\beta}^{t+\delta}}, t=l,\ldots,T-\delta,
    \end{equation}
since the strategy earns money whenever the sign of the forecast agrees with the sign of the future return. Correspondingly, the PnL of $D_\alpha$ on a given day $t+\delta$ is given by:

    \begin{equation}
    \text{PnL}_{D_\alpha}^{t+\delta} = \text{sign}(\text{EWMA}(ret_{D_\alpha}^{t-p}:ret_{D_\alpha}^{t}))\cdot\overline{ret_{D_\alpha}^{t+\delta}}, t=l,\ldots,T-\delta,
    \end{equation}

\noindent
where $ret_{D_\alpha}^{t-p}$ and $ret_{D_\alpha}^{t}$ are the excess return of $D_\alpha$ at $t-p$ and $t$, respectively, while $\text{EWMA}(ret_{D_\alpha}^{t-p}:ret_{D_\alpha}^{t})$ denotes the exponentially weighted moving average from the excess return of $D_\alpha$ from $t-p$ to $t$. Furthermore, $\overline{ret_{G_\beta}^{t+\delta}}$ depicts the mean of the excess return of $G_\beta$ at $t+\delta$, and $\overline{ret_{D_\alpha}^{t+\delta}}$ is the mean of the excess return of $D_\alpha$ at $t+\delta$.
We rescale the PnL by their volatility to target
equal risk assignment, and set our annualized volatility target $\sigma_{\text{tgt}}$ to be 0.15.

\begin{equation}
\text{PnL}_{\text{rescaled}} = \frac{\sigma_{\text{target}}}{\text{STD(PnL)}\cdot\sqrt{252}}\cdot\text{PnL}.
\end{equation}

The cumulative PnL sums the daily PnL across all trading days

\begin{equation}
\text{Cumulative PnL} = \sum\text{PnL}_{\text{rescaled}}.
\end{equation}

The annualized expected excess return (E[Returns]) is to measure the excess return earned by an investment over a benchmark index during a one-year period, and can be computed by

\begin{equation}
\text{E[Returns]} = \text{AVG}(\text{PnL}_{\text{rescaled}})\cdot 252.
\end{equation}

The annualized volatility is a measurement of the amount of risk associated with an investment over a one-year period

\begin{equation}
\text{Volatility} = \text{STD}(\text{PnL}_{\text{rescaled}})\cdot \sqrt{252}.
\end{equation}

Furthermore, we calculate the downside deviation and maximum drawdown to measure downside risk, and then the Sortino ratio is often used by investors who are more concerned with downside risk than with overall risk or volatility, which is derived by

\begin{equation}
\text{Sortino ratio} = \frac{\text{E[Returns]}}{\text{downside deviation}}.
\end{equation}

The Calmar ratio is often used by investors who are more concerned with long-term risk and downside protection. A higher Calmar ratio indicates that the strategy has generated higher returns relative to its maximum drawdown, while a lower Calmar ratio suggests that the strategy has underperformed given the level of risk it has taken. It is calculated by

\begin{equation}
\text{Calmar ratio} = \frac{\text{E[Returns]}}{\text{maximum drawdown}}.
\end{equation}

The hit rate measures the percentage of successful trades made by the strategy.  It is also known as the win rate or success rate, and is defined as 

\begin{equation}
\text{Hit rate} = \frac{|\text{PnL}_{\text{rescaled}}^{+}|}{|\text{PnL}_{\text{rescaled}}|},
\end{equation}
where $|\text{PnL}_{\text{rescaled}}^{+}|$ is the number of profitable trades, and $|\text{PnL}_{\text{rescaled}}|$ is the total number of trades.

The average profit / average loss (avg. profit / avg. loss) ratio measures the average size of profits relative to the average size of losses generated by the strategy.

\begin{equation}
\text{Avg. profit / avg. loss} = \frac{\text{AVG}(\text{PnL}_{\text{rescaled}}^{+})}{\text{AVG}(\text{PnL}_{\text{rescaled}}^{-})},
\end{equation}
where $\text{AVG}(\text{PnL}_{\text{rescaled}}^{+})$ is the average profit per trade, and $\text{AVG}(\text{PnL}_{\text{rescaled}}^{-})$ is the average loss per trade.

The PnL per trade illustrates the amount earned by the strategy, in basis points for each basket of $G_\beta$ or $D_\alpha$ traded in the markets (excluding transaction
costs), and is given by

\begin{equation}
\text{PnL per trade} = \text{AVG}(\text{PnL}_{\text{rescaled}})\cdot 10^{4},
\end{equation}
where we assume that the strategy trades the same amount of notional every day (i.e.,  a constant unit bet size is used every trading day).

We also compute the annualized Sharpe ratio to quantify the profit gained per unit of risk taken
\begin{equation}
\text{Sharpe ratio} = \frac{\text{AVG}(\text{PnL}_{\text{rescaled}})}{\text{STD}(\text{PnL}_{\text{rescaled}})} \cdot\sqrt{252}.
\end{equation}

It is important to assess the statistical significance of Sharpe ratio when back-testing a sample of hypothetical strategies [\cite{bailey2014deflated}, \cite{ledoit2008robust}, \cite{michael2022option}]. We use a test with the null hypothesis $H_0: \text{Sharpe ratio}=0$, and implement the method proposed by [\cite{bailey2014deflated}] to compute the test statistic

\begin{equation}
\frac{(\text{Sharpe ratio})\cdot\sqrt{T-1}}{\sqrt{1-\gamma_{1}\cdot (\text{Sharpe ratio})+(\gamma_{2}-1)\cdot(\text{Sharpe ratio})^{2} / 4}},
\end{equation}

\noindent
where the Sharpe ratio is what we are testing, $T$ is the length of the sample, and $\gamma_{1}$ and $\gamma_{2}$ are the skewness and kurtosis of the returns distribution for the selected strategy, respectively. This test statistic is assumed to be standard normal under the null hypothesis.

To evaluate the predictive performance of our method, we construct a simple trading strategy.  If the strategy is profitable with a statistically significant Sharpe ratio, this indicates that we are able to leverage the discovered lead-lag relationships for the prediction task.

\subsection{Results}

For the equity data set, full results with different tuning settings are reported in \href{https://data.mendeley.com/datasets/2djzdjvn96/2}{supplemental material} [\cite{zhang2023supplemental}] Table S1. We find that using the EWMA on the past seven days of the average winsorized time series excess returns of the $D_\alpha$ to predict the average future seven days of the excess return of the $G_\beta$ and $D_\alpha$ with $\alpha = 0.75$ performs consistently well across all methods. In Tables [\ref{tab:equity_metric_lag}, \ref{tab:equity_metric_lead}], we report the performance of the CCF and four methods based on the various metrics (rescaled to target volatility). Across the four methods we proposed, we note that the Sharpe ratio values are relatively high. In addition to the Sharpe ratio, the P-value associated with the Sharpe ratio also supports our hypothesis that the Sharpe ratio is statistically significant. It is evident that most P-values are 0 while others are significantly lower than 0.05, which provides us with more confidence when analyzing the results of the experiment. In particular, according to the $G_\beta$ strategy, it has been observed that KM\_Med outperforms other methods in terms of most metrics. Conversely, based on the $D_\alpha$ strategy, KM\_Mod performs better than other methods across most metrics. It is noteworthy that KM clustering produces similar results for both the $G_\beta$ and $D_\alpha$ strategies, and they are also much faster than SP clustering in terms of running time. In Figure \ref{tab:equity_pnl}, we depict the cumulative PnL (rescaled to target volatility) across trading days for the scenarios in Tables [\ref{tab:equity_metric_lag}, \ref{tab:equity_metric_lead}]. The left and right plots present $G_\beta$, and $D_\alpha$ strategies, respectively. We also present the benchmark performance for comparison.

\begin{table}[htbp]
  \centering
  \caption{Equity data set: performance metrics for $G_\beta$ strategy - rescaled to target volatility.\\
The experiment has been set with the values $p = 7$, $\delta = 7$, and $\alpha = 0.25$.}
    \begin{tabular}{p{3.7cm}p{1.5cm}p{1.5cm}p{1.5cm}p{1.5cm}p{1.5cm}}
        \toprule
        \multicolumn{1}{c}{\textbf{$G_\beta$ strategy}}  &\multicolumn{1}{c}{\textbf{Benchmark}}  &  \multicolumn{4}{c}{\textbf{Proposed}}  \\
        \cmidrule(lr){2-2} \cmidrule(lr){3-6}
         & CCF & KM\_Mod & KM\_Med & SP\_Mod & SP\_Med \\
        \midrule
        E[Returns]       &0.089  & 0.126 & 0.118  & \textbf{0.127*}  & 0.104    \\
        Volatility        &0.15   & 0.15  & 0.15   & 0.15   & 0.15    \\
        Downside deviation      &0.105  & 0.103 & \textbf{0.101*}  & 0.103  & 0.105  \\
        Maximum drawdown    &-0.313 & -0.26 & -0.215 & \textbf{-0.214*} & -0.287   \\
        Sortino ratio  &0.85   & 1.222 & 1.166  & \textbf{1.237*}  & 0.988    \\  
        Calmar ratio       &0.285  & 0.484 & 0.548  & \textbf{0.594*}  & 0.362   \\
        Hit rate        &0.499  & \textbf{0.521*} & 0.51   & 0.516  & 0.519   \\
        Avg. profit / avg. loss   &\textbf{1.117*}  & 1.068 & 1.107  & 1.091  & 1.051   \\
        PnL per trade   &3.542  & 4.996 & 4.672  & \textbf{5.041*}  & 4.119    \\
        Sharpe ratio &0.595  & 0.839 & 0.785  & \textbf{0.847*}  & 0.692    \\
        P-value     &0.009  & \textbf{0*}     & \textbf{0*}      & \textbf{0*}      & 0.002  \\
        \bottomrule
    \end{tabular}

\label{tab:equity_metric_lag}
\end{table}

\begin{table}[htbp]
  \centering
  \caption{Equity data set: performance metrics for $D_\alpha$ strategy - rescaled to target volatility.\\
The experiment has been set with the values $p = 7$, $\delta = 7$, and $\alpha = 0.25$.}
    \begin{tabular}{p{3.7cm}p{1.5cm}p{1.5cm}p{1.5cm}p{1.5cm}p{1.5cm}}
        \toprule
        \multicolumn{1}{c}{\textbf{$G_\beta$ strategy}}  &\multicolumn{1}{c}{\textbf{Benchmark}}  &  \multicolumn{4}{c}{\textbf{Proposed}}  \\
        \cmidrule(lr){2-2} \cmidrule(lr){3-6}
         & CCF & KM\_Mod & KM\_Med & SP\_Mod & SP\_Med \\
        \midrule
        E[Returns]       &0.101  & 0.122  & 0.095 & 0.129  & \textbf{0.134*}  \\
        Volatility        &0.15   & 0.15   & 0.15  & 0.15   & 0.15   \\
        Downside deviation      &\textbf{0.105*}  & 0.107  & 0.108 & 0.106  & \textbf{0.105*}  \\
        Maximum drawdown    &-0.288 & -0.251 & \textbf{-0.21*} & -0.283 & -0.227 \\
        Sortino ratio  &0.964  & 1.148  & 0.876 & 1.222  & \textbf{1.269*}  \\
        Calmar ratio       &0.352  & 0.488  & 0.452 & 0.456  & \textbf{0.59*}   \\
        Hit rate        &0.518  & 0.521  & 0.513 & 0.522  & \textbf{0.525*}  \\
        Avg. profit / avg. loss   &1.05   & 1.069  & 1.066 & \textbf{1.072*}  & 1.065  \\
        PnL per trade   &4.018  & 4.856  & 3.765 & 5.121  & \textbf{5.312*}  \\
        Sharpe ratio &0.675  & 0.816  & 0.632 & 0.86   & \textbf{0.892*}  \\
        P-value     &0.003  & \textbf{0*}      & 0.005 & \textbf{0*}      & \textbf{0*}     \\
        \bottomrule
    \end{tabular}

\label{tab:equity_metric_lead}
\end{table}


\newpage

\begin{table}[!htbp]
  \centering
    \begin{tabular}{p{6.9cm}p{6.9cm}}
      \multicolumn{1}{c}{\textbf{$G_\beta$ strategy}}    &  
      \multicolumn{1}{c}{\textbf{$D_\alpha$ strategy}}  \\

      \includegraphics[width=\linewidth,trim=0cm 0cm 0cm 0cm,clip]{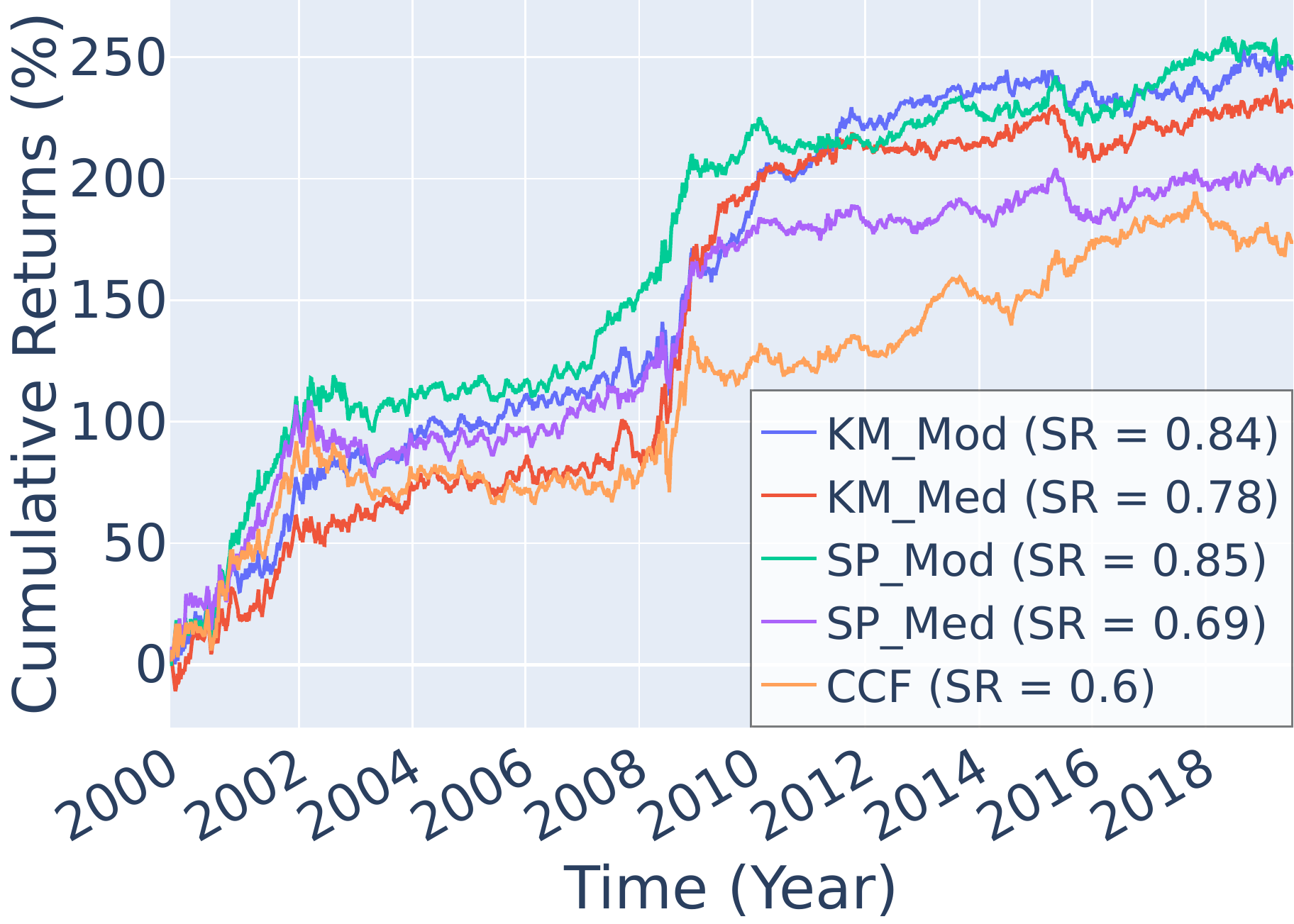}
      &
      \includegraphics[width=\linewidth,trim=0cm 0cm 0cm 0cm,clip]{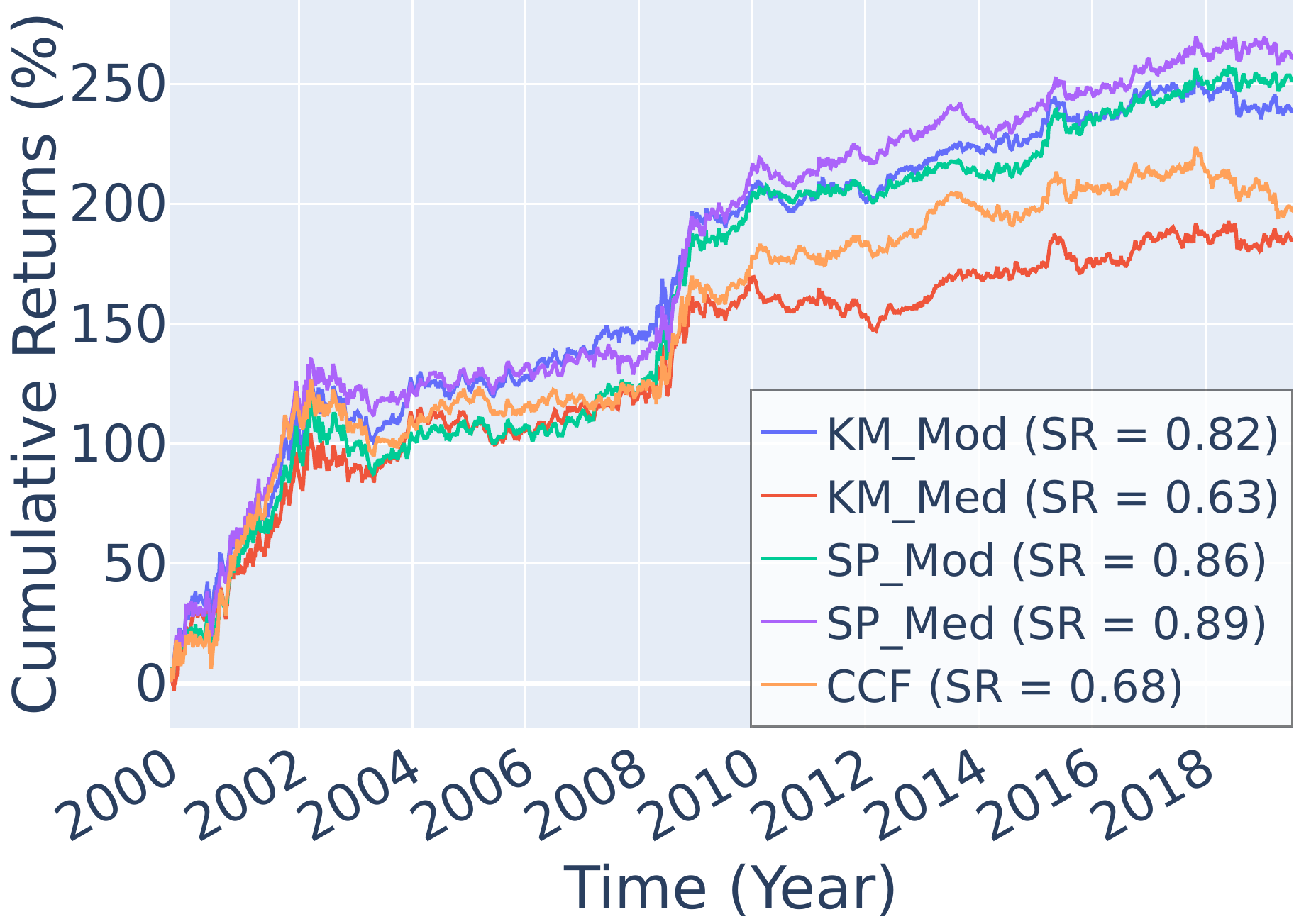}
    \end{tabular}

\vspace{-0.4cm}
\captionof{figure}{Equity data set: cumulative PnL for $G_\beta$ strategy (left) and $D_\alpha$ strategy (right) - rescaled to target volatility. The experiment has been set with the values $p = 7$, $\delta = 7$, and $\alpha = 0.25$.}
\label{tab:equity_pnl}
\end{table}

Tables [\ref{tab:etf_metric_lag}, \ref{tab:etf_metric_lead}] and Figure \ref{tab:etf_pnl} present results for the ETF data set using the same settings as for the equities data. Compared to equities data, we find no evidence of being able to consistently detect lead-lag relationships leading to a profitable outcome. We note, though, the performance is strong in some cases, namely the $D_{\alpha}$ strategy with the exception of KM\_Med. 
Full results across all tuning settings are reported in \href{https://data.mendeley.com/datasets/2djzdjvn96/2}{supplemental material} [\cite{zhang2023supplemental}] Table S2. Results for futures data, again with the same settings as the equities data, are reported in Tables [\ref{tab:futures_metric_lag}, \ref{tab:futures_metric_lead}] and Figure \ref{tab:futures_pnl}. For this data set, we do not see an ability to consistently detect profitable lead-lag relationships for any of the strategies. Full results across all tuning settings are reported in \href{https://data.mendeley.com/datasets/2djzdjvn96/2}{supplemental material} [\cite{zhang2023supplemental}] Table S3.

\begin{table}[htbp]
  \centering
  \caption{ETF data set: performance metrics for $G_\beta$ strategy - rescaled to target volatility.\\
The experiment has been set with the values $p = 7$, $\delta = 7$, and $\alpha = 0.25$.}
    \begin{tabular}{p{3.7cm}p{1.5cm}p{1.5cm}p{1.5cm}p{1.5cm}p{1.5cm}}
        \toprule
        \multicolumn{1}{c}{\textbf{$G_\beta$ strategy}}  &\multicolumn{1}{c}{\textbf{Benchmark}}  &  \multicolumn{4}{c}{\textbf{Proposed}}  \\
        \cmidrule(lr){2-2} \cmidrule(lr){3-6}
         & CCF & KM\_Mod & KM\_Med & SP\_Mod & SP\_Med \\
        \midrule
        E[Returns]       &    -0.019 & 0.02   & 0.022  & -0.005 & 0.013  \\ 
        Volatility        & 0.15   & 0.15   & 0.15   & 0.15   & 0.15   \\
        Downside deviation      &0.116  & 0.115  & 0.116  & 0.113  & 0.108  \\
        Maximum drawdown    &-0.668 & -0.525 & -0.465 & -0.579 & -0.369 \\
        Sortino ratio  &-0.165 & 0.176  & 0.188  & -0.043 & 0.122  \\
        Calmar ratio       &-0.029 & 0.038  & 0.047  & -0.008 & 0.036  \\
        Hit rate        &0.492  & 0.512  & 0.499  & 0.503  & 0.5    \\
        Avg. profit / avg. loss   &1.006  & 0.978  & 1.034  & 0.983  & 1.016  \\
        PnL per trade   &-0.76  & 0.8    & 0.87   & -0.191 & 0.527  \\
        Sharpe ratio &-0.128 & 0.134  & 0.146  & -0.032 & 0.088  \\
        P-value     &0.644  & 0.628  & 0.598  & 0.908  & 0.749 \\
        \bottomrule
    \end{tabular}

\label{tab:etf_metric_lag}
\end{table}

\begin{table}[htbp]
  \centering
  \caption{ETF data set: performance metrics for $D_\alpha$ strategy - rescaled to target volatility.\\
The experiment has been set with the values $p = 7$, $\delta = 7$, and $\alpha = 0.25$.}
    \begin{tabular}{p{3.7cm}p{1.5cm}p{1.5cm}p{1.5cm}p{1.5cm}p{1.5cm}}
        \toprule
        \multicolumn{1}{c}{\textbf{$G_\beta$ strategy}}  &\multicolumn{1}{c}{\textbf{Benchmark}}  &  \multicolumn{4}{c}{\textbf{Proposed}}  \\
        \cmidrule(lr){2-2} \cmidrule(lr){3-6}
         & CCF & KM\_Mod & KM\_Med & SP\_Mod & SP\_Med \\
        \midrule
        E[Returns]       &0.056  & 0.065  & 0.022  & 0.073  & 0.081  \\  
        Volatility        &0.15   & 0.15   & 0.15   & 0.15   & 0.15   \\ 
        Downside deviation      &0.097  & 0.097  & 0.105  & 0.1    & 0.105  \\ 
        Maximum drawdown    &-0.362 & -0.267 & -0.385 & -0.272 & -0.382 \\
        Sortino ratio  &0.581  & 0.676  & 0.213  & 0.728  & 0.771  \\ 
        Calmar ratio       &0.156  & 0.245  & 0.058  & 0.269  & 0.211  \\
        Hit rate        &0.504  & 0.502  & 0.5    & 0.51   & 0.504  \\
        Avg. profit / avg. loss   &1.056  & 1.077  & 1.027  & 1.05   & 1.087  \\
        PnL per trade   &2.234  & 2.591  & 0.889  & 2.901  & 3.2    \\
        Sharpe ratio &0.375  & 0.435  & 0.149  & 0.487  & 0.538  \\
        P-value     &0.169  & 0.105  & 0.588  & 0.074  & 0.051  \\
        \bottomrule
    \end{tabular}

\label{tab:etf_metric_lead}
\end{table}

\begin{table}[!htbp]
  \centering
    \begin{tabular}{p{6.9cm}p{6.9cm}}
      \multicolumn{1}{c}{\textbf{$G_\beta$ strategy}}    &  
      \multicolumn{1}{c}{\textbf{$D_\alpha$ strategy}}  \\

      \includegraphics[width=\linewidth,trim=0cm 0cm 0cm 0cm,clip]{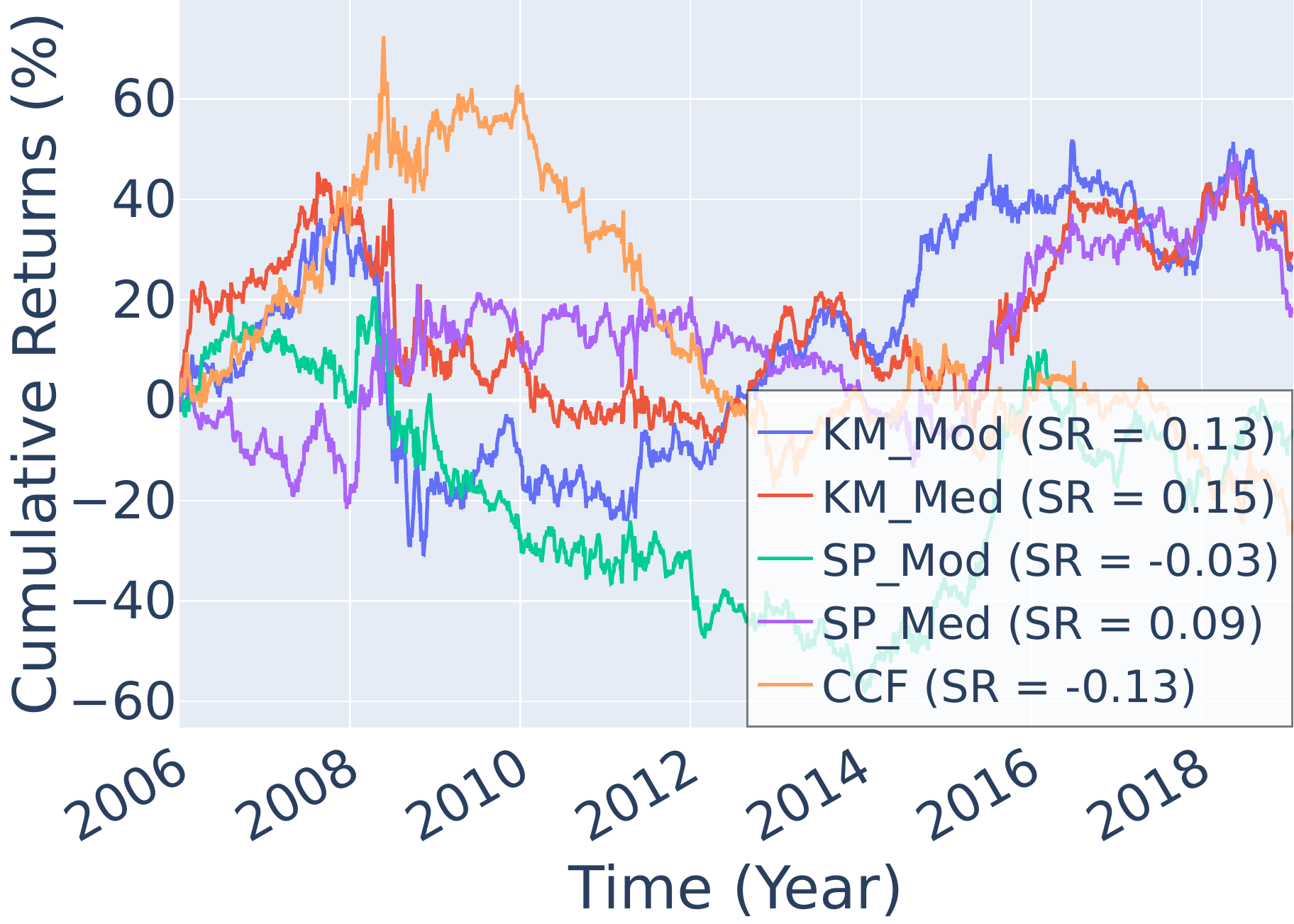}
      &
      \includegraphics[width=\linewidth,trim=0cm 0cm 0cm 0cm,clip]{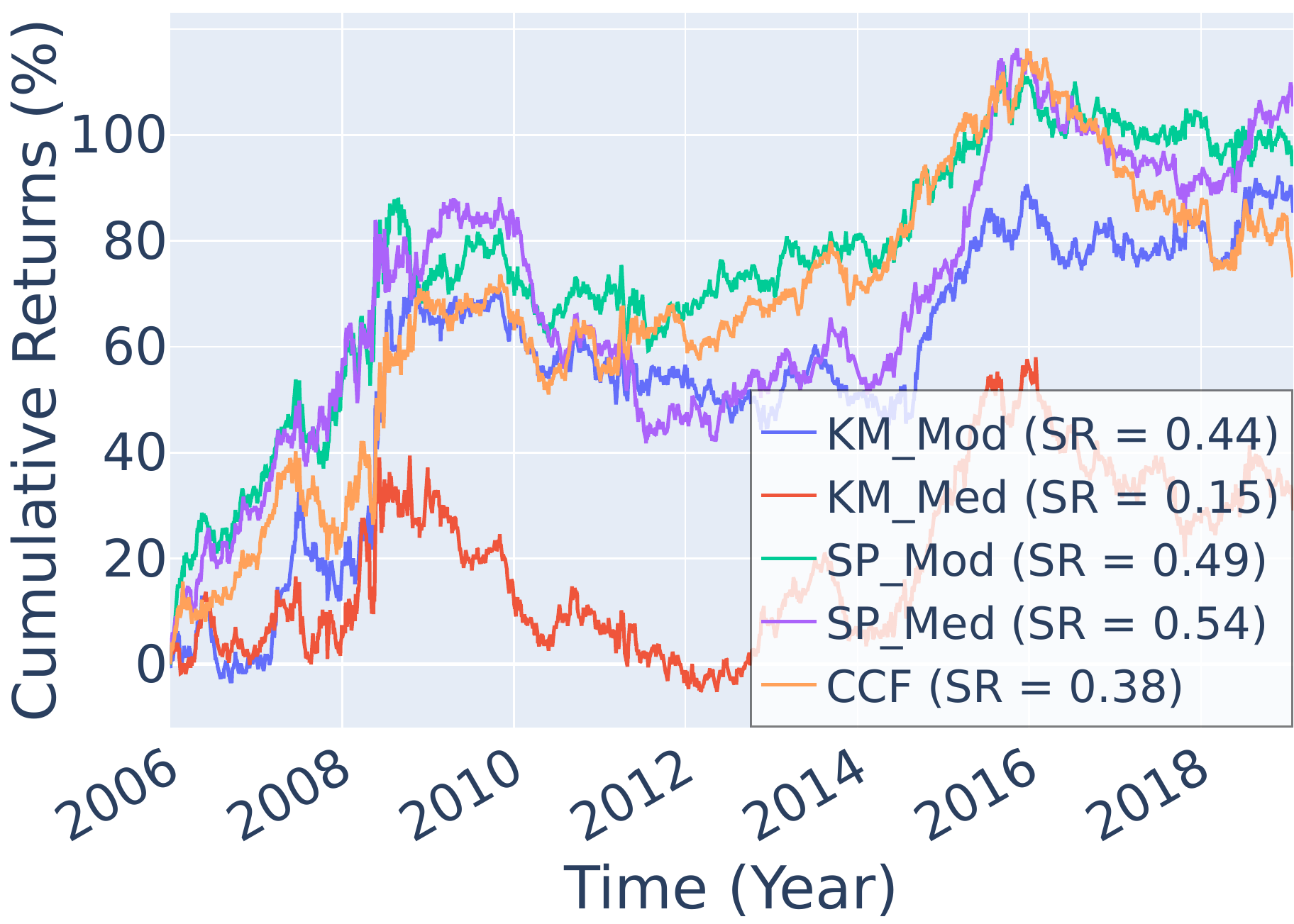}

    \\
    \end{tabular}
\vspace{-0.7cm}
\captionof{figure}{ETF data set: cumulative PnL for $G_\beta$ strategy (left) and $D_\alpha$ strategy (right) - rescaled to target volatility. The experiment has been set with the values $p = 7$, $\delta = 7$, and $\alpha = 0.25$.}
\label{tab:etf_pnl}
\end{table}

\begin{table}[htbp]
  \centering
  \caption{Futures data set: performance metrics for $G_\beta$ strategy - rescaled to target volatility.\\
The experiment has been set with the values $p = 7$, $\delta = 7$, and $\alpha = 0.25$.}
    \begin{tabular}{p{3.7cm}p{1.5cm}p{1.5cm}p{1.5cm}p{1.5cm}p{1.5cm}}
        \toprule
        \multicolumn{1}{c}{\textbf{$G_\beta$ strategy}}  &\multicolumn{1}{c}{\textbf{Benchmark}}  &  \multicolumn{4}{c}{\textbf{Proposed}}  \\
        \cmidrule(lr){2-2} \cmidrule(lr){3-6}
         & CCF & KM\_Mod & KM\_Med & SP\_Mod & SP\_Med \\
        \midrule
        E[Returns]       &0.013  & -0.001 & 0.005  & -0.01  & 0.022  \\
        Volatility        &0.15   & 0.15   & 0.15   & 0.15   & 0.15   \\
        Downside deviation      &0.104  & 0.102  & 0.103  & 0.102  & 0.105  \\
        Maximum drawdown    &-0.535 & -0.587 & -0.534 & -0.508 & -0.457 \\
        Sortino ratio  &0.121  & -0.007 & 0.045  & -0.097 & 0.209  \\
        Calmar ratio       &0.024  & -0.001 & 0.009  & -0.019 & 0.048  \\
        Hit rate        &0.502  & 0.498  & 0.495  & 0.494  & 0.508  \\
        Avg. profit / avg. loss   &1.007  & 1.009  & 1.028  & 1.013  & 0.994  \\
        PnL per trade   &0.499  & -0.028 & 0.185  & -0.392 & 0.869  \\
        Sharpe ratio &0.084  & -0.005 & 0.031  & -0.066 & 0.146  \\
        P-value     &0.705  & 0.983  & 0.888  & 0.766  & 0.51  \\
        \bottomrule
    \end{tabular}

\label{tab:futures_metric_lag}
\end{table}

\begin{table}[htbp]
  \centering
  \caption{Futures data set: performance metrics for $D_\alpha$ strategy - rescaled to target volatility.\\
The experiment has been set with the values $p = 7$, $\delta = 7$, and $\alpha = 0.25$.}
    \begin{tabular}{p{3.7cm}p{1.5cm}p{1.5cm}p{1.5cm}p{1.5cm}p{1.5cm}}
        \toprule
        \multicolumn{1}{c}{\textbf{$G_\beta$ strategy}}  &\multicolumn{1}{c}{\textbf{Benchmark}}  &  \multicolumn{4}{c}{\textbf{Proposed}}  \\
        \cmidrule(lr){2-2} \cmidrule(lr){3-6}
         & CCF & KM\_Mod & KM\_Med & SP\_Mod & SP\_Med \\
        \midrule
        E[Returns]       &0.061  & 0.036  & 0.036  & 0.027  & 0.031  \\
        Volatility        &0.15   & 0.15   & 0.15   & 0.15   & 0.15   \\
        Downside deviation      &0.107  & 0.108  & 0.108  & 0.108  & 0.107  \\
        Maximum drawdown    &-0.393 & -0.411 & -0.399 & -0.474 & -0.462 \\
        Sortino ratio  &0.574  & 0.336  & 0.335  & 0.253  & 0.29   \\
        Calmar ratio       &0.156  & 0.088  & 0.091  & 0.057  & 0.067  \\
        Hit rate        &0.511  & 0.502  & 0.499  & 0.498  & 0.501  \\
        Avg. profit / avg. loss   &1.032  & 1.038  & 1.05   & 1.043  & 1.037  \\
        PnL per trade   &2.439  & 1.441  & 1.443  & 1.079  & 1.23   \\
        Sharpe ratio &0.41   & 0.242  & 0.242  & 0.181  & 0.207  \\
        P-value     &0.064  & 0.274  & 0.273  & 0.412  & 0.35  \\
        \bottomrule
    \end{tabular}

\label{tab:futures_metric_lead}
\end{table}

\begin{table}[!htbp]
  \centering
    \begin{tabular}{p{6.9cm}p{6.9cm}}
      \multicolumn{1}{c}{\textbf{$G_\beta$ strategy}}    &  
      \multicolumn{1}{c}{\textbf{$D_\alpha$ strategy}}  \\

      \includegraphics[width=\linewidth,trim=0cm 0cm 0cm 0cm,clip]{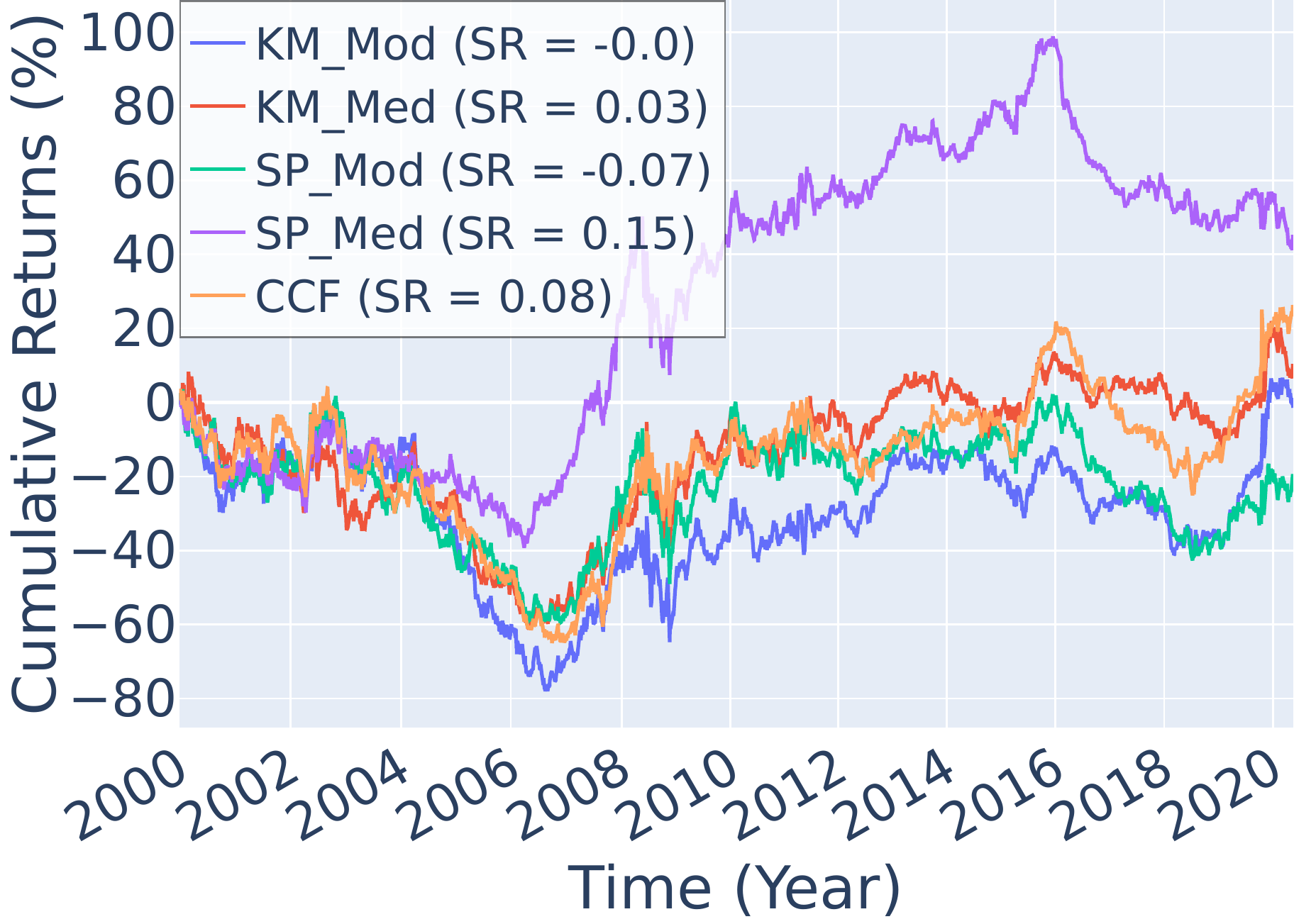}
      &
      \includegraphics[width=\linewidth,trim=0cm 0cm 0cm 0cm,clip]{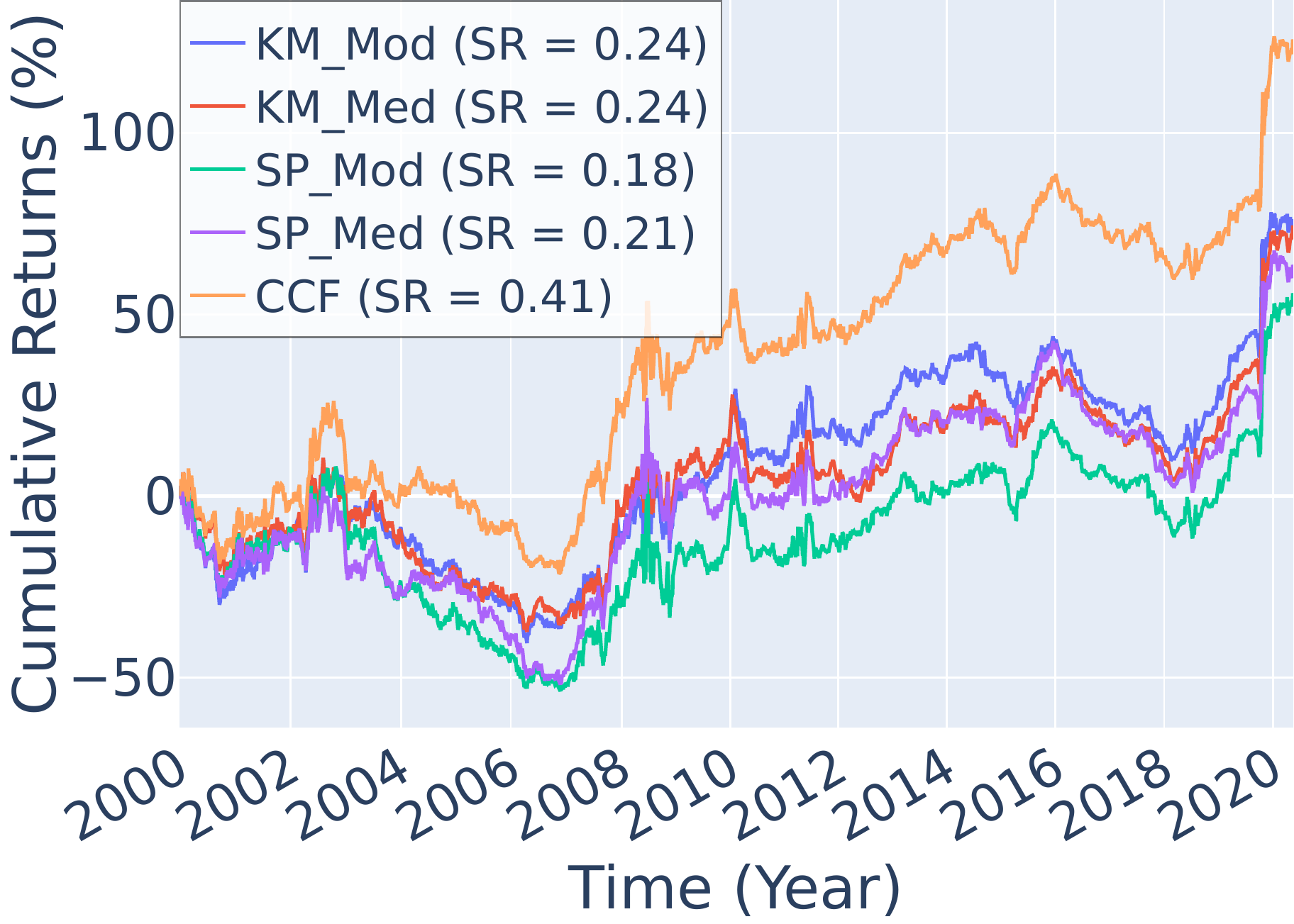}

    \\
    \end{tabular}
\vspace{-0.6cm}
\captionof{figure}{Futures data set: cumulative PnL for $G_\beta$ strategy (left) and $D_\alpha$ strategy (right) - rescaled to target volatility. The experiment has been set with the values $p = 7$, $\delta = 7$, and $\alpha = 0.25$.}
\label{tab:futures_pnl}
\end{table}

\newpage
\section{Robustness analysis}
\label{sec: robustness}

In this section, we test the robustness of the benchmark and our proposed methods by conducting experiments with different levels of $\alpha$.
For the equity data set, we consider $\alpha$ values ranging from $0.7$ to $0.85$ with an increment of $0.05$. In Table \ref{tab:equity_robust}, we note that the performance of all methods does not change significantly while maintaining a high Sharpe ratio. We notice that the P-values are almost all lower than 0.05, suggesting that all results are significant in our experiments. We also note that for the SP\_Med, the $G_\beta$ strategy has slightly lower performance. 

In Table \ref{tab:etf_robust}, we report on the performance of the four methods tested on the ETF data set. Here, we only consider the $\alpha$ values of $0.75$ and $0.85$ due to the smaller cross-section for this data set. The four methods achieve a fairly good performance for high alpha for the $D_{\alpha}$ strategy, except for KM\_Med.

Finally, in Table \ref{tab:futures_robust}, we provide the performance of the four methods on the futures data set, with $\alpha$ ranging from $0.7$ to $0.85$ in increments of $0.05$. We note that for this data set, the Sharpe ratio for the $G_{\beta}$ strategy tends to be more sensitive to change in $\alpha$ and it is hard to achieve profitability.

\begin{table}[htbp]
  \centering
  \caption{Equity data set: robustness analysis for $\alpha$ - rescaled to target volatility. The experiment has been set with the values $p = 7$, and $\delta = 7$.}
  \label{tab:equity_robust}
    \begin{tabular}{p{3.7cm}p{1cm}p{1cm}p{1cm}p{1cm}|p{1cm}p{1cm}p{1cm}p{1cm}}
        \toprule
        \multicolumn{1}{c}{\textbf{CCF}}  &\multicolumn{4}{c}{\textbf{$G_\beta$ strategy}}  &  \multicolumn{4}{c}{\textbf{$D_\alpha$ strategy}}  \\
        \cmidrule(lr){2-5} \cmidrule(lr){6-9}
         \textbf{$\alpha$} &  0.7 & 0.75 & 0.8 & 0.85 &  0.7 & 0.75 & 0.8 & 0.85 \\
        \midrule
        E[Returns]       &0.082  & 0.089  & 0.089 & 0.079  & 0.088  & 0.101  & 0.108  & 0.106  \\
        Volatility        &0.15   & 0.15   & 0.15  & 0.15   & 0.15   & 0.15   & 0.15   & 0.15   \\
        Downside deviation      &0.106  & 0.105  & 0.105 & 0.105  & 0.106  & 0.105  & 0.105  & 0.105  \\
        Maximum drawdown    &-0.313 & -0.313 & -0.29 & -0.259 & -0.267 & -0.288 & -0.254 & -0.254 \\
        Sortino ratio  &0.776  & 0.85   & 0.849 & 0.757  & 0.824  & 0.964  & 1.027  & 1.009  \\
        Calmar ratio       &0.263  & 0.285  & 0.306 & 0.306  & 0.328  & 0.352  & 0.425  & 0.418  \\
        Hit rate        &0.504  & 0.499  & 0.505 & 0.505  & 0.517  & 0.518  & 0.521  & 0.52   \\
        Avg. profit / avg. loss   &1.086  & 1.117  & 1.091 & 1.08   & 1.037  & 1.05   & 1.047  & 1.048  \\
        PnL per trade   &3.262  & 3.542  & 3.526 & 3.143  & 3.473  & 4.018  & 4.288  & 4.209  \\
        Sharpe ratio &0.548  & 0.595  & 0.592 & 0.528  & 0.583  & 0.675  & 0.72   & 0.707  \\
        P-value     &0.015  & 0.009  & 0.009 & 0.02   & 0.01   & 0.003  & 0.001  & 0.002 \\
        
        \midrule
        \midrule
        \multicolumn{1}{c}{\textbf{KM\_Mod}}  &\multicolumn{4}{c}{\textbf{$G_\beta$ strategy}}  &  \multicolumn{4}{c}{\textbf{$D_\alpha$ strategy}}  \\
        \cmidrule(lr){2-5} \cmidrule(lr){6-9}
         \textbf{$\alpha$} &  0.7 & 0.75 & 0.8 & 0.85 &  0.7 & 0.75 & 0.8 & 0.85 \\
         \midrule
        E[Returns]       &0.117  & 0.126 & 0.117  & 0.143  & 0.116  & 0.122  & 0.117  & 0.123  \\
        Volatility        &0.15   & 0.15  & 0.15   & 0.15   & 0.15   & 0.15   & 0.15   & 0.15   \\
        Downside deviation      &0.104  & 0.103 & 0.103  & 0.101  & 0.105  & 0.107  & 0.106  & 0.106  \\
        Maximum drawdown    &-0.245 & -0.26 & -0.236 & -0.213 & -0.207 & -0.251 & -0.289 & -0.288 \\
        Sortino ratio  &1.127  & 1.222 & 1.137  & 1.42   & 1.099  & 1.148  & 1.102  & 1.157  \\
        Calmar ratio       &0.476  & 0.484 & 0.494  & 0.672  & 0.559  & 0.488  & 0.405  & 0.427  \\
        Hit rate        &0.521  & 0.521 & 0.516  & 0.522  & 0.51   & 0.521  & 0.52   & 0.522  \\
        Avg. profit / avg. loss   &1.059  & 1.068 & 1.078  & 1.085  & 1.107  & 1.069  & 1.067  & 1.065  \\
        PnL per trade   &4.628  & 4.996 & 4.63   & 5.676  & 4.596  & 4.856  & 4.648  & 4.881  \\
        Sharpe ratio &0.778  & 0.839 & 0.778  & 0.954  & 0.772  & 0.816  & 0.781  & 0.82   \\
        P-value     &0.001  & 0     & 0.001  & 0      & 0.001  & 0      & 0.001  & 0     \\
        \midrule
        \midrule
\end{tabular}
\end{table}

\begin{table}[htbp]
  \centering
    \begin{tabular}{p{3.7cm}p{1cm}p{1cm}p{1cm}p{1cm}|p{1cm}p{1cm}p{1cm}p{1cm}}
        \midrule
        \multicolumn{1}{c}{\textbf{KM\_Med}}  &\multicolumn{4}{c}{\textbf{$G_\beta$ strategy}}  &  \multicolumn{4}{c}{\textbf{$D_\alpha$ strategy}}  \\
        \cmidrule(lr){2-5} \cmidrule(lr){6-9}
         \textbf{$\alpha$} &  0.7 & 0.75 & 0.8 & 0.85 &  0.7 & 0.75 & 0.8 & 0.85 \\
        \midrule
        E[Returns]       &0.115  & 0.118  & 0.121 & 0.138  & 0.094  & 0.095 & 0.086  & 0.1    \\
        Volatility        &0.15   & 0.15   & 0.15  & 0.15   & 0.15   & 0.15  & 0.15   & 0.15   \\
        Downside deviation      &0.103  & 0.101  & 0.099 & 0.097  & 0.109  & 0.108 & 0.109  & 0.108  \\
        Maximum drawdown    &-0.235 & -0.215 & -0.22 & -0.177 & -0.198 & -0.21 & -0.286 & -0.259 \\
        Sortino ratio  &1.115  & 1.166  & 1.225 & 1.428  & 0.866  & 0.876 & 0.788  & 0.925  \\ 
        Calmar ratio       &0.487  & 0.548  & 0.549 & 0.781  & 0.476  & 0.452 & 0.299  & 0.386  \\
        Hit rate        &0.517  & 0.51   & 0.509 & 0.517  & 0.515  & 0.513 & 0.512  & 0.514  \\
        Avg. profit / avg. loss   &1.074  & 1.107  & 1.113 & 1.101  & 1.058  & 1.066 & 1.058  & 1.068  \\
        PnL per trade   &4.545  & 4.672  & 4.796 & 5.486  & 3.741  & 3.765 & 3.397  & 3.97   \\
        Sharpe ratio &0.764  & 0.785  & 0.806 & 0.922  & 0.629  & 0.632 & 0.571  & 0.667  \\
        P-value     &0.001  & 0      & 0     & 0      & 0.005  & 0.005 & 0.012  & 0.003 \\
        \midrule
        \midrule

        \multicolumn{1}{c}{\textbf{SP\_Mod}}  &\multicolumn{4}{c}{\textbf{$G_\beta$ strategy}}  &  \multicolumn{4}{c}{\textbf{$D_\alpha$ strategy}}  \\
        \cmidrule(lr){2-5} \cmidrule(lr){6-9}
         \textbf{$\alpha$} &  0.7 & 0.75 & 0.8 & 0.85 &  0.7 & 0.75 & 0.8 & 0.85 \\
        \midrule
        E[Returns]       &0.137  & 0.127  & 0.129  & 0.129  & 0.139  & 0.129  & 0.129 & 0.123  \\
        Volatility        &0.15   & 0.15   & 0.15   & 0.15   & 0.15   & 0.15   & 0.15  & 0.15   \\
        Downside deviation      &0.102  & 0.103  & 0.103  & 0.102  & 0.105  & 0.106  & 0.105 & 0.105  \\
        Maximum drawdown    &-0.199 & -0.214 & -0.206 & -0.162 & -0.276 & -0.283 & -0.28 & -0.256 \\
        Sortino ratio  &1.346  & 1.237  & 1.258  & 1.267  & 1.333  & 1.222  & 1.226 & 1.18   \\
        Calmar ratio       &0.691  & 0.594  & 0.628  & 0.798  & 0.505  & 0.456  & 0.46  & 0.482  \\
        Hit rate        &0.523  & 0.516  & 0.522  & 0.522  & 0.524  & 0.522  & 0.52  & 0.521  \\
        Avg. profit / avg. loss   &1.076  & 1.091  & 1.069  & 1.07   & 1.074  & 1.072  & 1.079 & 1.068  \\
        PnL per trade   &5.456  & 5.041  & 5.135  & 5.131  & 5.532  & 5.121  & 5.115 & 4.895  \\
        Sharpe ratio &0.917  & 0.847  & 0.863  & 0.862  & 0.929  & 0.86   & 0.859 & 0.822  \\
        P-value     &0      & 0      & 0      & 0      & 0      & 0      & 0     & 0     \\
        \midrule
        \midrule

        \multicolumn{1}{c}{\textbf{SP\_Med}}  &\multicolumn{4}{c}{\textbf{$G_\beta$ strategy}}  &  \multicolumn{4}{c}{\textbf{$D_\alpha$ strategy}}  \\
        \cmidrule(lr){2-5} \cmidrule(lr){6-9}
         \textbf{$\alpha$} &  0.7 & 0.75 & 0.8 & 0.85 &  0.7 & 0.75 & 0.8 & 0.85 \\
        \midrule
        E[Returns]       &0.115  & 0.104  & 0.098  & 0.084  & 0.142  & 0.134  & 0.134  & 0.13   \\
        Volatility        &0.15   & 0.15  & 0.15   & 0.15   & 0.15   & 0.15   & 0.15   & 0.15   \\
        Downside deviation      &0.105  & 0.105  & 0.106  & 0.105  & 0.104  & 0.105  & 0.105  & 0.104  \\
        Maximum drawdown    &-0.279 & -0.287 & -0.271 & -0.256 & -0.221 & -0.227 & -0.217 & -0.215 \\
        Sortino ratio  &1.099  & 0.988  & 0.927  & 0.794  & 1.367  & 1.269  & 1.273  & 1.253  \\
        Calmar ratio       &0.414  & 0.362  & 0.363  & 0.327  & 0.644  & 0.59   & 0.617  & 0.604  \\
        Hit rate        &0.52   & 0.519  & 0.518  & 0.513  & 0.527  & 0.525  & 0.526  & 0.525  \\
        Avg. profit / avg. loss   &1.059  & 1.051  & 1.046  & 1.051  & 1.067  & 1.065  & 1.06   & 1.059  \\
        PnL per trade   &4.579  & 4.119  & 3.904  & 3.321  & 5.648  & 5.312  & 5.316  & 5.15   \\
        Sharpe ratio &0.769  & 0.692  & 0.656  & 0.558  & 0.949  & 0.892  & 0.893  & 0.865  \\
        P-value     &0.001  & 0.002  & 0.004  & 0.014  & 0      & 0      & 0      & 0    \\

\bottomrule 
\end{tabular}
\end{table}

\begin{table}[htbp]
  \centering
  \caption{ETF data set: robustness analysis for $\alpha$ - rescaled to target volatility. The experiment has been set with the values $p = 7$, and $\delta = 7$.}
  \label{tab:etf_robust}
    \begin{tabular}{p{3.7cm}p{1cm}p{1cm}|p{1cm}p{1cm}}
        \toprule
        \multicolumn{1}{c}{\textbf{CCF}}  &\multicolumn{2}{c}{\textbf{$G_\beta$ strategy}}  &  \multicolumn{2}{c}{\textbf{$D_\alpha$ strategy}}  \\
        \cmidrule(lr){2-3} \cmidrule(lr){4-5}
         \textbf{$\alpha$} & 0.75 & 0.85 & 0.75 & 0.85 \\
        \midrule
        E[Returns]       & -0.019 & -0.038 & 0.056  & 0.041  \\
        Volatility        & 0.15   & 0.15   & 0.15   & 0.15   \\
        Downside deviation      &0.116  & 0.116  & 0.097  & 0.097  \\
        Maximum drawdown    &-0.668 & -0.672 & -0.362 & -0.382 \\
        Sortino ratio  &  -0.165 & -0.324 & 0.581  & 0.422  \\
        Calmar ratio       &-0.029 & -0.056 & 0.156  & 0.107  \\
        Hit rate        & 0.492  & 0.501  & 0.504  & 0.494  \\
        Avg. profit / avg. loss   & 1.006  & 0.949  & 1.056  & 1.077  \\
        PnL per trade   &-0.76  & -1.49  & 2.234  & 1.617  \\
        Sharpe ratio &-0.128 & -0.25  & 0.375  & 0.272  \\
        P-value     &0.644  & 0.365  & 0.169  & 0.321  \\
\midrule
\midrule

        \multicolumn{1}{c}{\textbf{KM\_Mod}}  &\multicolumn{2}{c}{\textbf{$G_\beta$ strategy}}  &  \multicolumn{2}{c}{\textbf{$D_\alpha$ strategy}}  \\
        \cmidrule(lr){2-3} \cmidrule(lr){4-5}
         \textbf{$\alpha$} & 0.75 & 0.85 & 0.75 & 0.85 \\
        \midrule
        E[Returns]       &0.02   & -0.06  & 0.065  & 0.083  \\
        Volatility        & 0.15   & 0.15   & 0.15   & 0.15   \\
        Downside deviation      &0.115  & 0.122  & 0.097  & 0.096  \\
        Maximum drawdown    &-0.525 & -0.707 & -0.267 & -0.292 \\
        Sortino ratio  &0.176  & -0.487 & 0.676  & 0.86   \\
        Calmar ratio       &0.038  & -0.084 & 0.245  & 0.283  \\
        Hit rate        &0.512  & 0.505  & 0.502  & 0.505  \\
        Avg. profit / avg. loss   & 0.978  & 0.911  & 1.077  & 1.086  \\
        PnL per trade   & 0.8    & -2.363 & 2.591  & 3.28   \\
        Sharpe ratio & 0.134  & -0.397 & 0.435  & 0.551  \\
        P-value     & 0.628  & 0.147  & 0.105  & 0.039  \\

        \midrule
        \midrule
        \multicolumn{1}{c}{\textbf{KM\_Med}}  &\multicolumn{2}{c}{\textbf{$G_\beta$ strategy}}  &  \multicolumn{2}{c}{\textbf{$D_\alpha$ strategy}}  \\
        \cmidrule(lr){2-3} \cmidrule(lr){4-5}
         \textbf{$\alpha$} & 0.75 & 0.85 & 0.75 & 0.85 \\
        \midrule
        E[Returns]       &0.022  & 0.004  & 0.022  & 0.052  \\ 
        Volatility        &0.15   & 0.15   & 0.15   & 0.15   \\
        Downside deviation      &0.116  & 0.113  & 0.105  & 0.104  \\
        Maximum drawdown    &-0.465 & -0.436 & -0.385 & -0.336 \\
        Sortino ratio  &0.188  & 0.031  & 0.213  & 0.504  \\
        Calmar ratio       &0.047  & 0.008  & 0.058  & 0.155  \\
        Hit rate        &0.499  & 0.505  & 0.5    & 0.502  \\
        Avg. profit / avg. loss   &1.034  & 0.984  & 1.027  & 1.06   \\
        PnL per trade   &0.87   & 0.14   & 0.889  & 2.071  \\
        Sharpe ratio &0.146  & 0.024  & 0.149  & 0.348  \\
        P-value     &0.598  & 0.932  & 0.588  & 0.205 \\
        \midrule
        \midrule
        \multicolumn{1}{c}{\textbf{SP\_Mod}}  &\multicolumn{2}{c}{\textbf{$G_\beta$ strategy}}  &  \multicolumn{2}{c}{\textbf{$D_\alpha$ strategy}}  \\
        \cmidrule(lr){2-3} \cmidrule(lr){4-5}
         \textbf{$\alpha$} & 0.75 & 0.85 & 0.75 & 0.85 \\
        \midrule
        E[Returns]       &-0.005 & -0.076 & 0.073  & 0.12   \\
        Volatility        &0.15   & 0.15   & 0.15   & 0.15   \\
        Downside deviation      &0.113  & 0.118  & 0.1    & 0.095  \\
        Maximum drawdown    &-0.579 & -0.726 & -0.272 & -0.258 \\
        Sortino ratio  &-0.043 & -0.648 & 0.728  & 1.259  \\ 
        Calmar ratio       &-0.008 & -0.105 & 0.269  & 0.464  \\
        Hit rate        &0.503  & 0.497  & 0.51   & 0.511  \\ 
        Avg. profit / avg. loss   &0.983  & 0.918  & 1.05   & 1.107  \\
        PnL per trade   &-0.191 & -3.035 & 2.901  & 4.747  \\
        Sharpe ratio &-0.032 & -0.51  & 0.487  & 0.797  \\
        P-value     &0.908  & 0.064  & 0.074  & 0.003  \\
        \midrule
        \midrule
\end{tabular}
\end{table}

\begin{table}[htbp]
  \centering
    \begin{tabular}{p{3.7cm}p{1cm}p{1cm}|p{1cm}p{1cm}}
        \toprule
        \multicolumn{1}{c}{\textbf{SP\_Med}}  &\multicolumn{2}{c}{\textbf{$G_\beta$ strategy}}  &  \multicolumn{2}{c}{\textbf{$D_\alpha$ strategy}}  \\
        \cmidrule(lr){2-3} \cmidrule(lr){4-5}
         \textbf{$\alpha$} & 0.75 & 0.85 & 0.75 & 0.85 \\
        \midrule
        E[Returns]       &0.013  & -0.025 & 0.081  & 0.08   \\
        Volatility        &0.15   & 0.15   & 0.15   & 0.15   \\
        Downside deviation      &0.108  & 0.11   & 0.105  & 0.103  \\
        Maximum drawdown    &-0.369 & -0.452 & -0.382 & -0.348 \\
        Sortino ratio  &0.122  & -0.232 & 0.771  & 0.779  \\  
        Calmar ratio       &0.036  & -0.056 & 0.211  & 0.23   \\
        Hit rate        &0.5    & 0.495  & 0.504  & 0.51   \\
        Avg. profit / avg. loss   &1.016  & 0.987  & 1.087  & 1.061  \\
        PnL per trade   &0.527  & -1.01  & 3.2    & 3.18   \\
        Sharpe ratio &0.088  & -0.17  & 0.538  & 0.534  \\
        P-value     &0.749  & 0.54   & 0.051  & 0.051 \\
\bottomrule

    \end{tabular}

\end{table}

\begin{table}[htbp]
  \centering
  \caption{Futures data set: robustness analysis for $\alpha$ - rescaled to target volatility. The experiment has been set with the values $p = 7$, and $\delta = 7$.}
  \label{tab:futures_robust}
    \begin{tabular}{p{3.7cm}p{1cm}p{1cm}p{1cm}p{1cm}|p{1cm}p{1cm}p{1cm}p{1cm}}
        \toprule
        \multicolumn{1}{c}{\textbf{CCF}}  &\multicolumn{4}{c}{\textbf{$G_\beta$ strategy}}  &  \multicolumn{4}{c}{\textbf{$D_\alpha$ strategy}}  \\
        \cmidrule(lr){2-5} \cmidrule(lr){6-9}
         \textbf{$\alpha$} &  0.7 & 0.75 & 0.8 & 0.85 &  0.7 & 0.75 & 0.8 & 0.85 \\
        \midrule
        E[Returns]       &0.011  & 0.013  & 0.006  & -0.025 & 0.053  & 0.061  & 0.064 & 0.045  \\
        Volatility        &0.15   & 0.15   & 0.15   & 0.15   & 0.15   & 0.15   & 0.15  & 0.15   \\
        Downside deviation      &0.104  & 0.104  & 0.105  & 0.107  & 0.107  & 0.107  & 0.106 & 0.107  \\
        Maximum drawdown    &-0.594 & -0.535 & -0.545 & -0.667 & -0.432 & -0.393 & -0.38 & -0.426 \\
        Sortino ratio  &0.102  & 0.121  & 0.06   & -0.236 & 0.496  & 0.574  & 0.603 & 0.422  \\
        Calmar ratio       &0.018  & 0.024  & 0.012  & -0.038 & 0.123  & 0.156  & 0.169 & 0.106  \\
        Hit rate        &0.501  & 0.502  & 0.504  & 0.5    & 0.509  & 0.511  & 0.503 & 0.504  \\
        Avg. profit / avg. loss   &1.008  & 1.007  & 0.993  & 0.971  & 1.031  & 1.032  & 1.072 & 1.042  \\
        PnL per trade   &0.421  & 0.499  & 0.249  & -1.001 & 2.116  & 2.439  & 2.548 & 1.793  \\
        Sharpe ratio &0.071  & 0.084  & 0.042  & -0.168 & 0.355  & 0.41   & 0.428 & 0.301  \\
        P-value     &0.749  & 0.705  & 0.85   & 0.447  & 0.108  & 0.064  & 0.053 & 0.173 \\
\midrule
\midrule
        
        \multicolumn{1}{c}{\textbf{KM\_Mod}}  &\multicolumn{4}{c}{\textbf{$G_\beta$ strategy}}  &  \multicolumn{4}{c}{\textbf{$D_\alpha$ strategy}}  \\
        \cmidrule(lr){2-5} \cmidrule(lr){6-9}
         \textbf{$\alpha$} &  0.7 & 0.75 & 0.8 & 0.85 &  0.7 & 0.75 & 0.8 & 0.85 \\
        \midrule
        E[Returns]       &0.006  & -0.001 & 0.006  & 0.006  & 0.048  & 0.036  & 0.038 & 0.025  \\ 
        Volatility        &0.15   & 0.15   & 0.15   & 0.15   & 0.15   & 0.15   & 0.15  & 0.15   \\
        Downside deviation      &0.103  & 0.102  & 0.1    & 0.099  & 0.108  & 0.108  & 0.108 & 0.108  \\
        Maximum drawdown    &-0.484 & -0.587 & -0.517 & -0.505 & -0.326 & -0.411 & -0.35 & -0.452 \\
        Sortino ratio  &0.06   & -0.007 & 0.064  & 0.059  & 0.446  & 0.336  & 0.352 & 0.234  \\
        Calmar ratio       &0.013  & -0.001 & 0.012  & 0.011  & 0.147  & 0.088  & 0.109 & 0.056  \\
        Hit rate        &0.503  & 0.498  & 0.501  & 0.496  & 0.501  & 0.502  & 0.502 & 0.496  \\
        Avg. profit / avg. loss   &0.996  & 1.009  & 1.004  & 1.025  & 1.059  & 1.038  & 1.039 & 1.05   \\
        PnL per trade   &0.246  & -0.028 & 0.254  & 0.23   & 1.901  & 1.441  & 1.514 & 1.004  \\
        Sharpe ratio &0.041  & -0.005 & 0.043  & 0.039  & 0.319  & 0.242  & 0.254 & 0.169  \\
        P-value     &0.852  & 0.983  & 0.847  & 0.862  & 0.148  & 0.274  & 0.25  & 0.446 \\ \midrule
        \midrule
\end{tabular}
\end{table}

\begin{table}[htbp]
  \centering
  
    \begin{tabular}{p{3.7cm}p{1cm}p{1cm}p{1cm}p{1cm}|p{1cm}p{1cm}p{1cm}p{1cm}}
        \toprule

        \multicolumn{1}{c}{\textbf{KM\_Med}}  &\multicolumn{4}{c}{\textbf{$G_\beta$ strategy}}  &  \multicolumn{4}{c}{\textbf{$D_\alpha$ strategy}}  \\
        \cmidrule(lr){2-5} \cmidrule(lr){6-9}
         \textbf{$\alpha$} &  0.7 & 0.75 & 0.8 & 0.85 &  0.7 & 0.75 & 0.8 & 0.85 \\
        \midrule
        E[Returns]       &0.004  & 0.005  & 0.009  & 0.017  & 0.031  & 0.036  & 0.037  & 0.03   \\
        Volatility        &0.15   & 0.15   & 0.15   & 0.15   & 0.15   & 0.15   & 0.15   & 0.15   \\
        Downside deviation      &0.103  & 0.103  & 0.102  & 0.102  & 0.108  & 0.108  & 0.108  & 0.108  \\
        Maximum drawdown    &-0.551 & -0.534 & -0.479 & -0.454 & -0.414 & -0.399 & -0.434 & -0.459 \\
        Sortino ratio  &0.042  & 0.045  & 0.09   & 0.163  & 0.284  & 0.335  & 0.343  & 0.28   \\ 
        Calmar ratio       &0.008  & 0.009  & 0.019  & 0.037  & 0.074  & 0.091  & 0.085  & 0.066  \\
        Hit rate        &0.496  & 0.495  & 0.495  & 0.498  & 0.501  & 0.499  & 0.5    & 0.498  \\
        Avg. profit / avg. loss   &1.023  & 1.028  & 1.03   & 1.028  & 1.036  & 1.05   & 1.047  & 1.046  \\
        PnL per trade   &0.172  & 0.185  & 0.363  & 0.661  & 1.217  & 1.443  & 1.472  & 1.202  \\
        Sharpe ratio &0.029  & 0.031  & 0.061  & 0.111  & 0.204  & 0.242  & 0.247  & 0.202  \\
        P-value     &0.896  & 0.888  & 0.783  & 0.616  & 0.356  & 0.273  & 0.264  & 0.362 \\        
        \midrule
        \midrule
        \multicolumn{1}{c}{\textbf{SP\_Mod}}  &\multicolumn{4}{c}{\textbf{$G_\beta$ strategy}}  &  \multicolumn{4}{c}{\textbf{$D_\alpha$ strategy}}  \\
        \cmidrule(lr){2-5} \cmidrule(lr){6-9}
         \textbf{$\alpha$} &  0.7 & 0.75 & 0.8 & 0.85 &  0.7 & 0.75 & 0.8 & 0.85 \\
        \midrule
        E[Returns]       &-0.011 & -0.01  & 0     & -0.007 & 0.033  & 0.027  & 0.029  & 0.031  \\
        Volatility        &0.15   & 0.15   & 0.15  & 0.15   & 0.15   & 0.15   & 0.15   & 0.15   \\
        Downside deviation      &0.103  & 0.102  & 0.102 & 0.102  & 0.108  & 0.108  & 0.108  & 0.108  \\
        Maximum drawdown    &-0.528 & -0.508 & -0.43 & -0.504 & -0.465 & -0.474 & -0.467 & -0.502 \\
        Sortino ratio  &-0.106 & -0.097 & 0.002 & -0.073 & 0.309  & 0.253  & 0.271  & 0.283  \\ 
        Calmar ratio       &-0.021 & -0.019 & 0     & -0.015 & 0.072  & 0.057  & 0.062  & 0.061  \\
        Hit rate        &0.496  & 0.494  & 0.495 & 0.502  & 0.5    & 0.498  & 0.5    & 0.503  \\
        Avg. profit / avg. loss   &1.004  & 1.013  & 1.021 & 0.984  & 1.044  & 1.043  & 1.037  & 1.027  \\
        PnL per trade   &-0.435 & -0.392 & 0.007 & -0.295 & 1.322  & 1.079  & 1.158  & 1.213  \\
        Sharpe ratio &-0.073 & -0.066 & 0.001 & -0.05  & 0.222  & 0.181  & 0.195  & 0.204  \\
        P-value     &0.742  & 0.766  & 0.996 & 0.823  & 0.315  & 0.412  & 0.379  & 0.357 \\        
        \midrule
        \midrule
        \multicolumn{1}{c}{\textbf{SP\_Med}}  &\multicolumn{4}{c}{\textbf{$G_\beta$ strategy}}  &  \multicolumn{4}{c}{\textbf{$D_\alpha$ strategy}}  \\
        \cmidrule(lr){2-5} \cmidrule(lr){6-9}
         \textbf{$\alpha$} &  0.7 & 0.75 & 0.8 & 0.85 &  0.7 & 0.75 & 0.8 & 0.85 \\
        \midrule
        E[Returns]       &0.007  & 0.022  & 0.02   & 0.004 & 0.03   & 0.031  & 0.035  & 0.036 \\
        Volatility        &0.15   & 0.15   & 0.15   & 0.15  & 0.15   & 0.15   & 0.15   & 0.15  \\
        Downside deviation      &0.105  & 0.105  & 0.103  & 0.104 & 0.107  & 0.107  & 0.107  & 0.107 \\
        Maximum drawdown    &-0.467 & -0.457 & -0.455 & -0.44 & -0.433 & -0.462 & -0.504 & -0.44 \\
        Sortino ratio  &0.071  & 0.209  & 0.191  & 0.039 & 0.282  & 0.29   & 0.323  & 0.333 \\ 
        Calmar ratio       &0.016  & 0.048  & 0.044  & 0.009 & 0.069  & 0.067  & 0.069  & 0.081 \\
        Hit rate        &0.503  & 0.508  & 0.497  & 0.492 & 0.503  & 0.501  & 0.501  & 0.501 \\
        Avg. profit / avg. loss   &0.999  & 0.994  & 1.036  & 1.037 & 1.024  & 1.037  & 1.04   & 1.04  \\
        PnL per trade   &0.295  & 0.869  & 0.786  & 0.162 & 1.192  & 1.23   & 1.371  & 1.417 \\
        Sharpe ratio &0.05   & 0.146  & 0.132  & 0.027 & 0.2    & 0.207  & 0.23   & 0.238 \\
        P-value     &0.823  & 0.51   & 0.551  & 0.902 & 0.365  & 0.35   & 0.298  & 0.282\\

\bottomrule

    \end{tabular}
\end{table}

\clearpage
\newpage
\section{CO2 emissions data}
\label{sec: CO2}

\subsection{Data description}
We acquire CO2 emissions data (metric tons per capita) for the period 1990–2019 from Climate Watch. This data is depicted in Figure \ref{fig:CO2_countries} in the appendix, and Table \ref{tab: country_code} depicts the 31 countries' names and corresponding codes.


\begin{table}[htbp]
  \centering
  \caption{31 Countries and their corresponding codes.}
    \begin{tabular}{p{1.7cm}p{0.8cm}|p{2.4cm}p{0.8cm}|p{1.7cm}p{0.8cm}|p{1.7cm}p{0.8cm}}
        \toprule
        \textbf{Country} & \textbf{Code} & \textbf{Country} & \textbf{Code} & \textbf{Country} & \textbf{Code} & \textbf{Country} & \textbf{Code} \\
        \midrule
        Austria & AUT & Belgium & BEL & Bulgaria & BGR & Croatia & HRV\\ 
        Cyprus & CYP & Czech Republic & CZE & Denmark & DNK & Estonia & EST \\  
        Finland & FIN & France & FRA & Germany & DEU & Greece & GRC \\ 
        Hungary & HUN & Iceland & ISL & Ireland & IRL & Italy & ITA \\ 
        Latvia & LVA & Lithuania & LTU & Luxemburg & LUX & Malta & MLT \\ 
        Netherlands & NLD & Norway & NOR & Poland & POL & Portugal & PRT \\ 
        Romania & ROU & Slovakia & SVK & Slovenia & SVN & Spain & ESP \\ 
        Sweden & SWE & Switzerland & CHE & UK & GBR \\ 
        \bottomrule
    \end{tabular}

\label{tab: country_code}
\end{table}

\subsection{Pipeline}

Given that $n=31$ for the CO2 time series with a period of $T=30$ years, we extract STS from the data with length $q = 16$ via a sliding window that is shifted by $s=1$. Next, we apply SP\_Med to cluster the STS, and calculate the lead-lag matrix $\Gamma_{n \times n}$ using a voting threshold ($\theta = 3$). From this, we rank the time series from the most leading to the most lagging by using RowSum ranking.

\subsection{Results}

We find that the first among the leading countries is Poland, while there are five countries that are tied for second place. These countries are Sweden, Belgium, Slovakia, Denmark and Germany. Moreover, we find that the last two lagging countries are Bulgaria and Estonia. We summarize the top ten leading countries, and the last ten lagging countries in Tables [\ref{tab: top10}, \ref{tab: last10}]. 

By investigating the top-ranked countries Poland and Sweden, we can attempt to explain the trend in these countries' CO2 emission rates. We note that these countries belong to the same political group, the European Union. Thus, one possibility is that they follow similar protocols in terms of regulations and restrictions. These restrictions and regulations will help alleviate CO2 emissions.

The combustion of fossil fuels such as gasoline, natural gas and coal represents one of the main contributors to carbon dioxide pollution. \cite{zuk2021coal} suggested that indigenous coal is the primary source of mixed energy in Poland, and since the country joined the European Union in 2004, it has become clear that Poland will do much to protect its domestic coal sector, thus, will reject demands for aggressive harmonized decarbonization efforts, despite being identified as a leader with rapid emissions reduction taking place in the 1990s and early 2000s. At the same time, \cite{kolasa2015stepwise} proposed that the Polish energy sector is actively participating in efforts to reduce emissions of greenhouse gases (GHG) into the environment by lowering the quantity of coal in the fuel mix and increasing the usage of renewable energy sources. Also, \cite{rybak2022impact} posited that both Poland and Sweden are the leading countries that effectively reduce GHG and apply environmental taxes. 

We note that since joining the EU in 1995, Sweden experienced a rapid drop in CO2 emissions. In Sweden, \cite{de2021detecting} proffered that many incentives have been designed to encourage consumers to purchase cleaner automobiles that produce fewer CO2 emissions. Also, \cite{de2021detecting} further argued that the Swedish government introduced a new bonus-penalty system for the purpose of administering incentives and taxes on light vehicles. \cite{de2021detecting} suggested this is based on the strong leadership of Germany, because it is one of the first countries in the world to formulate an approach that proved favourable to the environment, particularly in the field of transportation. For example, the German government first implemented an Eco-tax in 1999, which led to increases in the price of fuel and parking fees, and a reduction in the amount of available parking space for vehicles. Indeed, Germany is a leading country in reducing CO2 emissions itself.

In contrast, we also perform an analysis of the two most lagging countries. In Estonia, the work of \cite{moora2017determination} demonstrated that shale oil is a low-grade carbonaceous fossil fuel, which is utilized locally as their main source of energy. Burning shale oil produces high CO2 emissions, which sets it apart from other forms of fossil fuel combustion. Moreover, the power sector appears to be the most profound emitter of CO2 and is responsible for significant amounts of waste. Bulgaria, which has a fleet of old cars ranging from 15 to 20 years, has the extremely difficult challenge of reducing automobile pollution in practice. Moreover, a study by \cite{hristov2019study} demonstrated that upgrading the transportation infrastructure does not always have a positive impact on the decrease in transportation emissions. Furthermore, the only native energy source is lignite, which has a low caloric content. Although the future of fossil fuels is uncertain after 2030, it should be noted that EU Regulation 2018/842 will not impose stringent limitations on GHG emissions in Bulgaria until 2030, as posited by \cite{vitkov2020greenhouse}. 

\vspace{0.5cm}
\begin{minipage}{\textwidth}
\begin{minipage}[b]{0.5\textwidth}
  \centering
  \captionof{table}{Top ten ranking of leading countries.}
    \begin{tabular}{lccccc}
        \toprule
          \textbf{Country}                & \textbf{Rank}  \\
        \midrule
            Poland & $1$ \\ 
            Sweden & $2$ \\  
            Belgium & $2$ \\
            Slovakia & $2$ \\ 
            Denmark & $2$ \\  
            Germany & $2$ \\
            Netherlands  & $7$ \\ 
            Hungary & $8$ \\  
            France  & $9$ \\
            UK & $10$ \\ 
        \bottomrule
    \end{tabular}

\label{tab: top10}

\end{minipage}
\hfill
\begin{minipage}[b]{0.49\textwidth}

  \centering
  \captionof{table}{Last ten ranking of lagging countries.}
    \begin{tabular}{lccccc}
        \toprule
          \textbf{Country}                & \textbf{Rank}  \\
        \midrule
            Austria & $22$ \\
            Slovenia & $23$ \\
            Croatia & $23$ \\
            Luxembourg& $25$ \\ 
            Romania  & $26$ \\
            Malta & $27$ \\ 
            Latvia  & $28$ \\ 
            Lithuania & $28$ \\
            Bulgaria & $30$ \\ 
            Estonia & $31$ \\ 
        \bottomrule
    \end{tabular}

\label{tab: last10}
\end{minipage}
\end{minipage}
\vspace{0.5cm}

\section{Conclusion}
\label{sec: conclusion}

We develop a clustering-driven methodology for robust detection of lead-lag relationships in high-dimensional multivariate time series, in the setting of lagged multi-factor models. We consider a collection of time series as input, and generate an expanded universe by extracting STS from each time series via a sliding window. Next, we employ several clustering approaches which include KM and SP clustering to cluster the resulting STS. After extracting the clusters, lead-lag estimates from the clusters are merged to help identify the consistent relationships that existed in the original universe. 

When applied to financial data sets, our proposed methods attain promising Sharpe ratios, and are statistically significant when compared to the benchmark. In addition to the financial domain, our methods can also be applied to other fields, such as environmental sciences, which is consistent with the results of others’ previous work. Thus, our method is generally applicable to a variety of multivariate time series data sets.
One possible future work is to fine-tune the weights of the portfolios of laggers. In our current work related to financial data, we buy or sell baskets of lagging assets based on equal weight according to the trading signal, while adjusting the weight of the portfolios of laggers based on the ranking of the laggers. In another direction of future work, we intend to experiment with other clustering methods applicable in our setting [\cite{cucuringu2020hermitian}, \cite{cucuringu2019sponge}], since our current work has only experimented with KM and SP clustering. Moreover, we also intend to test various alternative ranking algorithms [\cite{bradley1952rank}, \cite{fogel2014serialrank}, \cite{page1999pagerank}, \cite{cucuringu2016sync}, \cite{he2022gnnrank}, \cite{d2021ranking}], and consider momentum reversal strategies within the trading pipeline. Lastly, one could further explore the more challenging mixed membership model defined in Section \ref{sec: model}.



\newpage
\printbibliography

\newpage
\appendix
\section{Appendix} \label{appendix}

\subsection{Synthetic data experiment: spectral clustering}
\label{appendix_spectral_clustering}
Figures [\ref{fig:spectral_vote_matrix_before_plus_error_matrix_befores}, \ref{fig:spectral_vote_matrix_after_plus_error_matrix_afters}]
display the result of applying SP clustering, the voting matrix and the error matrix based on mode and median estimation without and with voting threshold with different $k = \{1,2,3\}$. 

\begin{table}[!htbp]
  \centering
    \begin{tabular}{p{4.5cm}|p{4.5cm}p{4.5cm}}
      \multicolumn{3}{c}{\textbf{Spectral clustering}} \vspace{0.3cm}\\
      
      \multicolumn{1}{c}{\textbf{Homogeneous Setting}}    &  \multicolumn{2}{c}{\textbf{Heterogeneous Setting}}  \\

      \includegraphics[width=\linewidth,trim=0cm 0cm 1.5cm 1cm,clip]{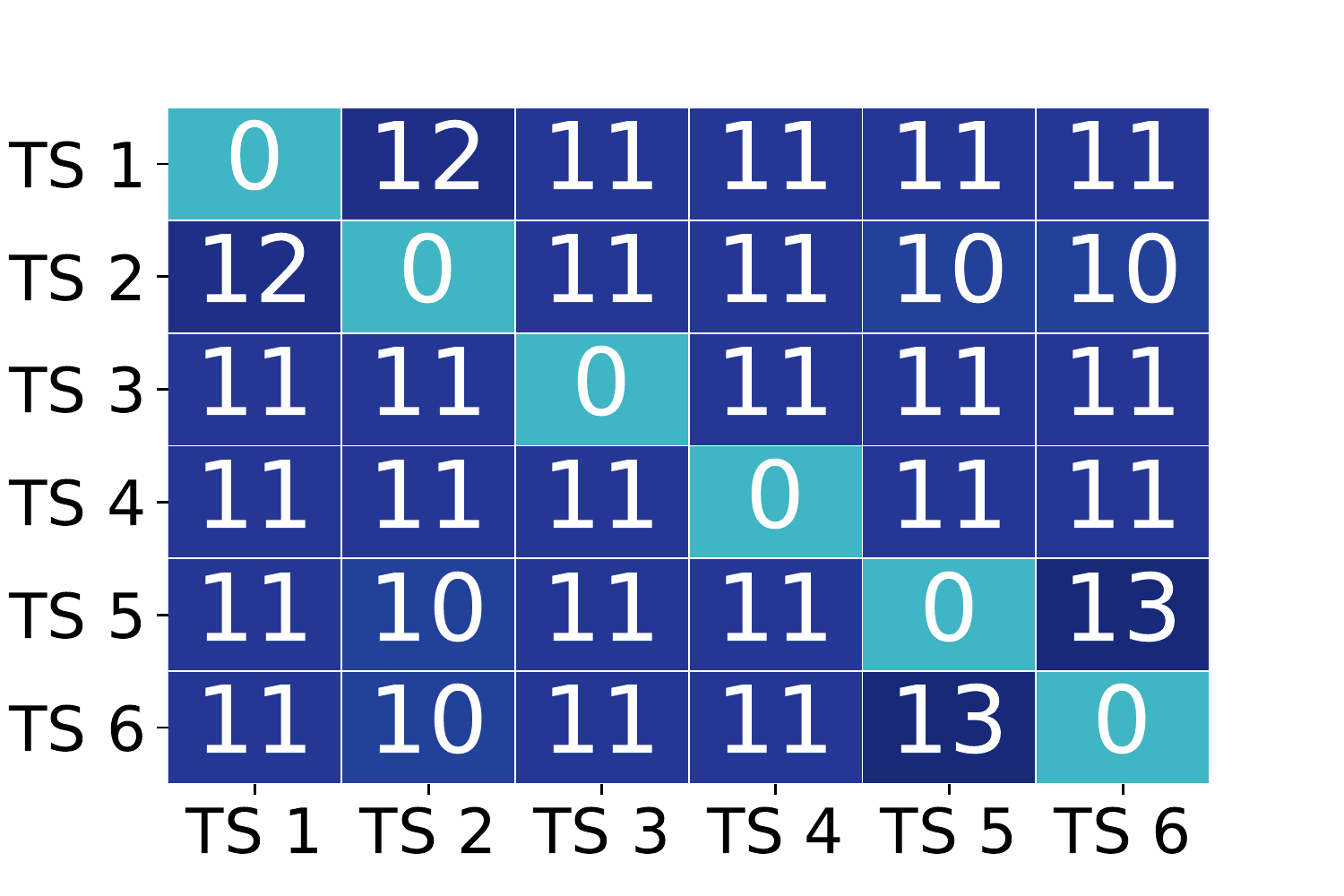}
      
      \includegraphics[width=\linewidth,trim=0cm 0cm 1.5cm 1cm,clip] {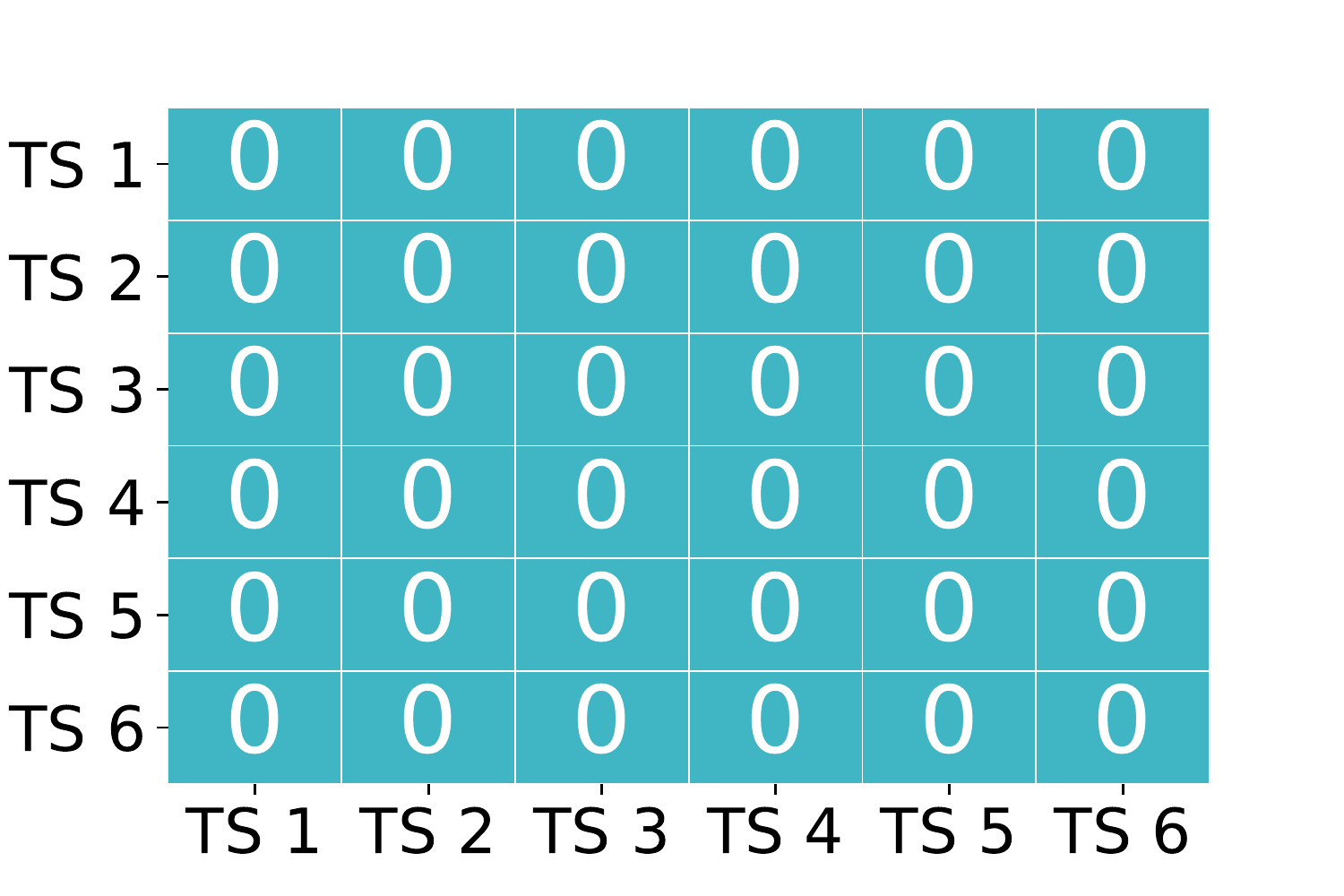}

    \includegraphics[width=\linewidth,trim=0cm 0cm 1.5cm 1cm,clip] {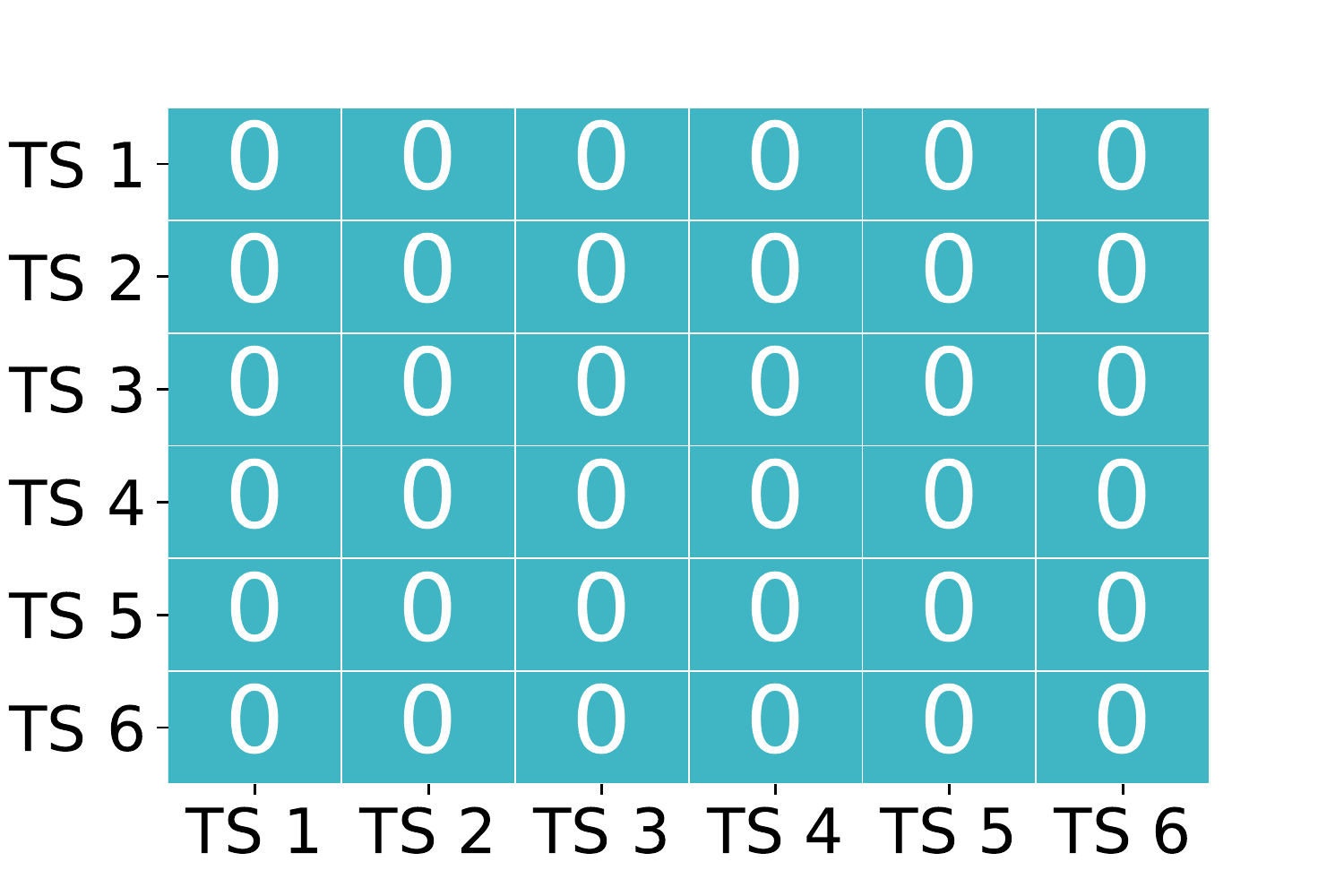}

    &
      \includegraphics[width=\linewidth,trim=0cm 0cm 1.5cm 1cm,clip]{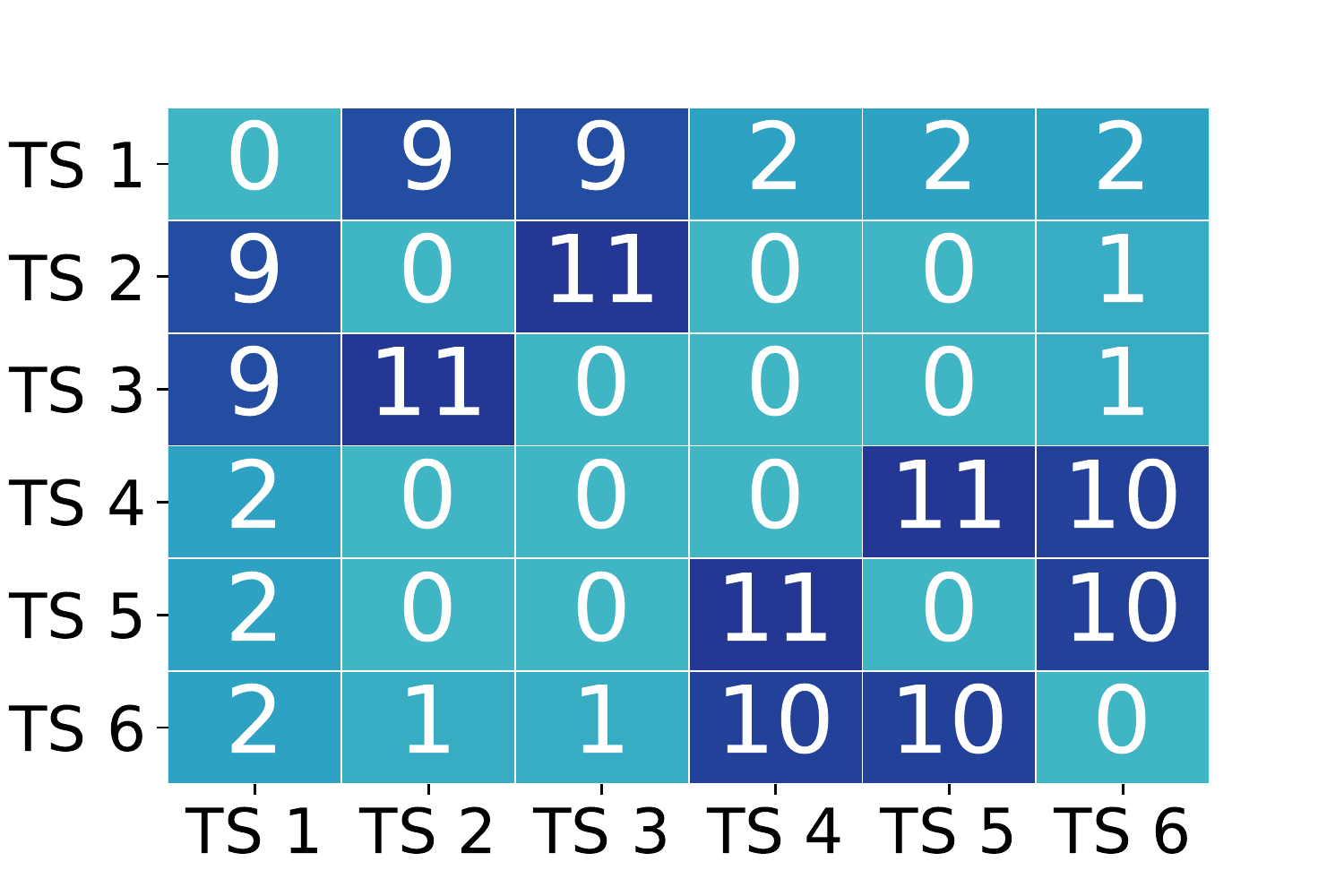}
      
      \includegraphics[width=\linewidth,trim=0cm 0cm 1.5cm 1cm,clip] {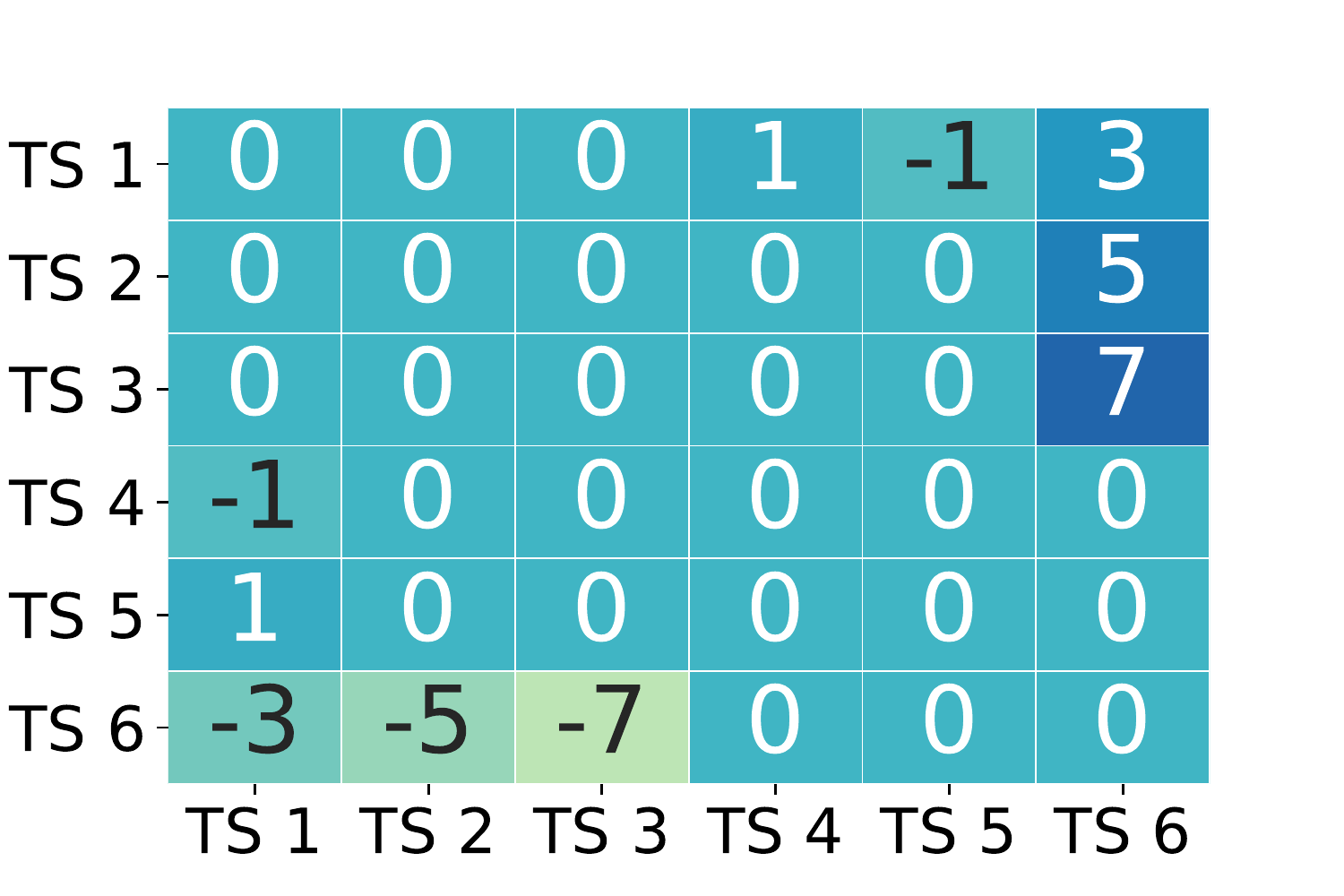}

    \includegraphics[width=\linewidth,trim=0cm 0cm 1.5cm 1cm,clip] {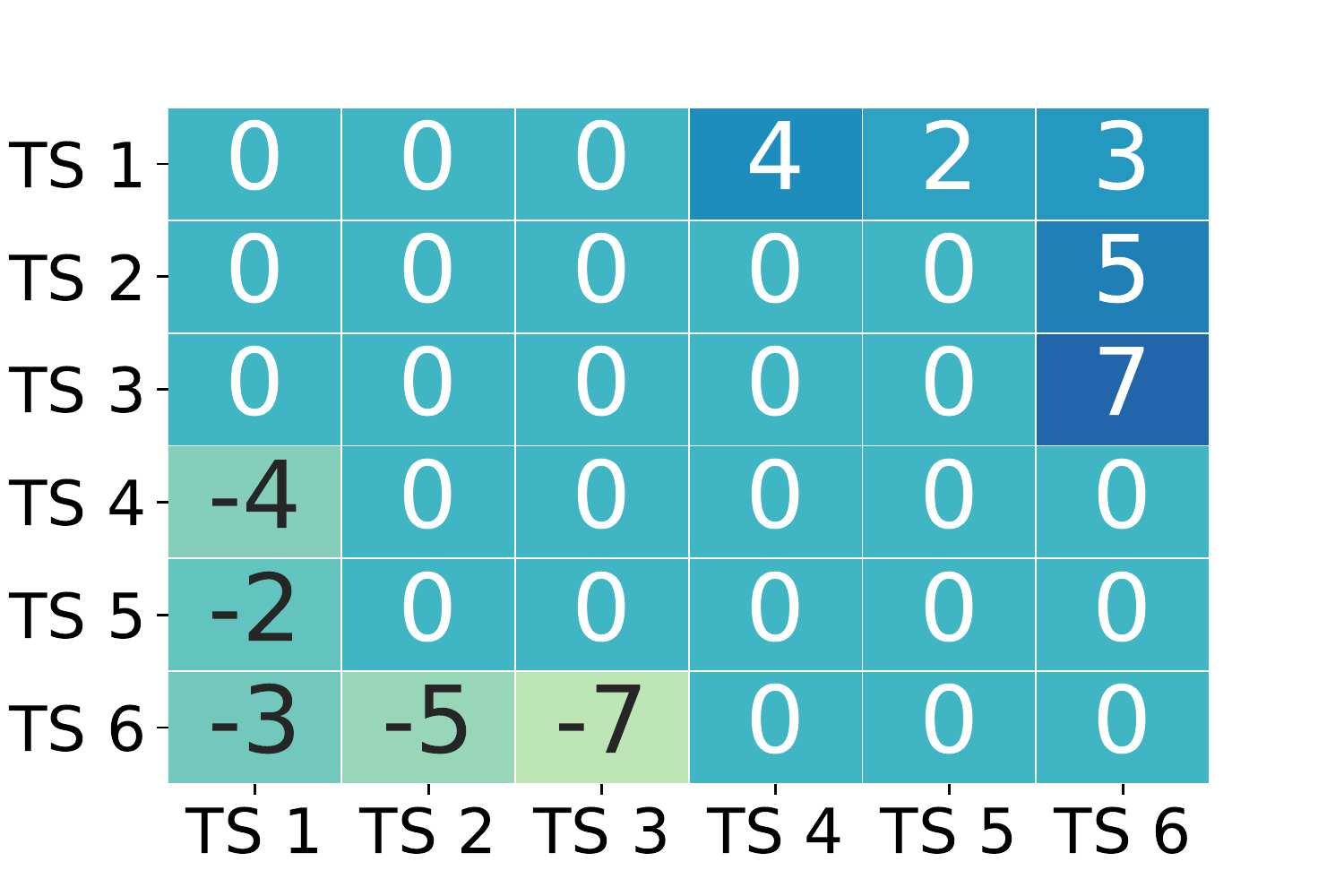}
    &
      \includegraphics[width=\linewidth,trim=0cm 0cm 1.5cm 1cm,clip]{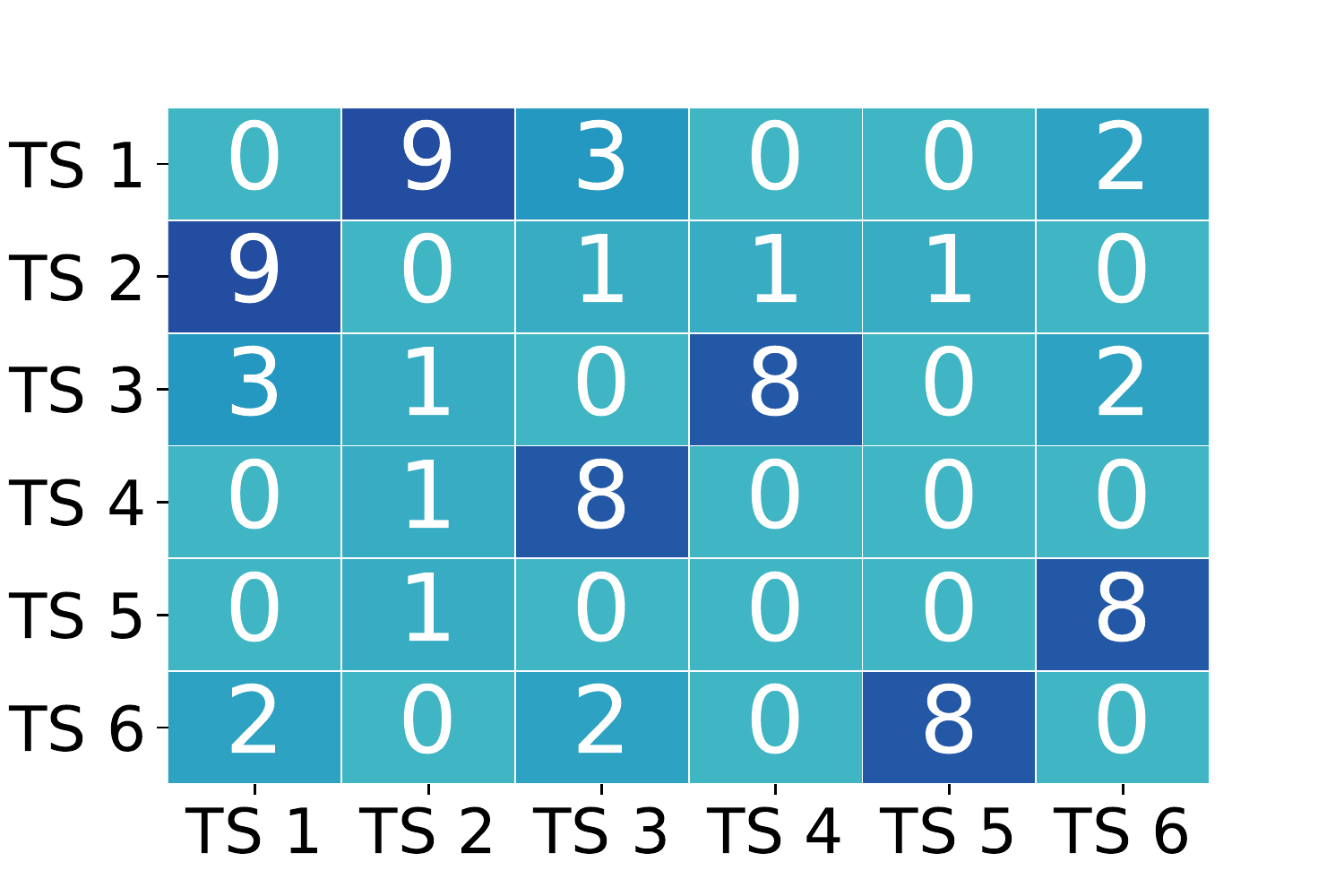}
      
      \includegraphics[width=\linewidth,trim=0cm 0cm 1.5cm 1cm,clip] {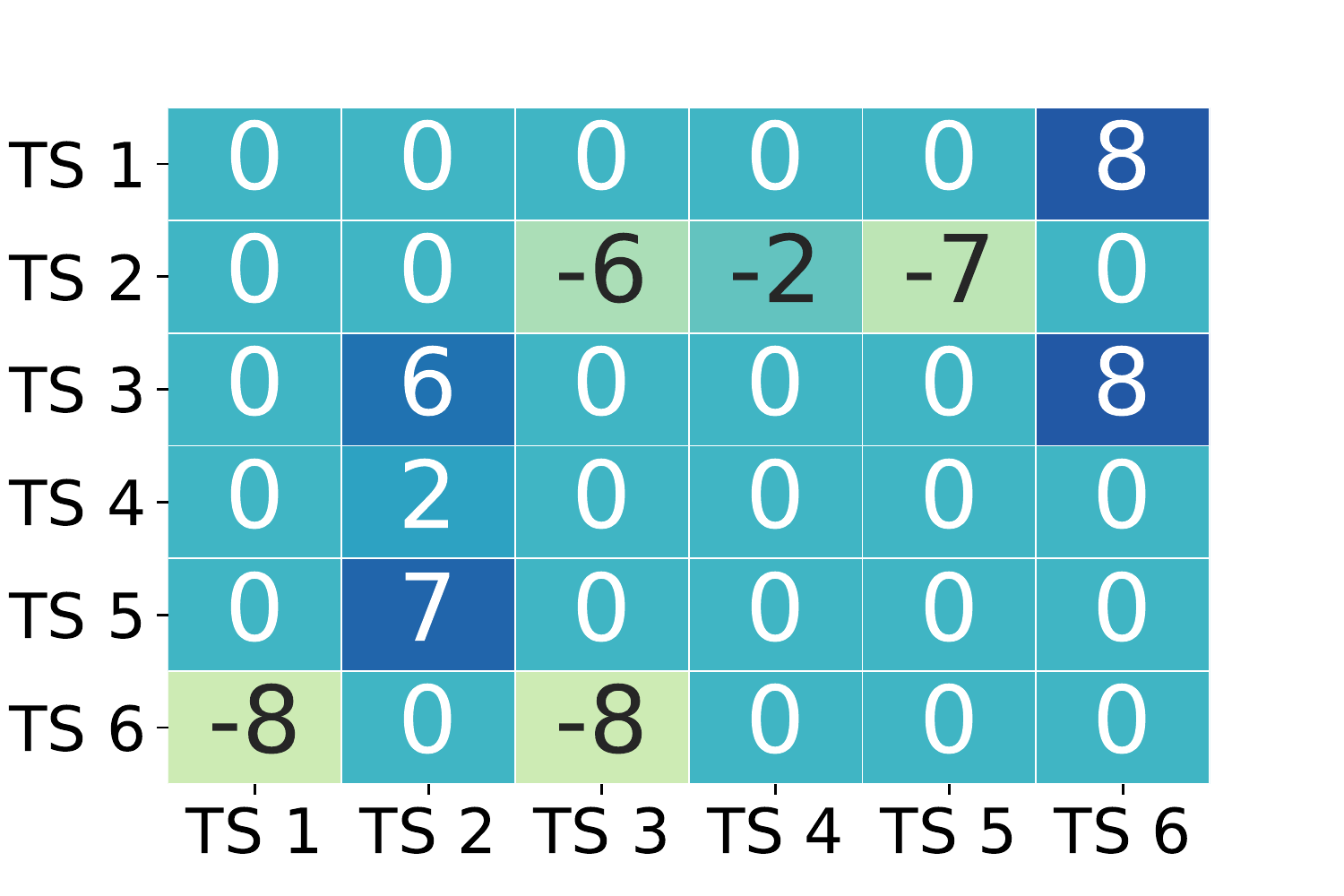}

      \includegraphics[width=\linewidth,trim=0cm 0cm 1.5cm 1cm,clip] {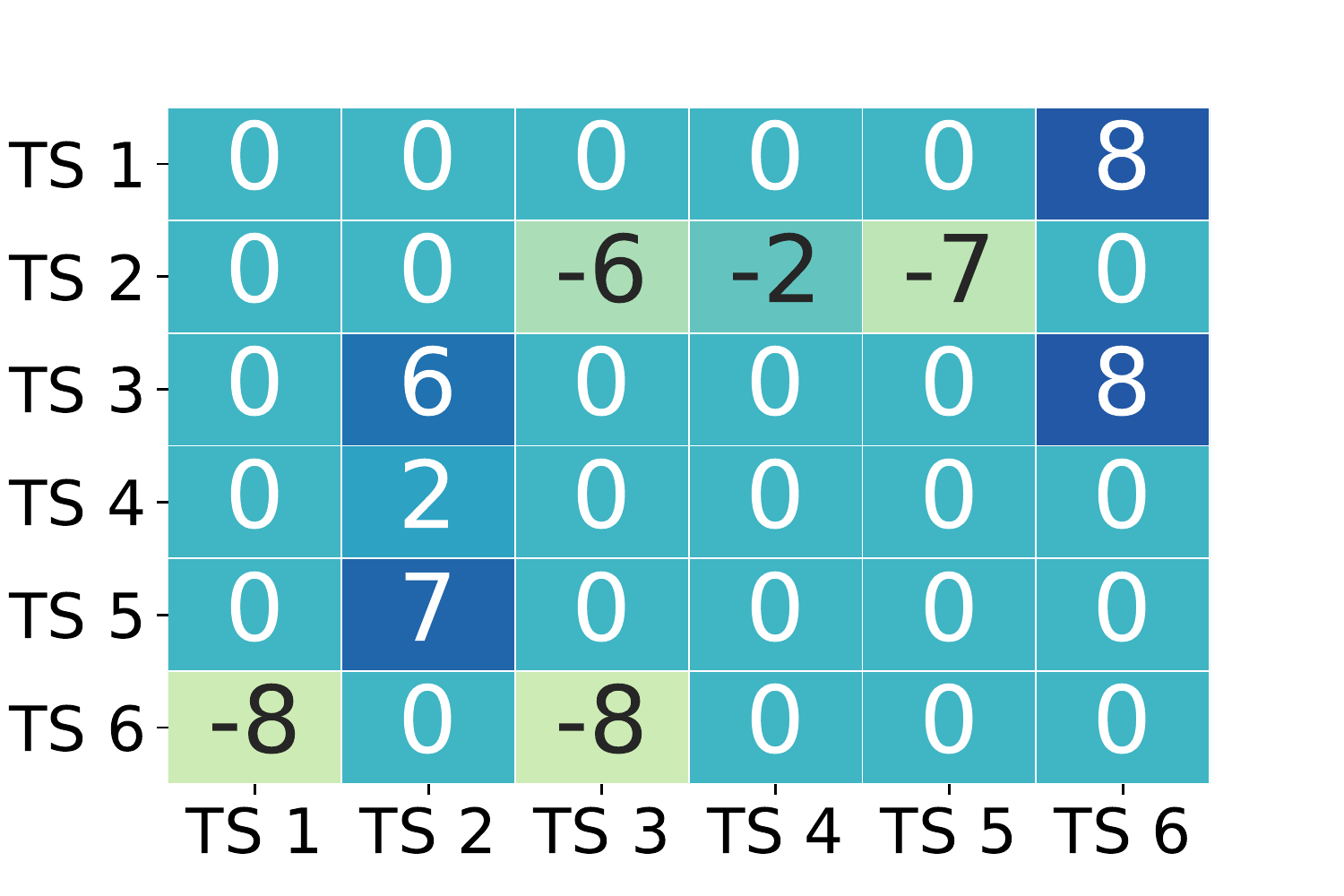}
    \\
    \multicolumn{1}{c}{$k=1$} & \multicolumn{1}{c}{$k=2$} & \multicolumn{1}{c}{$k=3$}  \\
    \end{tabular}
\captionof{figure}{Top panel: Voting matrix without voting threshold ($\theta = 1$). Middle panel: Error matrix based on mode estimation without the voting threshold ($\theta = 1$). Bottom panel: Error matrix based on median estimation without the voting threshold ($\theta = 1$).}
\label{fig:spectral_vote_matrix_before_plus_error_matrix_befores}
\end{table}

      





\begin{table}[!htbp]
  \centering
    \begin{tabular}{p{4.5cm}|p{4.5cm}p{4.5cm}}
      \multicolumn{3}{c}{\textbf{Spectral clustering}} \vspace{0.3cm}\\
      
      \multicolumn{1}{c}{\textbf{Homogeneous Setting}}    &  \multicolumn{2}{c}{\textbf{Heterogeneous Setting}}  \\

      \includegraphics[width=\linewidth,trim=0cm 0cm 1.5cm 1cm,clip]{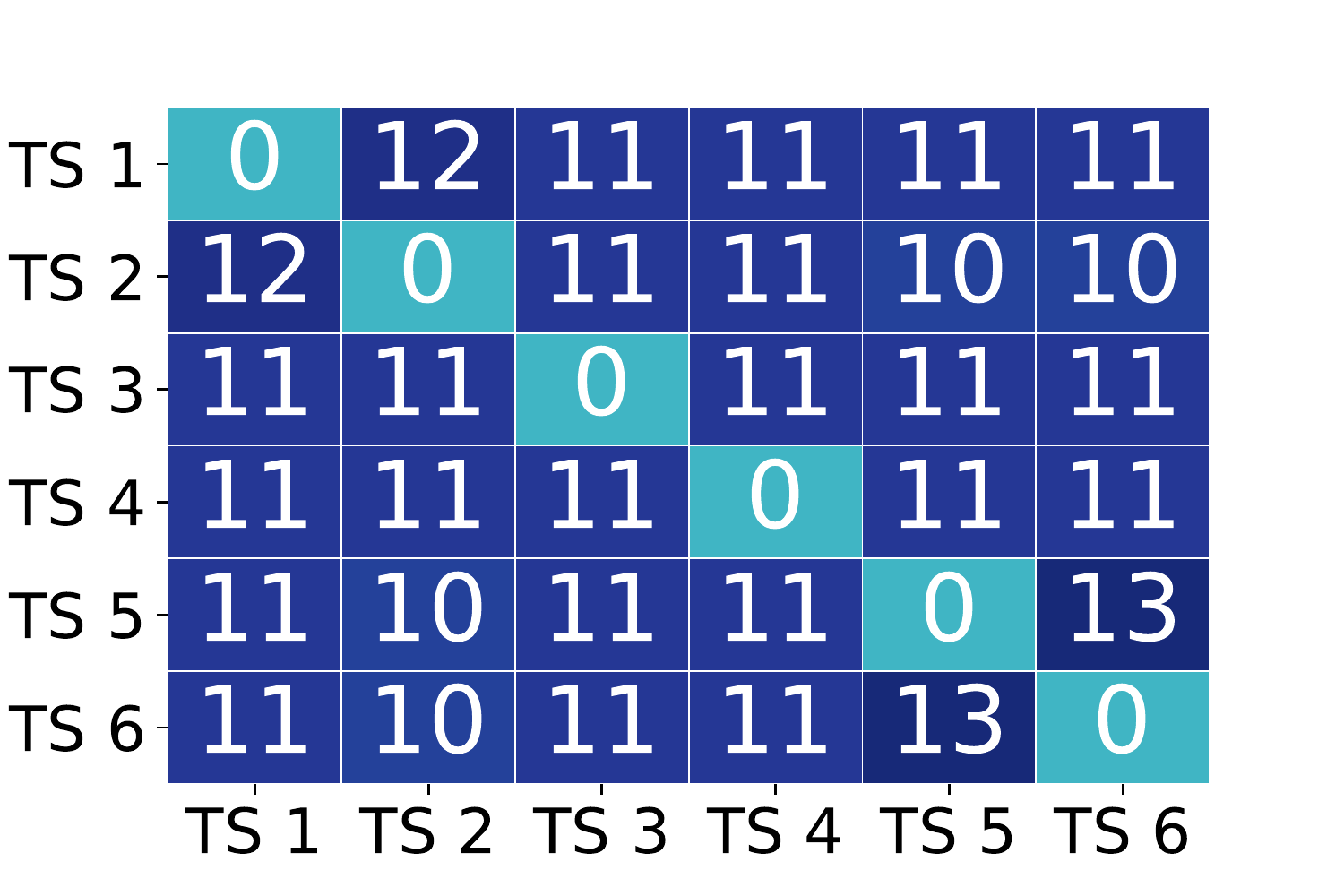}
      
      \includegraphics[width=\linewidth,trim=0cm 0cm 1.5cm 1cm,clip] {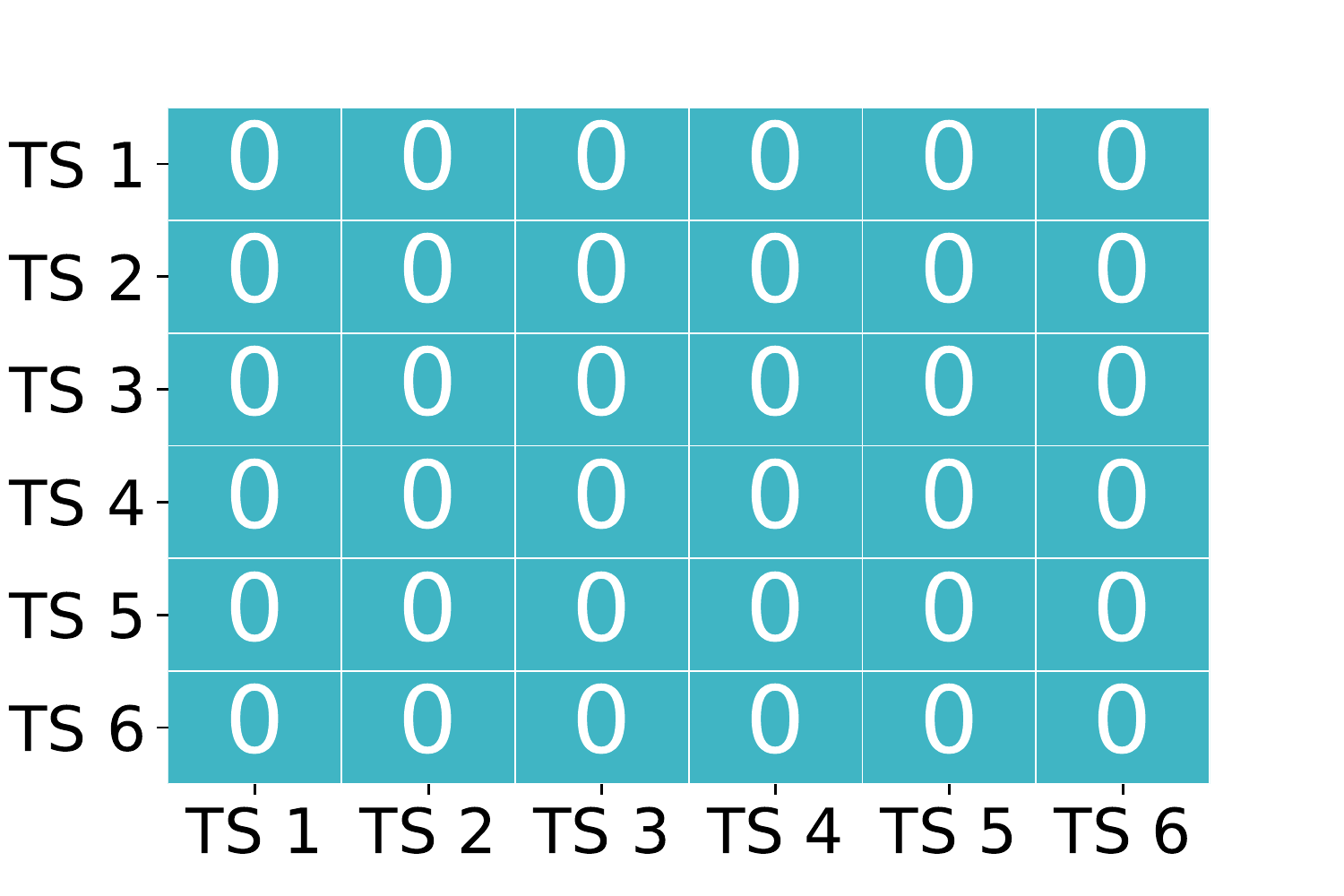}

    \includegraphics[width=\linewidth,trim=0cm 0cm 1.5cm 1cm,clip] {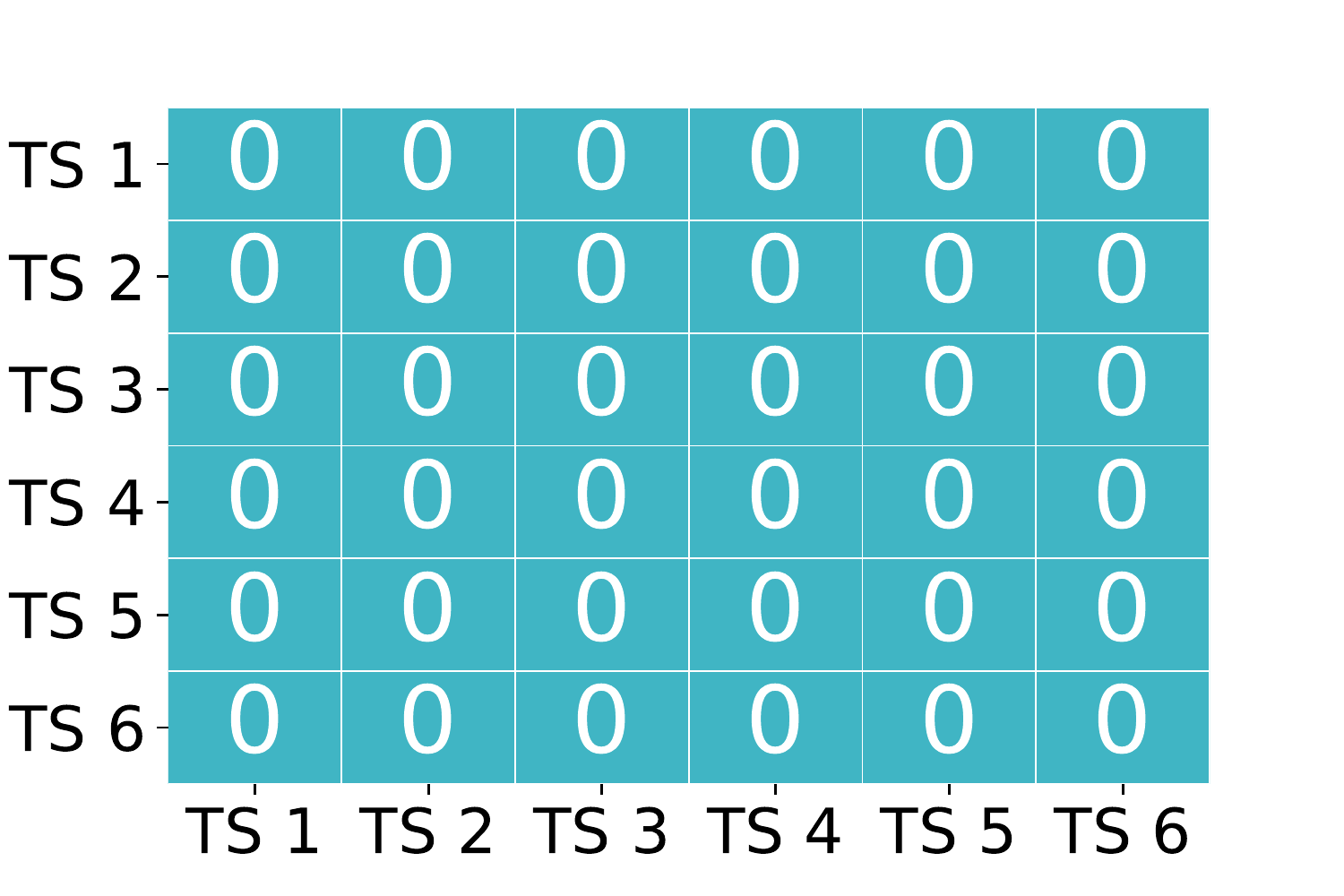}

    &
      \includegraphics[width=\linewidth,trim=0cm 0cm 1.5cm 1cm,clip]{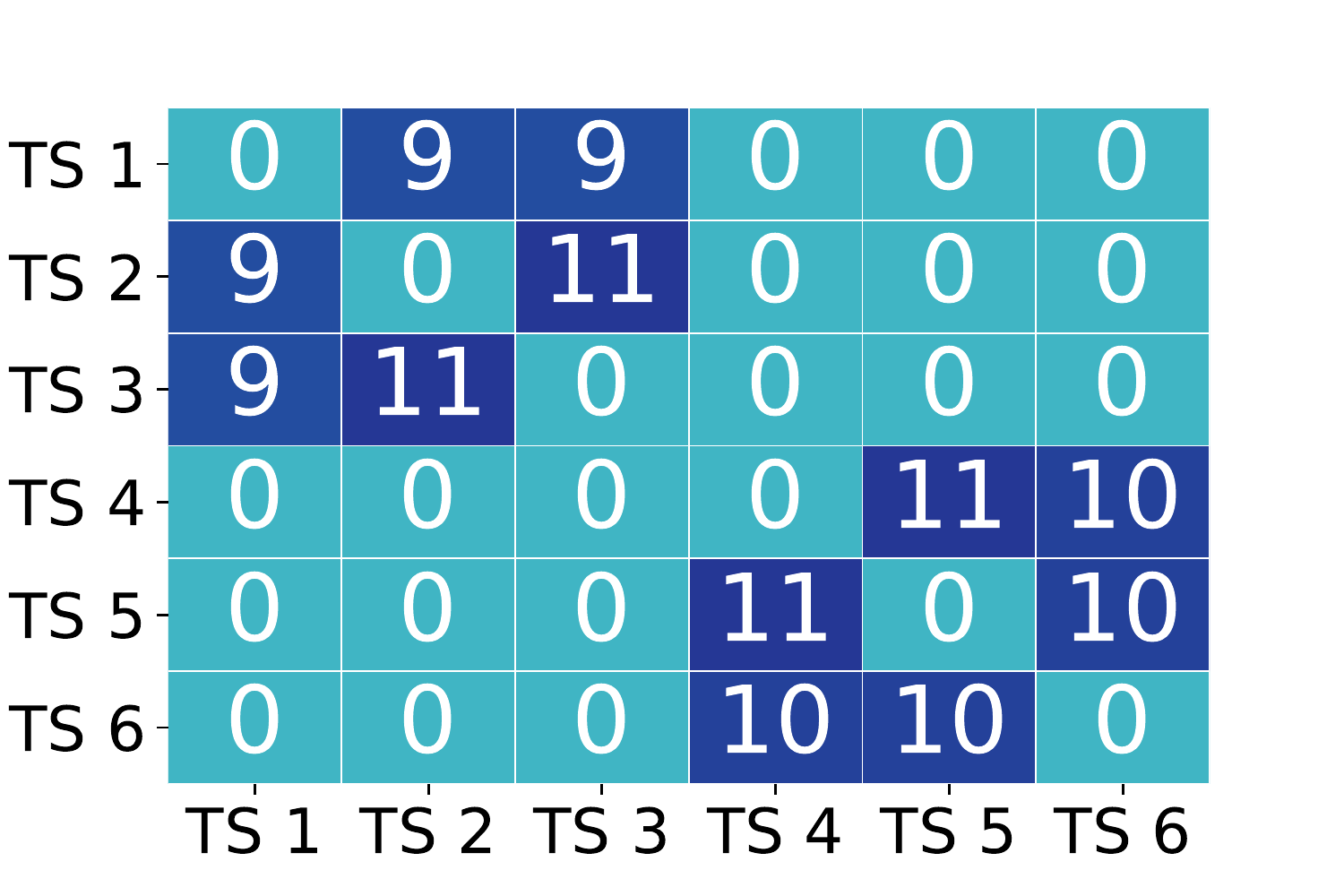}
      
      \includegraphics[width=\linewidth,trim=0cm 0cm 1.5cm 1cm,clip] {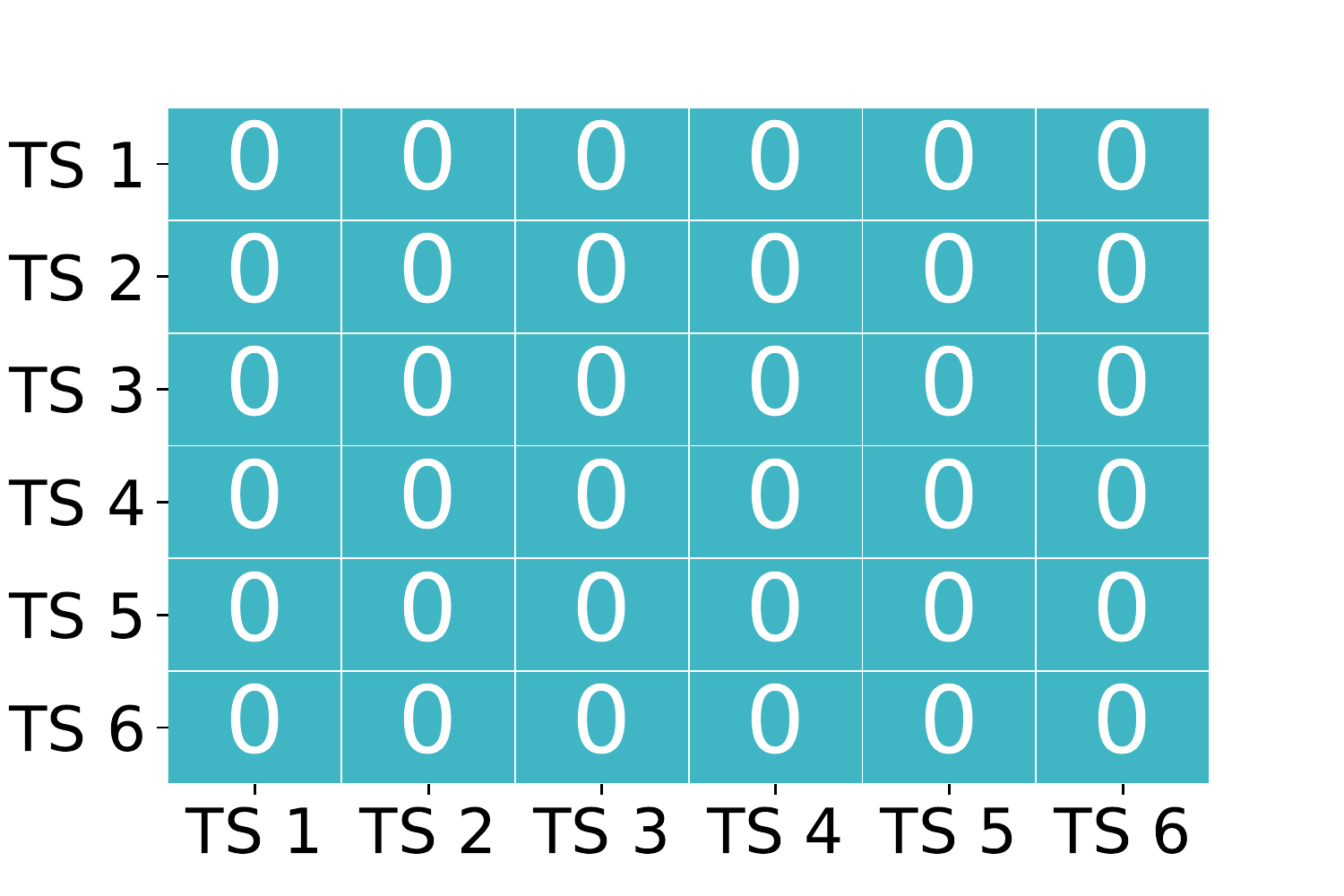}

    \includegraphics[width=\linewidth,trim=0cm 0cm 1.5cm 1cm,clip] {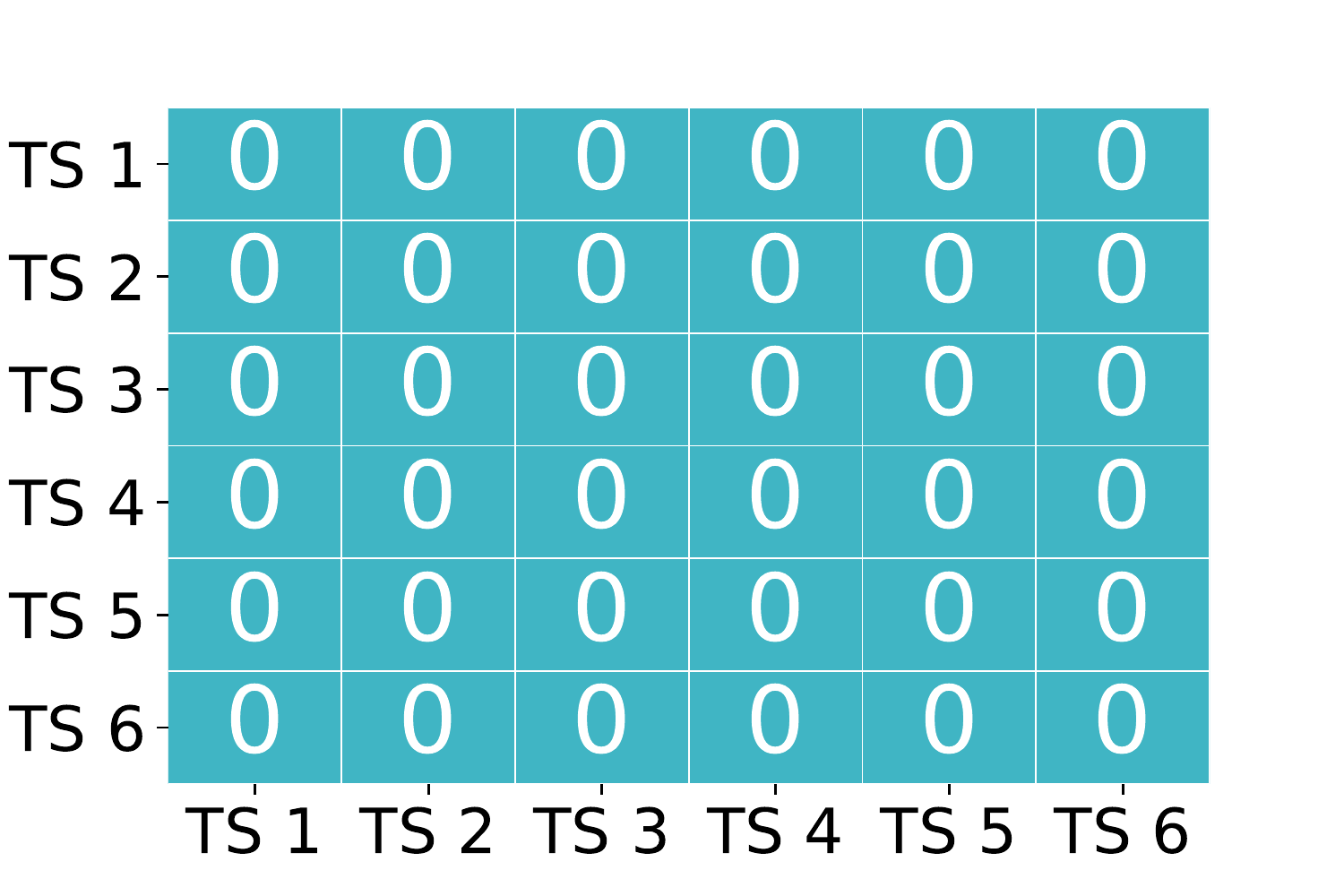}
    &
      \includegraphics[width=\linewidth,trim=0cm 0cm 1.5cm 1cm,clip]{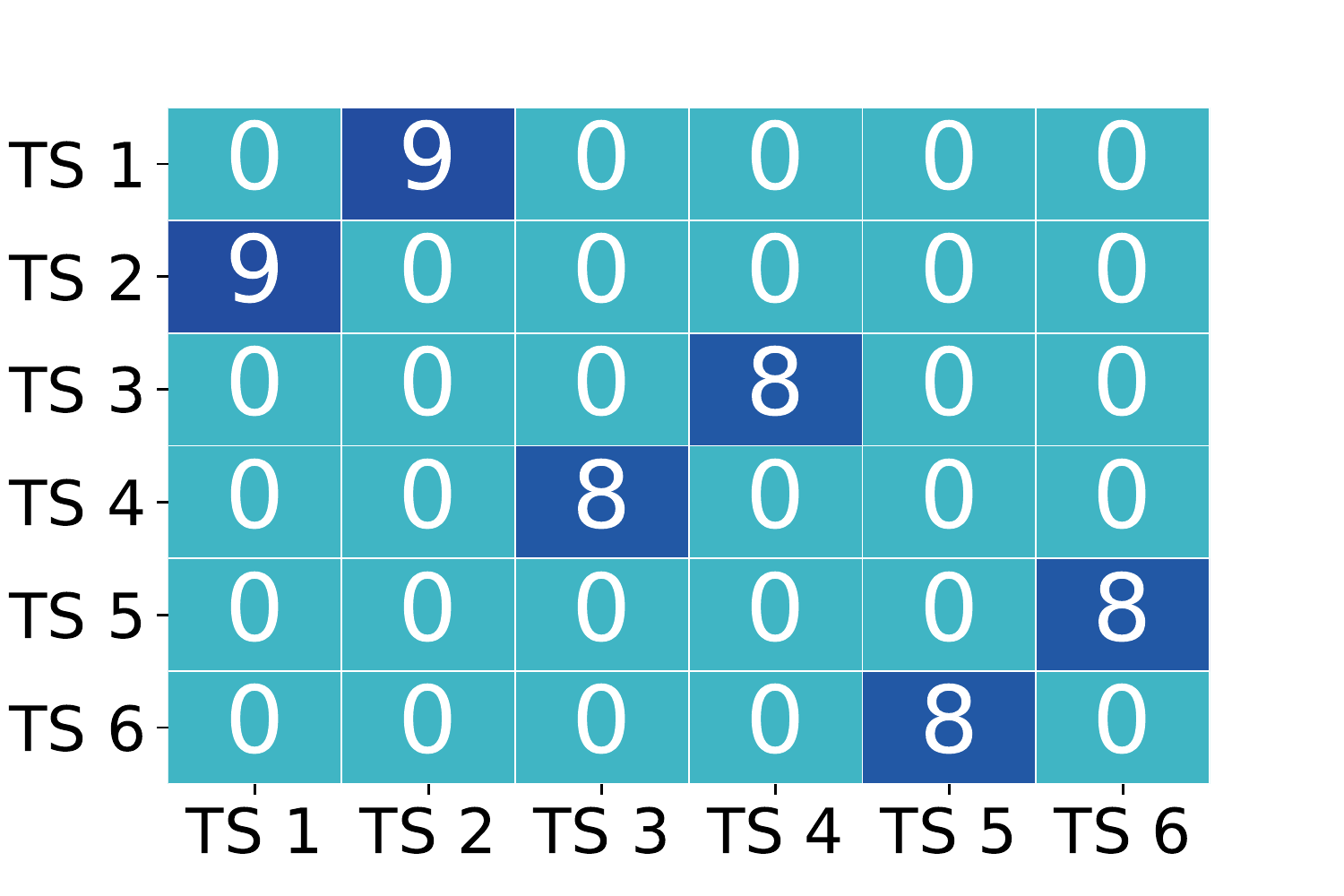}
      
      \includegraphics[width=\linewidth,trim=0cm 0cm 1.5cm 1cm,clip] {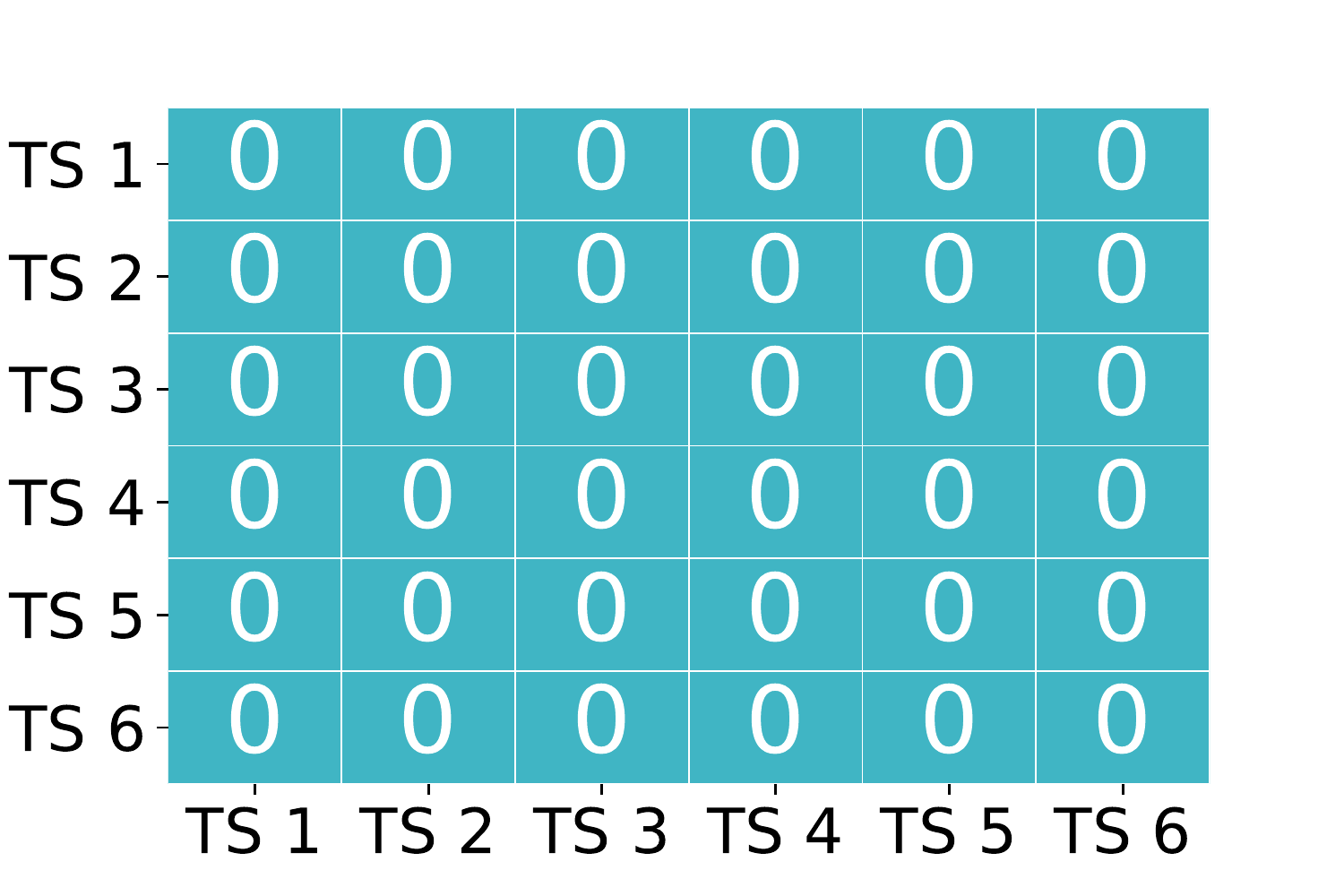}

        \includegraphics[width=\linewidth,trim=0cm 0cm 1.5cm 1cm,clip] {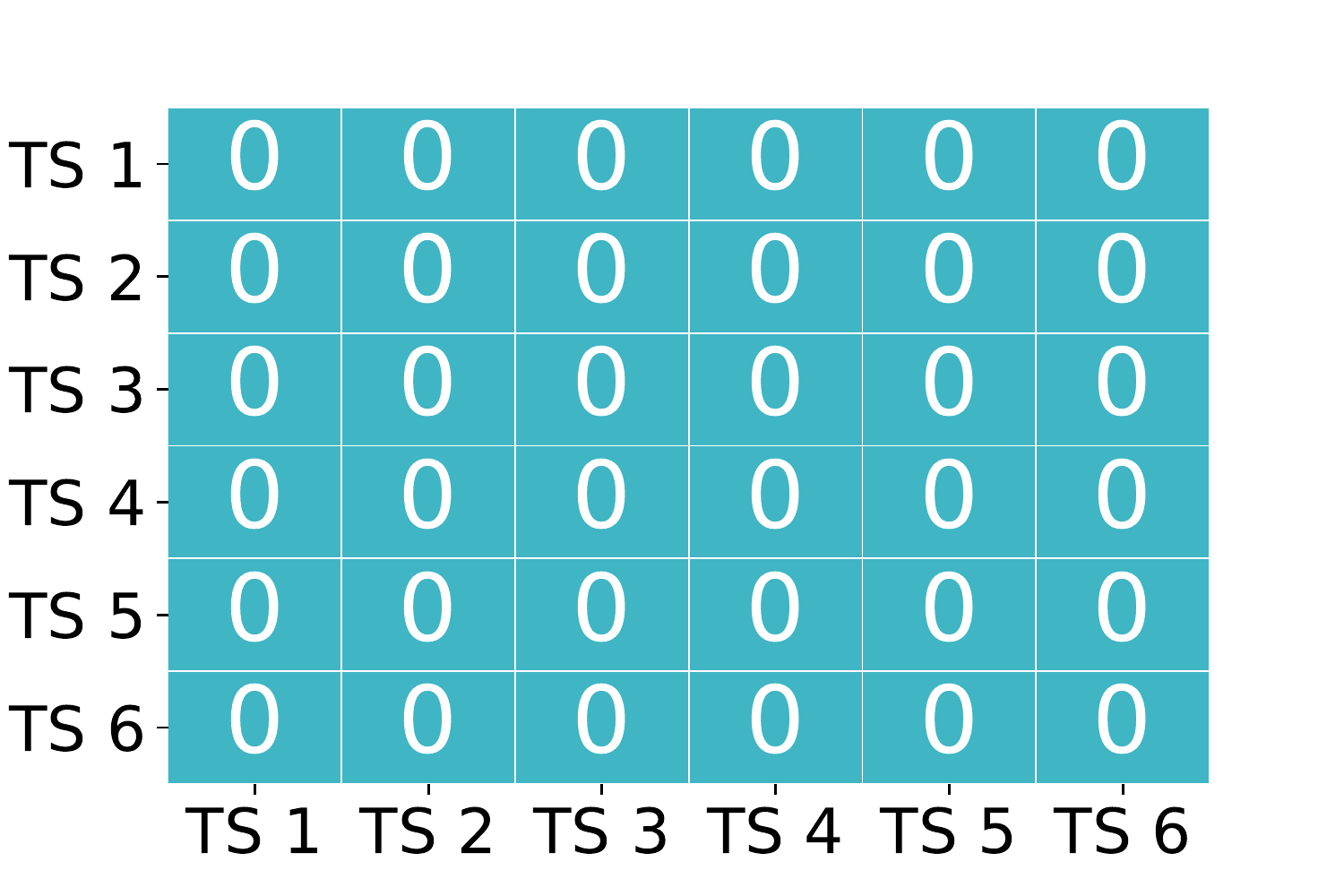}
    \\
    \multicolumn{1}{c}{$k=1$} & \multicolumn{1}{c}{$k=2$} & \multicolumn{1}{c}{$k=3$}  \\
    \end{tabular}
\captionof{figure}{Top panel: Voting matrix with voting threshold ($\theta = 6$). Middle panel: Error matrix based on mode estimation with the voting threshold ($\theta = 6$). Bottom panel: Error matrix based on median estimation with the voting threshold ($\theta = 6$).}
\label{fig:spectral_vote_matrix_after_plus_error_matrix_afters}
\end{table}

\newpage
\subsection{CO2 emissions plots}

\begin{figure}[!htbp]
\centering
\includegraphics[width=\textwidth,trim=0cm 0cm 0cm 0cm,clip]{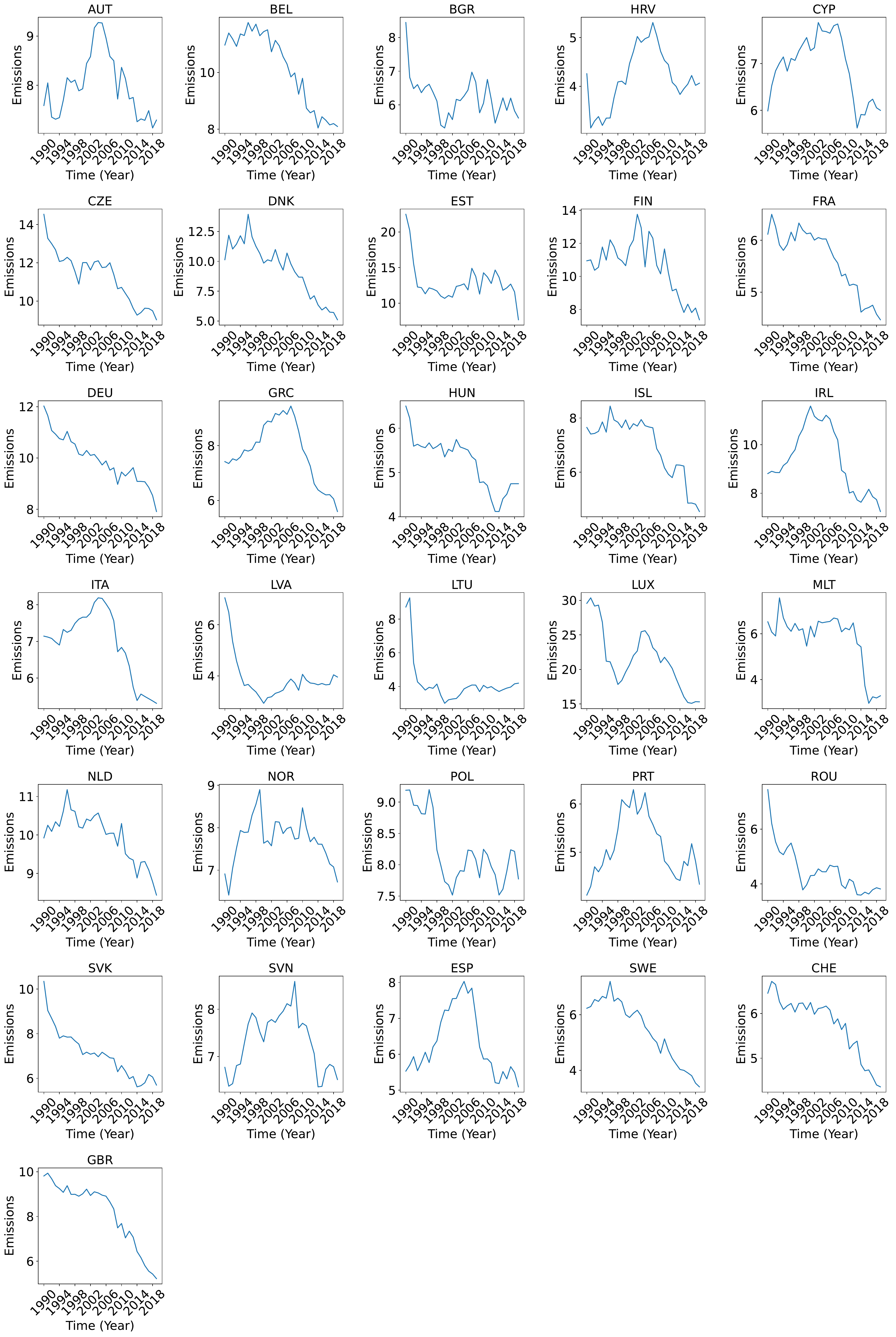}
\caption{31 countries' CO2 emissions (metric tons per capita) during the period 1990–2019.}
\label{fig:CO2_countries}
\end{figure}

Figure \ref{fig:CO2_countries} shows 31 countries’ CO2 emissions data (metric tons per capita) for the period 1990–2019 from Climate Watch.

 \newpage

\subsection{Futures data set details}
\label{sec:pinnacle}

Tables [\ref{tab: Grains}, \ref{tab: Meats}, \ref{tab: Foodfibr}, \ref{tab: Metals}, \ref{tab: Indexes}, \ref{tab: Bonds}, \ref{tab: Currency}, \ref{tab: Oils}] show the futures contracts we used and its description from the Pinnacle Data Corp CLC Database. The data is based on ratio-adjusted methods, which removes the contract-to-contract gap, yet it will not go negative as it reduces the size of the price bars if they go lower and increases them if they go higher.

\begin{table}[!htbp]
  \centering
  \caption{Grains}
    \begin{tabular}{lccccc}
        \toprule
          \textbf{Identifier}    & \textbf{Description}  \\
        \midrule
            KW  & WHEAT, KC \\
            MW  & WHEAT, MINN \\
            NR  & ROUGH RICE \\
            W\_  & WHEAT, CBOT \\ 
            ZC  & CORN, Electronic \\
            ZL  & SOYBEAN OIL, Electronic \\
            ZM  & SOYBEAN MEAL, Electronic \\
            ZO  & OATS, Electronic \\ 
            ZR  & ROUGH RICE, Electronic \\
            ZS  & SOYBEANS, Electronic \\
            ZW  & WHEAT, Electronic \\
        \bottomrule
    \end{tabular}
\label{tab: Grains}
\end{table}

\begin{table}[!htbp]
  \centering
  \caption{Meats}
    \begin{tabular}{lccccc}
        \toprule
          \textbf{Identifier}    & \textbf{Description}  \\
        \midrule
            DA  & MILK III, Comp. \\
            ZF  & FEEDER CATTLE, Electronic \\
            ZT  & LIVE CATTLE, Electronic \\
            ZZ  & LEAN HOGS, Electronic \\
        \bottomrule
    \end{tabular}
\label{tab: Meats}
\end{table}

\begin{table}[!htbp]
  \centering
  \caption{Wood fibre}
    \begin{tabular}{lccccc}
        \toprule
          \textbf{Identifier}    & \textbf{Description}  \\
        \midrule
            LB  & LUMBER \\
        \bottomrule
    \end{tabular}

\label{tab: Foodfibr}
\end{table}

\begin{table}[!htbp]
  \centering
  \caption{Metals}
    \begin{tabular}{lccccc}
        \toprule
          \textbf{Identifier}    & \textbf{Description}  \\
        \midrule
            ZG  & GOLD, Electronic \\
            ZI  & SILVER, Electronic \\
            ZP  & PLATINUM, electronic \\
            ZA  & PALLADIUM, electronic \\
            ZK  & COPPER, electronic \\
        \bottomrule
    \end{tabular}
\label{tab: Metals}
\end{table}

\begin{table}[!htbp]
  \centering
  \caption{Indexes}
    \begin{tabular}{lccccc}
        \toprule
          \textbf{Identifier}    & \textbf{Description}  \\
        \midrule
            AX  & GERMAN DAX INDEX \\
            CA  & CAC40 INDEX \\
            DX  & US DOLLAR INDEX \\
            EN  & NASDAQ, MINI \\
            ES  & S \& P 500, MINI \\
            GI  & GOLDMAN SAKS C. I. \\
            LX  & FTSE 100 INDEX \\
            MD  & S \& P 400 (Mini electronic) \\
            NK  & NIKKEI INDEX \\
            SC  & S \& P 500, composite \\
        \bottomrule
    \end{tabular}
\label{tab: Indexes}
\end{table}

\begin{table}[!htbp]
  \centering
  \caption{Bonds}
    \begin{tabular}{lccccc}
        \toprule
         \textbf{Identifier}    & \textbf{Description}  \\
        \midrule
            DT &  EURO BOND (BUND) \\
            FB &  T-NOTE, 5yr composite \\
            GS &  GILT, LONG  BOND \\
            SS & STERLING, SHORT \\
            TY & T-NOTE, 10yr composite \\
            TU & T-NOTES, 2yr composite \\
            US & T-BONDS, composite \\
            UB & EURO BOBL \\
            UZ & EURO SCHATZ \\
        \bottomrule
    \end{tabular}
\label{tab: Bonds}
\end{table}

\begin{table}[!htbp]
  \centering
  \caption{Currency}
    \begin{tabular}{lccccc}
        \toprule
          \textbf{Identifier}    & \textbf{Description}  \\
        \midrule
            AN & AUSTRALIAN \$\$\, composite\\ 
            BN & BRITISH POUND, composite \\
            CN & CANADIAN \$\$\, composite \\
            EC & EURODOLLAR, composite \\
            FN & EURO, composite \\
            JN & JAPANESE YEN, composite \\
            MP & MEXICAN PESO \\
            SN & SWISS FRANC, composite \\
        \bottomrule
    \end{tabular}
\label{tab: Currency}
\end{table}

\begin{table}[!htbp]
  \centering
  \caption{Oils}
    \begin{tabular}{lccccc}
        \toprule
          \textbf{Identifier}    & \textbf{Description}  \\
        \midrule
            ZB  & RBOB, Electronic \\
            ZH  & HEATING OIL, electronic \\
            ZN  & NATURAL GAS, electronic \\
            ZU  & CRUDE OIL, Electronic \\
        \bottomrule
    \end{tabular}
\label{tab: Oils}
\end{table}



\end{document}